%% file: neurips_2026.tex
\documentclass{article}

\PassOptionsToPackage{numbers,compress}{natbib}

\usepackage[preprint]{neurips_2026}

\usepackage[utf8]{inputenc}
\usepackage[T1]{fontenc}

\usepackage{hyperref}
\usepackage{url}

\usepackage{booktabs}
\usepackage{multirow}
\usepackage{makecell}
\usepackage{graphicx}
\usepackage{adjustbox}

\usepackage[table]{xcolor}
\usepackage{pifont}

\usepackage{amsfonts}
\usepackage{nicefrac}
\usepackage{microtype}
\usepackage{enumitem}
\usepackage{wrapfig}
\usepackage{amsmath}

\usepackage{array}

\usepackage[most]{tcolorbox}
\usepackage{caption}

\usepackage{tabularx}

\title{MentalHospital: A Virtual Environment for Evaluating Psychiatric Clinical Encounters}

\author{%
  \textbf{Yuming Yang\textsuperscript{1},
  Xiao Sun\textsuperscript{1},
  Yuanwei Zou\textsuperscript{1},
  Zhengxiao Wu\textsuperscript{1},
  Yun Chen\textsuperscript{2}}
  \\
  \textbf{Jiang Zhong\textsuperscript{1,\dag},
  Haoyang Zeng\textsuperscript{1},
  Jingwang Huang\textsuperscript{1},
  Kaiwen Wei\textsuperscript{1,\dag}}
  \\
  \textsuperscript{1}School of Computer Science, Chongqing University, Chongqing, China
  \\
  \textsuperscript{2}School of Computer Science, Hunan University, Changsha, China
  \\
  \textsuperscript{\dag}Corresponding authors.
  \\
  \texttt{ymyang@cqu.edu.cn},
  \texttt{zhongjiang@cqu.edu.cn},
  \texttt{weikaiwen@cqu.edu.cn}
}

\begin{document}

\maketitle

\begin{abstract}
\label{sec:abstract}
Large language models (LLMs) have shown strong performance on isolated psychiatric tasks, including dialogue, diagnosis, and treatment planning, yet existing benchmarks rarely simulate complete psychiatric clinical encounters. We introduce \textbf{MentalHospital}, a virtual evaluation environment for LLM-based psychiatric clinical encounters. MentalHospital instantiates the Subjective Interviewing, Objective Examination, Diagnostic Assessment, and Treatment Planning (S.O.A.P.) workflow, using skill-augmented standardized patients constructed from 1,193 de-identified psychiatric electronic health record (EHR) cases spanning all major ICD-11 categories and 76 disorders. Each encounter is assessed through a dual-track protocol that combines objective comparison against EHR-derived references with subjective assessment of clinical process quality. To scale specialist judgment, we develop \textbf{MentalEval}, five domain-specific evaluators covering communication empathy, interviewing professionalism, clinical-note quality, diagnostic rigor, and treatment appropriateness, trained with rubric-grounded SFT and expert-guided DPO. Survey responses from 22 clinicians support MentalHospital's clinical fidelity (3.88/5), while MentalEval achieves strong expert alignment with an average QWK of 0.944. Benchmarking shows that even the strongest LLM trails clinicians by 37.28 percentage points in objective psychiatric competence, with mental status assessment as a key bottleneck.
\end{abstract}

\vspace{-0.5em}
\section{Introduction}
\vspace{-0.5em}
Large language models (LLMs) have made substantial progress in psychiatry~\citep{lee2021artificial,kolding2025use}. Existing studies have reported strong performance on key clinical components, including dialogue~\citep{arora2025healthbench,zhang2024cpsycoun,racha2026omind,liu2025interactive}, medical record understanding~\citep{patel2022neuroblu}, diagnostic reasoning~\citep{sun2026mentalseek}, and treatment planning~\citep{zhang2025cbt}. However, success on isolated clinical components remains insufficient to reflect the integrated capability required in real-world clinical encounters.


Recent studies have increasingly recognized the limitations of isolated evaluation~\citep{agrawal2025evaluation,li2025counselbench}. Accordingly, several works have begun exploring virtual medical environments for more comprehensive clinical simulation and assessment~\citep{schmidgall2024agentclinic,fan2025ai,liu2026openhospital,jiang2025medagentbench}. Yet, as summarized in Table~\ref{tab:benchmark_comparison}, these environments still exhibit three limitations: \textit{(1) Encounter Incompleteness}. Existing environments still center on doctor--patient communication, rather than complete encounters across multiple clinical scenarios and tasks. 
\textit{(2) Data Inauthenticity}. The environmental evidence and reference targets are often derived from synthetic or web-based data, limiting clinical fidelity. 
\textit{(3) Patient Simplification}. Virtual patients are often reactive symptom carriers, rather than faithful and distinctive psychiatric individuals.

To address these limitations, we introduce \textbf{MentalHospital}, to our knowledge the first virtual evaluation environment designed for psychiatric clinical encounters. MentalHospital moves beyond dialogue-centered simulation by instantiating the full S.O.A.P. workflow~\citep{donohoe2015implementing}, from interviewing and examination to diagnosis and treatment planning. It is grounded in 1,193 de-identified psychiatric EHR cases covering all major ICD-11 psychiatric categories and 76 disorder-level diagnoses, and constructs skill-augmented standardized patients with individualized symptom presentation and dynamic memory recall. In each interaction, LLMs must elicit evidence from the standardized patient, request auxiliary examinations from hospital-side modules, generate clinical notes and disorder-level decisions, and finally produce individualized treatment plans.

\input{tables/benchmark_comparation}

MentalHospital evaluates each encounter through \textit{objective comparison} against EHR-derived references and \textit{subjective assessment} of clinical process quality. The former can be computed from structured clinical targets, whereas the latter requires specialist judgment and is difficult to scale with clinician review alone. Our validation further shows that general-purpose LLM judges are insufficiently reliable for specialist clinical assessment: LLM-as-a-Judge achieves only a QWK of 0.677, an accuracy of 0.225, and an MAE of 0.875. To address this gap, we develop \textbf{MentalEval}, a suite of five domain-specific evaluators covering communication empathy, interviewing professionalism, clinical-note quality, diagnostic rigor, and treatment appropriateness. MentalEval is trained in two stages: \textit{(1) Rubric-grounded SFT}, which teaches evaluators to generate rubric-based rationales and ratings from multi-level evaluation trajectories, and \textit{(2) Expert-guided DPO}, which further aligns evaluator preferences with clinician-selected judgments.

We conduct experiments with LLMs, clinicians, medical trainees, and crowdworkers. Expert and trainee ratings validate the clinical fidelity of MentalHospital, with an overall score of 3.96/5, while ablations show that the full patient construction achieves the best objective faithfulness and subjective fidelity. MentalEval also achieves strong expert alignment, reaching an average QWK of 0.944 after DPO and outperforming LLM-as-a-Judge, Multi-LLM Jury, and individual crowdworker baselines. Using this evaluation setting, we find that current LLMs still lag behind clinicians in objective psychiatric competence: even the strongest medical-specific model trails medical trainees by 27.17 percentage points and human experts by 37.28 points on average across objective metrics, while showing complementary potential in empathic communication and treatment assistance. Fine-grained analysis further reveals a key weakness in mental status assessment. Finally, survey responses from 22 clinicians indicate positive acceptance of MentalHospital and its perceived value for psychiatric training. Our contributions are summarized as follows:
\begin{itemize}[leftmargin=1.5em, itemsep=0.2em, topsep=0.2em, parsep=0pt, partopsep=0pt]
    \item We introduce \textbf{MentalHospital}, an EHR-grounded benchmark environment for full S.O.A.P. psychiatric encounters, using skill-augmented standardized patients constructed from 1,193 de-identified psychiatric EHR cases spanning all major ICD-11 categories and 76 disorders.
    
    \item We design a dual-track evaluation protocol that combines objective comparison against EHR-derived references with subjective assessment of clinical process quality, and develop \textbf{MentalEval}, a suite of specialist-aligned evaluators for scalable subjective assessment.
    
    \item We conduct comprehensive experiments with LLMs, clinicians, medical trainees, and crowdworkers, validating MentalHospital's clinical fidelity and MentalEval's expert alignment while revealing persistent gaps between current LLMs and clinicians in mental status assessment.
\end{itemize}

\begin{figure}[t]
\centering
  \includegraphics[width=\linewidth]{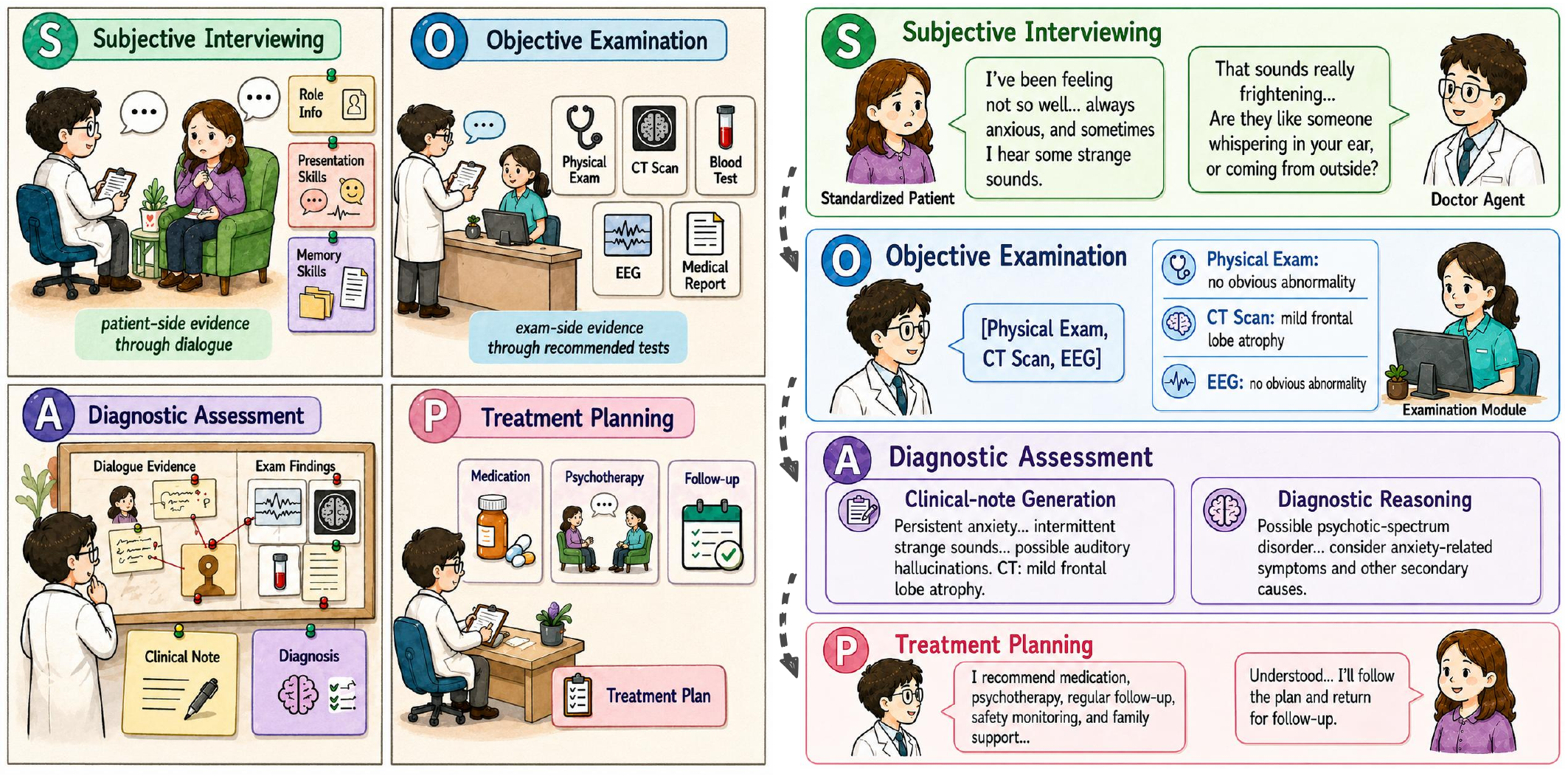}
\caption{
Overview of MentalHospital. A virtual psychiatric evaluation environment for LLM-based medical AI, built on the S.O.A.P. workflow and designed for dual-track evaluation.
}
\vspace{-1.0em}
  \label{fig:intro}
\end{figure}

\vspace{-0.5em}
\section{\textsc{MentalHospital}}
\vspace{-0.5em}
\vspace{-0.5em}
MentalHospital is constructed from 1,193 de-identified psychiatric EHR cases obtained through an approved collaborative data-use partnership with a coordinating medical center. The resulting dataset comprises multi-center records sourced from six medical centers across three regions. The dataset includes 885 single-diagnosis cases and 308 comorbid cases. These cases cover all major ICD-11 psychiatric categories and 76 disorder-level diagnoses, providing broad disorder-level diversity for virtual psychiatric simulation (see Appendix~\ref{data_source_and_case_composistion}). Each case was independently reviewed by two licensed psychiatrists to ensure clinical completeness and consistency (Cohen's $\kappa > 0.8$). Details see Appendix~\ref{app:data_distribution}. Each EHR case is converted into a structured MentalHospital case $c$ with patient-side evidence, examination-side evidence, and reference clinical targets:
\begin{equation}
x_c = \{\mathcal{K}^{\mathrm{pat}}_c, \mathcal{K}^{\mathrm{exam}}_c\},
\qquad
\mathcal{Y}^{*}_c =
\{y^{\mathrm{cat}*}_c, y^{\mathrm{dis}*}_c, y^{\mathrm{treat}*}_c\}.
\end{equation}
Here, $\mathcal{K}^{\mathrm{pat}}_c$ contains patient-side evidence such as chief complaint, illness history, past and family history, and mental status; $\mathcal{K}^{\mathrm{exam}}_c$ contains examination-side evidence such as physical and auxiliary examination findings. The reference targets $\mathcal{Y}^{*}_c$ include the diagnostic category $y^{\mathrm{cat}*}_c$, the disorder-level diagnosis $y^{\mathrm{dis}*}_c$, and EHR-derived treatment information $y^{\mathrm{treat}*}_c$. In MentalHospital, $\mathcal{K}^{\mathrm{pat}}_c$ is used to instantiate the standardized patient, $\mathcal{K}^{\mathrm{exam}}_c$ supports the hospital-side examination module, and $\mathcal{Y}^{*}_c$ serves as the reference for diagnostic and treatment evaluation. Details see Appendix~\ref{app:data_processing}.

To protect privacy while preserving clinical meaning, each EHR is processed using an on-premise \texttt{DeepSeek-R1}~\citep{liu2025deepseek}, rule-based detectors, and expert review. Sensitive events, exact dates, locations, institutions, and other identifiable details are generalized or masked. Clinically relevant sensitive events are rewritten into semantically similar but non-identifiable descriptions, so that psychiatric evidence remains usable for simulation and evaluation.

The study protocol and annotation procedures were approved by the institutional ethics boards of all participating clinical centers. All data collection, processing, and annotation followed the \textit{Declaration of Helsinki}. Since records were de-identified before construction, individual informed consent was waived. All processing was conducted within secure hospital environments using locally deployed models, and no identifiable patient information or intermediate records left institutional governance.

\vspace{-0.5em}
\subsection{Interaction Process}
\vspace{-0.5em}

Given a psychiatric case $c$, MentalHospital instantiates a case-conditioned sequential environment:
\begin{equation}
\mathcal{E}_c = \{\mathcal{P}_c, \mathcal{M}^{\mathrm{exam}}_c\},
\end{equation}
where $\mathcal{P}_c$ denotes the standardized patient constructed from patient-side evidence $\mathcal{K}^{\mathrm{pat}}_c$, and $\mathcal{M}^{\mathrm{exam}}_c$ denotes the hospital-side examination module supported by examination-side evidence $\mathcal{K}^{\mathrm{exam}}_c$. The doctor agent cannot directly access $\mathcal{K}^{\mathrm{pat}}_c$, $\mathcal{K}^{\mathrm{exam}}_c$, or $\mathcal{Y}^{*}_c$, but must recover clinical evidence and infer decisions through interaction.

As shown in Figure~\ref{fig:intro}, MentalHospital follows the clinical S.O.A.P. workflow. In the subjective interviewing stage, the doctor asks psychiatric interview questions and receives patient responses from $\mathcal{P}_c$, aiming to recover patient-side evidence such as symptoms, illness history, and mental status. In the objective examination stage, the doctor recommends examinations, and $\mathcal{M}^{\mathrm{exam}}_c$ returns the corresponding EHR-derived results when available. In the diagnostic assessment stage, the doctor integrates the dialogue trajectory and examination findings to generate a clinical note and diagnostic hypotheses. In the treatment planning stage, the doctor produces a treatment plan based on the accumulated evidence and diagnostic assessment. The interaction history is updated step by step:
\begin{equation}
h_t = h_{t-1} \cup \{(a_t,o_t)\},
\end{equation}
where $a_t$ is the doctor action and $o_t$ is the environment observation returned by either the standardized patient or the examination module. After completing the workflow, each episode produces
\begin{equation}
\hat{y}_c =
\left\{
\tau_c,\,
\hat{\mathcal{A}}^{\mathrm{exam}}_c,\,
\hat{n}_c,\,
\hat{d}_c,\,
\hat{p}_c
\right\},
\end{equation}
where $\tau_c$ is the doctor--patient trajectory, $\hat{\mathcal{A}}^{\mathrm{exam}}_c$ is the recommended examination set, $\hat{n}_c$ is the clinical note, $\hat{d}_c$ is the diagnosis with reasoning, and $\hat{p}_c$ is the treatment plan. Detailed interaction details are provided in the Appendix~\ref{app:interaction_prompt}.

\vspace{-0.5em}
\subsection{Environment Modules}
\vspace{-0.5em}

MentalHospital consists of an environment controller, a standardized patient agent, a hospital-side examination module, and the doctor agent under evaluation. The controller manages the S.O.A.P. stage transitions, routes doctor actions to the corresponding module, records intermediate outputs, and issues instructions for completing the episode. See the Appendix~\ref{app: module_prompts} for implementation details.
\vspace{-0.5em}
\paragraph{Environment Controller.}
The controller coordinates the S.O.A.P. interaction process. It maintains the interaction history $h_t$, records the doctor--patient trajectory $\tau_c$, collects examination requests $\hat{\mathcal{A}}^{\mathrm{exam}}_c$, and stores downstream outputs including the clinical note $\hat{n}_c$, diagnosis $\hat{d}_c$, and treatment plan $\hat{p}_c$. It ensures that the doctor agent interacts with the appropriate module at each stage.
\vspace{-0.5em}
\paragraph{Standardized Patient.}
The standardized patient $\mathcal{P}_c$ is constructed from patient-side evidence $\mathcal{K}^{\mathrm{pat}}_c$ as a skill-augmented patient configuration:
\begin{equation}
\mathcal{C}_c=\{\mathcal{R}_c,\mathcal{S}^{\mathrm{pre}}_c,\mathcal{S}^{\mathrm{mem}}_c\},
\end{equation}
where $\mathcal{R}_c$ specifies role information, $\mathcal{S}^{\mathrm{pre}}_c$ encodes individualized symptom presentation grounded in the EHR mental status examination, and $\mathcal{S}^{\mathrm{mem}}_c$ stores patient-side evidence as topic-level memory units. During interaction, the patient maintains an active memory state rather than exposing all memory units at once. Given the doctor action $a_t$ and history $h_{t-1}$, the memory skill determines whether the current topic should be maintained or switched:
\begin{equation}
m_t=\mathrm{UpdateMem}(m_{t-1},a_t,h_{t-1},\mathcal{S}^{\mathrm{mem}}_c).
\end{equation}
After updating the active memory, the patient generates a content-level response from the current recalled experience and role information, and then renders it with the presentation skill:
\begin{equation}
\tilde{u}_t^{\mathrm{pat}}
\sim
\pi_{\mathrm{pat}}^c(\cdot\mid h_{t-1},a_t,\mathcal{R}_c,m_t),
\qquad
o_t^{\mathrm{pat}}
=
\mathrm{Render}(\tilde{u}_t^{\mathrm{pat}},\mathcal{S}^{\mathrm{pre}}_c).
\end{equation}
The final observation $o_t^{\mathrm{pat}}$ contains both behavioral presentation and verbal response, enabling the doctor agent to infer clinical evidence from both how the patient presents and what the patient states. This hybrid design ensures that the patient agent discloses only information supported by the retrieved memory while avoiding unrelated, unasked, or future-stage clinical information.

\vspace{-0.5em}
\paragraph{Examination Module.}
The examination module $\mathcal{M}^{\mathrm{exam}}_c$ is implemented as a hospital-side question-answering interface based on \texttt{DeepSeek-V4-Flash}~\citep{liu2025deepseek}. It receives examination requests from the doctor agent and returns results according to the examination-side evidence $\mathcal{K}^{\mathrm{exam}}_c$. If a requested examination exists in $\mathcal{K}^{\mathrm{exam}}_c$, the module returns the corresponding EHR-derived report; otherwise, it returns \textit{``result unavailable''}. The module also records the requested examination set $\hat{\mathcal{A}}^{\mathrm{exam}}_c$ for comparison with the EHR-derived reference set when calculating examination coverage.
\vspace{-0.5em}
\subsection{Dual-track Evaluation}
\vspace{-0.5em}
MentalHospital evaluates each completed episode through \textit{objective comparison} and \textit{subjective assessment}. Objective comparison measures outcome-level task correctness against EHR-derived references. Subjective assessment evaluates process-level clinical professionalism and interpretability using expert-defined rubrics. Detailed metrics are provided in Section~\ref{sec:experiments}.

\vspace{-0.5em}
\section{\textsc{MentalEval}}
\vspace{-0.5em}
We explored two approaches to scalable subjective evaluation: crowdworker review and rubric-prompted general-purpose LLM judges. Experiment~\ref{MentalEval_test} shows that crowdworker review is reasonably aligned with specialists (QWK = 0.904, Acc. = 0.625, MAE = 0.375), yet costly to scale, while LLM-as-a-Judge shows limited specialist agreement (QWK = 0.677, Acc. = 0.225, MAE = 0.875).We therefore develop \textbf{MentalEval}, a suite of five Qwen3-8B-based domain-specific evaluators trained with rubric-grounded SFT followed by expert-guided DPO. The evaluators assess communication empathy, interviewing professionalism, clinical-note quality, diagnostic rigor, and treatment appropriateness. Given an evaluation context such as dialogue, clinical note, diagnostic reasoning, or treatment recommendation, each evaluator outputs a rubric-grounded rationale and a Likert-style 1--5 rating. Detailed rubrics are provided in Appendix~\ref{app:rubrics}.

\begin{figure}[t]
\centering
  \includegraphics[width=\linewidth]{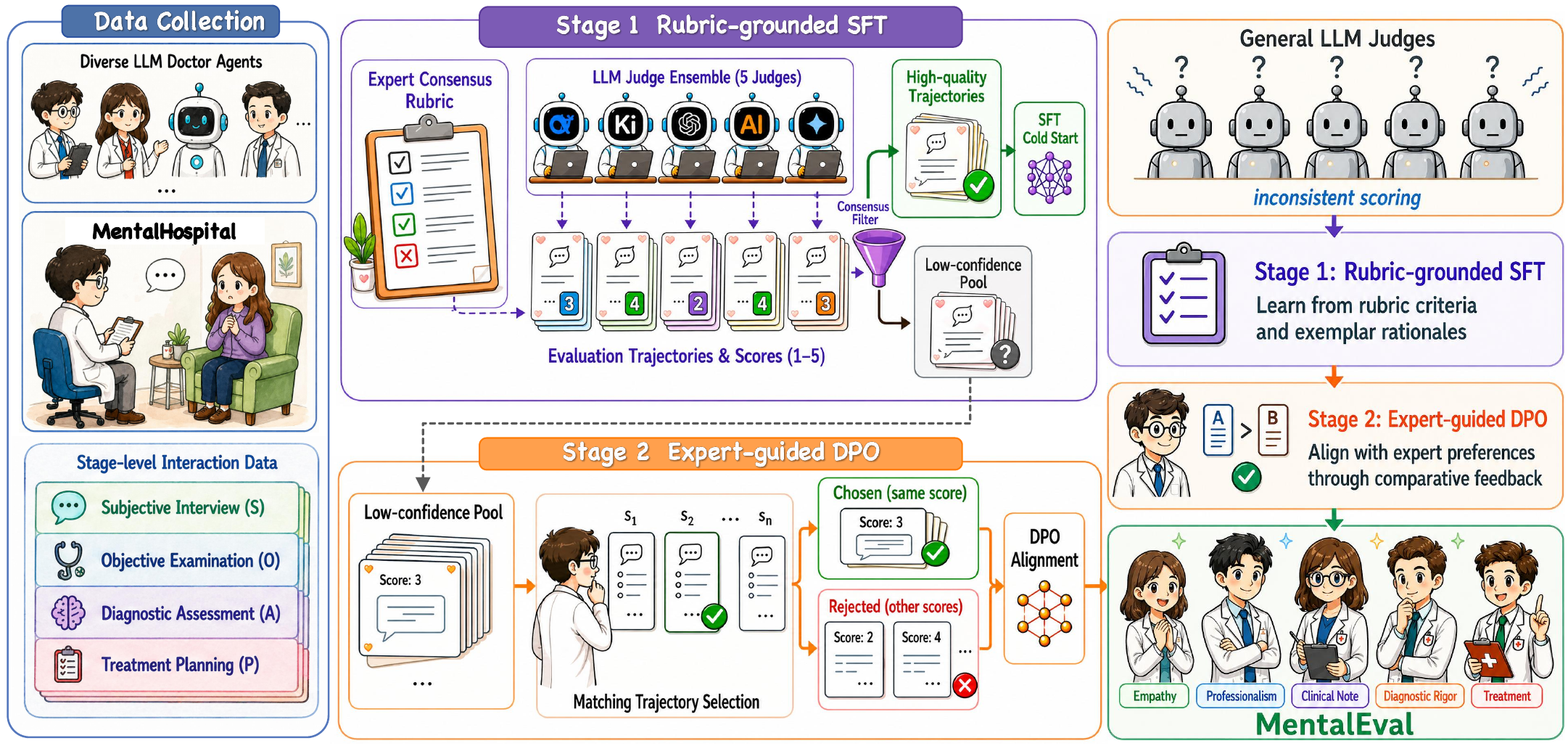}
\caption{Overview of MentalEval construction. MentalEval is trained on MentalHospital interaction data via rubric-grounded SFT and expert-guided DPO for subjective clinical evaluation.}
\vspace{-1.0em}
  \label{fig:mentaleval}
\end{figure}

\vspace{-0.5em}
\subsection{Stage 1: Rubric-grounded SFT}
\vspace{-0.5em}
The first stage provides a cold start for learning rubric-grounded evaluation patterns. We sample 200 cases for evaluator training, then select 20 LLMs covering different providers, model scales, and specialties, and run each model three times on each case, yielding complete interaction episodes. As shown in Figure~\ref{fig:mentaleval}, these episodes are further decomposed into stage-level interaction data for the five subjective evaluation dimensions. For each stage-level sample $z_i$, we use a judge ensemble $\mathcal{J}$ of five LLMs (\texttt{DeepSeek-V4-Pro}~\citep{liu2025deepseek}, \texttt{KIMI-K2.6}~\citep{team2025kimi}, \texttt{GPT-5.5}~\citep{singh2025openai}, \texttt{Claude-Opus-4.6}~\citep{bai2022constitutional}, and \texttt{Gemini-3.1-Pro}~\citep{team2024gemma}) to generate rubric-based evaluation trajectories and ratings:
\begin{equation}
(s_{i,j}, r_{i,j}) = J_j(z_i, \mathcal{B}), 
\qquad j \in \mathcal{J}, \quad r_{i,j}\in\{1,\ldots,5\},
\end{equation}
where $\mathcal{B}$ denotes the expert consensus rubric, $s_{i,j}$ is the generated evaluation trajectory, and $r_{i,j}$ is the assigned score. To filter reliable supervision, we retain only judge outputs with sufficient score agreement. Specifically, the consensus score for $z_i$ is defined as:
\begin{equation}
r_i^{*}
=
\arg\max_{r\in\{1,\ldots,5\}}
\sum_{j\in\mathcal{J}}\mathbf{1}[r_{i,j}=r].
\end{equation}
If at least $k$ judges assign $r_i^{*}$, the trajectories with score $r_i^{*}$ are retained as high-quality SFT targets; otherwise, the candidate set is treated as low-confidence and reserved for preference construction. Because extreme scores are underrepresented, we additionally select near-boundary samples and rewrite 2-point cases toward 1-point quality and 4-point cases toward 5-point quality. These augmented samples are retained only after judge-ensemble verification. The SFT objective is:
\begin{equation}
\mathcal{L}_{\mathrm{SFT}}
=
-\mathbb{E}_{(z,s)\sim\mathcal{D}_{\mathrm{SFT}}}
\log \pi_{\theta}(s\mid z).
\end{equation}

\vspace{-1em}
\subsection{Stage 2: Expert-guided DPO}
\vspace{-0.5em}

After cold-start training, we further align MentalEval with specialist judgment using expert-guided preference optimization. For each low-confidence candidate set
$\mathcal{S}_i=\{(s_{i,1},r_{i,1}),\ldots,(s_{i,5},r_{i,5})\}$, we present the original stage-level interaction data $z_i$ and the candidate evaluation trajectories to clinicians. The clinician selects the trajectory $s_i^{*}$ whose rating and rationale best match the clinical quality of $z_i$. Candidate trajectories with the same score as $s_i^{*}$ are treated as chosen responses, while the remaining trajectories are treated as rejected responses:
\begin{equation}
\mathcal{S}_i^{+}=\{s_{i,j}\mid r_{i,j}=r_i^{*}\},
\qquad
\mathcal{S}_i^{-}=\{s_{i,j}\mid r_{i,j}\ne r_i^{*}\}.
\end{equation}
We then construct preference triples $(z_i,s_i^{+},s_i^{-})$ for DPO training. The DPO objective is:
\begin{equation}
\mathcal{L}_{\mathrm{DPO}}
=
-\mathbb{E}_{(z,s^+,s^-)}
\log \sigma\left(
\beta
\left[
\log\frac{\pi_{\theta}(s^+\mid z)}{\pi_{\mathrm{ref}}(s^+\mid z)}
-
\log\frac{\pi_{\theta}(s^-\mid z)}{\pi_{\mathrm{ref}}(s^-\mid z)}
\right]
\right).
\end{equation}
Detailed MentalEval training settings are provided in Appendix~\ref{app:training_settings}.

\input{tables/experiments_main_comparasion}

\vspace{-1em}
\section{Experiments}
\vspace{-0.5em}

\paragraph{Task Definition.}
We formulate evaluation as a multi-stage, mixed-type clinical task. Given a case $c$, a doctor agent completes an episode covering subjective interviewing, objective examination, diagnostic assessment, and treatment planning. The task combines three forms: (1) open-ended generation, including dialogue, clinical-note writing, diagnostic reasoning, and treatment planning; (2) multi-label decision making, including examination recommendation and category-level diagnosis; and (3) ranked disorder prediction, for definitive diagnosis. We evaluate the resulting trajectory and stage outputs from two perspectives: (i) objective comparison, which measures EHR-grounded outcomes, and (ii) subjective assessment, which measures process-level clinical quality.
\vspace{-0.5em}
\paragraph{Evaluation Metrics.}
\label{sec:experiments}
For EHR-grounded outputs, including examination recommendation (Exam.), clinical interviewing (Inter.), and clinical-note generation (Note), we measure reference coverage:
\begin{equation}
\mathrm{Coverage}(y_c, \mathcal{K}_c)
=
\frac{1}{|\mathcal{K}_c|}
\sum_{k \in \mathcal{K}_c}
\mathbb{I}\left[ k \preceq y_c \right],
\end{equation}
where $\mathcal{K}_c$ denotes the EHR-derived reference checkpoint set for case $c$, and $k \preceq y_c$ indicates that checkpoint $k$ is semantically entailed by the generated output $y_c$. We use a conservative semantic similarity threshold of $\tau=0.85$ for high-precision automatic matching, and adjudicate pairs below the threshold with an empirically validated Multi-LLM-assisted semantic matching procedure. For definitive diagnosis, we report category identification (Cate.) and disorder-level diagnosis ranking (Diso.), with nDCG used to evaluate diagnostic prioritization. 
For subjective assessment, MentalEval scores communication empathy (Emp.), interviewing professionalism (Prof.), clinical-note quality (Qual.), diagnostic rigor (Rigor), and treatment appropriateness (App.) using expert-defined rubrics. For more details see Appendix~\ref{app:metrics}.

\vspace{-0.5em}
\paragraph{Comparison Groups.}
We consider five comparison groups: human experts, medical trainees, crowdworkers, general-purpose LLMs, and medical-specific LLMs. The expert group consists of 5 licensed psychiatrists and 3 psychology experts. We also include 14 medical trainees and 8 non-specialist crowdworkers. For LLMs, we evaluate a set of general-purpose models, including Llama-3-8B~\citep{grattafiori2024llama}, Qwen3-14B~\citep{yang2025qwen3}, GPT-OSS-20B~\citep{agarwal2025gpt}, QwQ-32B~\citep{yang2025qwen3}, GPT-OSS-120B~\citep{agarwal2025gpt}, DeepSeek-V4-Pro~\citep{liu2025deepseek}, GPT-5.5~\citep{singh2025openai}, and Claude-Sonnet-4-6~\citep{bai2022constitutional}. We also include medical-oriented LLMs, including MedGemma-4B-it~\citep{sellergren2025medgemma}, MedGemma-27B-it~\citep{sellergren2025medgemma}, and Baichuan-M2~\citep{dou2025baichuan}. 

\vspace{-0.5em}
\subsection{Can LLMs Interact Like Clinicians?}
\vspace{-0.5em}

We evaluate 12 LLMs, 8 human experts, and 14 medical trainees in MentalHospital. As shown in Table~\ref{tab:main_results}, current LLMs still show a clear gap from clinicians in objective clinical competence. Compared with the strongest medical-specific model, Baichuan-M2, medical trainees achieve higher scores across all five objective metrics, with an average gap of 27.17 percentage points; human experts further widen this gap to 37.28 points. In subjective assessment, LLMs and clinicians show complementary strengths rather than uniform superiority of either group. Human experts exhibit stronger interviewing professionalism (4.25), reflecting more efficient, goal-directed clinical information gathering, but receive lower scores in expressed communication empathy (1.22). In contrast, the best-performing LLMs reach or exceed human experts in expressed empathy (2.16 vs. 1.22), diagnostic rigor (3.14 vs. 2.54), and treatment appropriateness (3.48 vs. 2.96), while remaining lower in interviewing professionalism (3.72 vs. 4.25) and clinical-note quality (3.89 vs. 4.01). These results suggest that LLMs are not yet reliable substitutes for clinicians, but may provide complementary value in verbal affective support and evidence-informed treatment assistance.
\input{tables/experiments_comparasion_coverage}

\vspace{-0.5em}
\subsection{Where Do LLMs Fall Short in Clinical Interaction?}
\vspace{-0.5em}
\input{tables/experiments_patient_ablation}
To identify the core gap, we decompose coverage in clinical interviewing and note generation into Chief Complaint (CC) and Mental Status (MS) in Table~\ref{tab:interview_note}, current LLMs fail to recognize the clinical importance of mental status examination in psychiatric assessment and diagnosis. For example, \texttt{GPT-5.5} shows a large gap in interviewing (85.38 vs. 55.72), and \texttt{DeepSeek-v4-Pro} exhibits a similar pattern in note generation (59.94 vs. 43.91). 



\vspace{-0.5em}
\subsection{Does Skill-Augmented Patient Construction Improve Fidelity?}
\vspace{-0.5em}

We evaluate patient construction fidelity through objective evaluation and subjective assessment, with detailed experimental settings provided in Appendix~\ref{app:patient_ablation}. As shown in Table~\ref{tab:patient_ablation}, naive prompting achieves reasonable factual faithfulness but performs poorly in subjective fidelity, indicating that directly exposing the EHR does not yield realistic patient simulation. Representation skill substantially improves clinical realism and patient distinctiveness, while memory skill improves evidence coverage, precision, and hallucination control. The full construction achieves the best overall performance.

\vspace{-0.5em}
\input{tables/experiments_mentaleval_ablation}
\vspace{-0.5em}
\begin{figure}[t]
\centering
  \includegraphics[width=\linewidth]{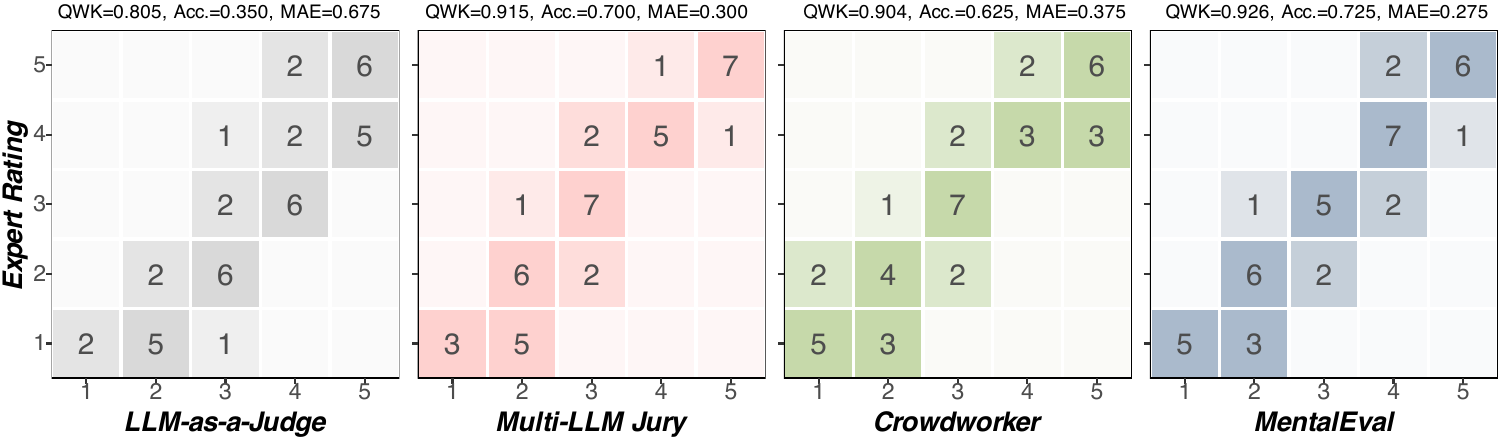}
  \vspace{-1.5em}
\caption{Agreement with expert ratings, where MentalEval shows the strongest alignment.}
\label{fig:agreement_heatmaps}
\end{figure}

\subsection{Does MentalEval Align with Expert Judgment?}
\label{MentalEval_test}
\vspace{-0.5em}

MentalEval is designed as a task-specific evaluator for MentalHospital, rather than a general-purpose LLM judge across arbitrary clinical tasks. Our goal is therefore to test whether it can reproduce expert judgments within the MentalHospital evaluation setting.To avoid evaluation leakage, we additionally sampled 40 psychiatric EHR cases from the coordinating medical center to construct the MentalEval test set. These cases were strictly excluded from all MentalEval training data. For each case, we sample model-generated responses and evaluate them across the five MentalEval dimensions, yielding 200 dimension-level test samples in total. Each sample is independently rated by three experts, and only unanimously assigned samples are retained as expert-validated references. We ablate the two-stage training procedure on this held-out set of unseen expert-labeled cases. As shown in Table~\ref{tab:mentaleval_ablation}, rubric-grounded SFT consistently improves all five evaluators, increasing average QWK from 0.759 to 0.879 and reducing MAE from 0.710 to 0.380. Expert-guided DPO further improves agreement, reaching an average QWK of 0.944, accuracy of 0.790, and MAE of 0.210. We then compare the trained MentalEval with LLM-as-a-Judge, Multi-LLM Jury, and individual crowdworker ratings in Figure~\ref{fig:agreement_heatmaps}. MentalEval achieves the strongest expert alignment, with the highest QWK (0.926) and accuracy (0.725), and the lowest MAE (0.275).

\subsection{Does MentalHospital Provide a Clinically Faithful Environment?}

\begin{wrapfigure}[20]{r}{0.55\linewidth}
  \centering
   \vspace{-1em}
  \includegraphics[width=\linewidth]{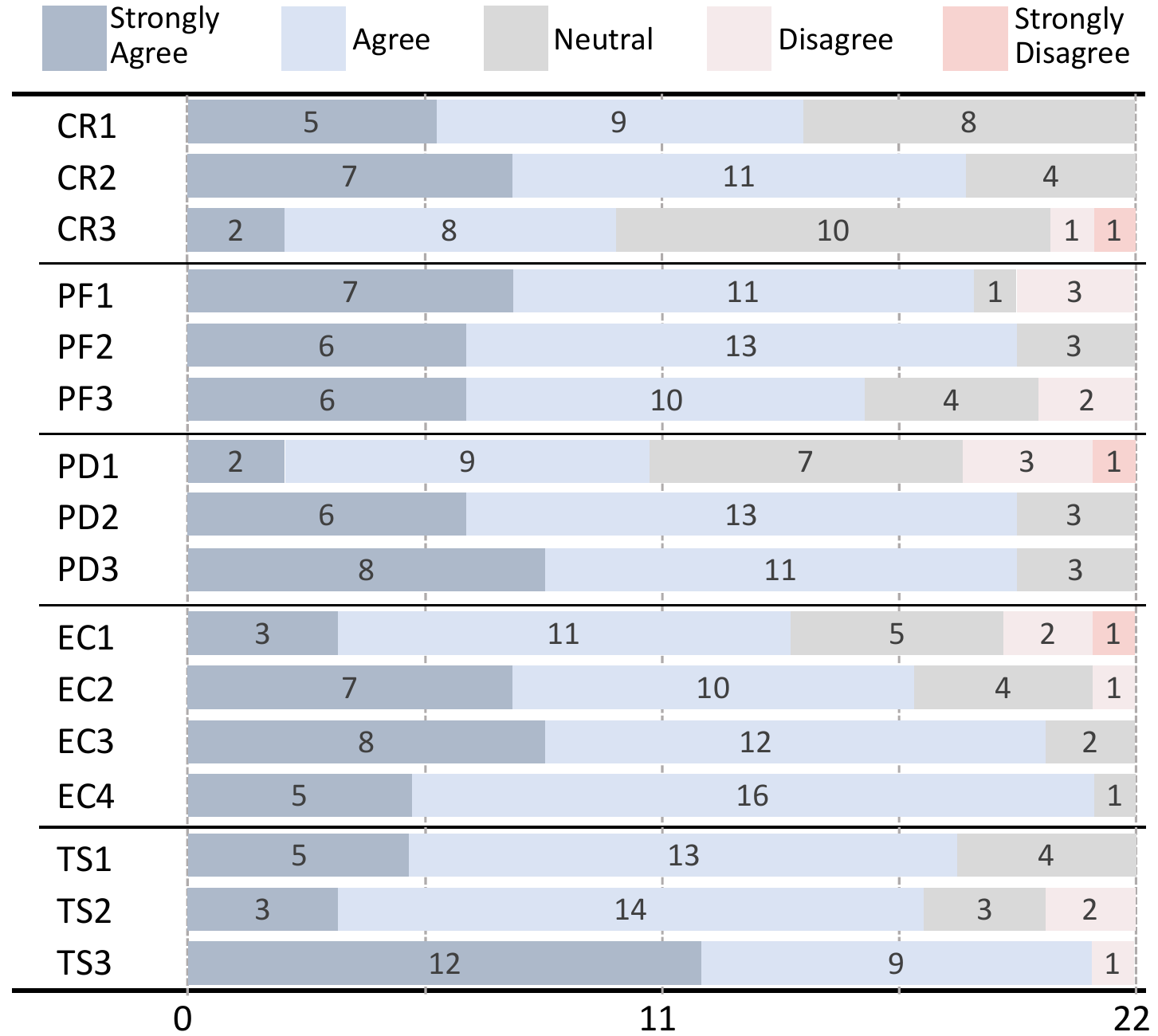}
   \caption{Clinical fidelity ratings for MentalHospital.}
  \label{fig:faithful_scale}
\end{wrapfigure}
We conduct a survey to assess the clinical faithfulness with 5 psychiatrists, 3 psychologists, and 14 medical trainees (full questionnaire is see Appendix~\ref{app:faithfulness_questionnaire}). The questionnaire covers five dimensions: clinical realism (CR), patient fidelity (PF), patient distinctiveness (PD), evaluation credibility (EC), and training suitability (TS), using a 5-point Likert scale. As shown in Figure~\ref{fig:faithful_scale}, MentalHospital receives consistently positive ratings, with an overall mean score of 3.88/5 (95\% bootstrap CI: [3.78, 3.98]), a positive response rate of 69.9\% (95\% CI: [64.8\%, 75.0\%]), and a negative response rate of only 5.1\% (95\% CI: [3.4\%, 6.8\%]). The ratings are particularly strong for training suitability and evaluation credibility, with positive response rates of 84.8\% and 75.0\%, suggesting that MentalHospital provides a clinically plausible environment.

\vspace{-0.5em}
\subsection{Qualitative Error Analysis}
\vspace{-0.5em}

\begin{wrapfigure}[20]{r}{0.55\linewidth}
  \centering
  \vspace{-2em}
  \includegraphics[width=\linewidth]{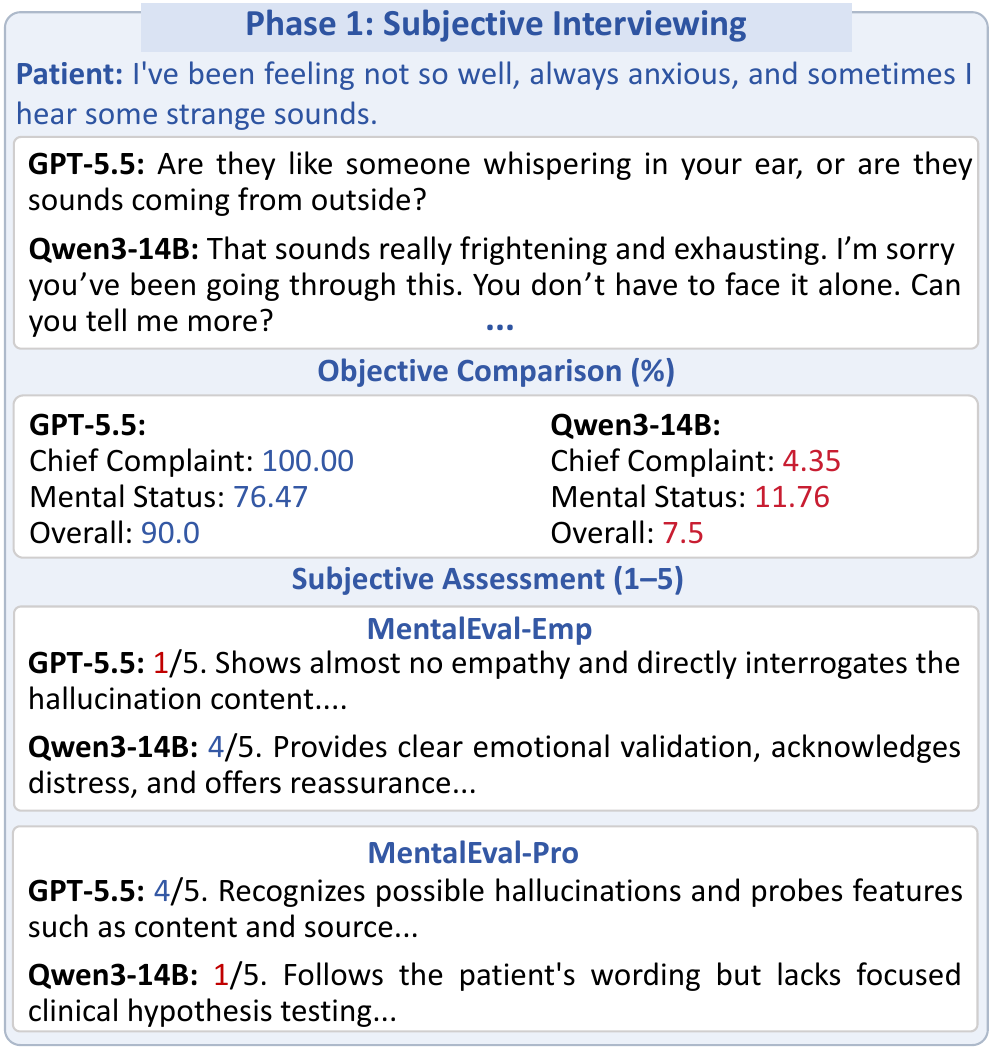}
  \vspace{-1.5em}
  \caption{Representative cases in MentalHospital.}
  \vspace{-1.5em}
  \label{fig:case_study}
\end{wrapfigure}

Figure~\ref{fig:case_study} illustrates two representative failure modes observed in MentalHospital. In a case involving anxiety and possible perceptual disturbance, \texttt{GPT-5.5} directly probes hallucination-related features such as source and content, yielding higher objective evidence coverage than \texttt{Qwen3-14B}. However, its response contains limited empathic acknowledgment, resulting in a lower MentalEval empathy score. Conversely, \texttt{Qwen3-14B} provides more explicit emotional validation and reassurance, but fails to pursue the hallucination hypothesis through targeted psychiatric questioning. This contrast shows that evidence-seeking behavior and process-level communication quality can diverge substantially. A model may recover key diagnostic evidence while offering poor affective support, whereas another may appear empathic but remain clinically under-informative. 


\vspace{-0.7em}
\section{Related Work}
\vspace{-0.5em}
\paragraph{Psychiatric Benchmarks.}
Early medical LLM evaluation mainly relied on static question-answering benchmarks, testing medical knowledge recall and exam-style clinical reasoning~\citep{singhal2025toward, tu2024towards}. This paradigm was later expanded to broader medical evaluation suites and real-world clinical benchmarks covering diverse tasks, safety, and clinical reliability~\citep{chen2024gmai, bedi2025medhelm, zhang2025llmeval, arora2025healthbench}, with subsequent work further examining clinical safety, expert consensus, and specialized reasoning~\citep{wang2025novel, mehandru2025er, yan2026livemedbench, zheng2026clinconsensus}. In parallel, mental-health benchmarks began to assess counseling, emotional support, and psychotherapy~\citep{li2025counselbench, xu2025mentalchat16k, zhang2025cbt}, while clinically oriented psychiatric benchmarks focused on psychiatric practice, disorder recognition, and diagnostic reasoning~\citep{liu2025psychbench, fouda2025psychiatrybench, sun2026mentalseek, song2026mentalbench}. These efforts connect subclinical mental-health support with threshold-level psychiatric assessment, but still largely evaluate isolated responses or task-specific outputs. Recent work has moved toward multi-turn dialogue and simulated clinical environments, testing contextual interaction across turns~\citep{zhang2024cpsycoun, racha2026omind, liu2025interactive, li2024mediq, yao2025dischargesim, pombal2025mindeval} and patient--doctor simulation~\citep{schmidgall2024agentclinic, fan2025ai, liu2026openhospital, sviridov20253mdbench, almansoori2025self,xu2026lingxidiagbench}. However, existing benchmarks still rarely support virtual psychiatric encounters.

\vspace{-1em}

\paragraph{Evaluation Methods.}
Early evaluation of LLM outputs mainly relied on task accuracy and surface-level automatic metrics, including lexical overlap and embedding-based similarity~\citep{papineni2002bleu,lin2004rouge,banerjee2005meteor,zhang2019bertscore}. These metrics remain common in benchmark-based assessment and text generation evaluation~\citep{singhal2023large, zhang2025llmeval}, but they are insufficient for clinically nuanced behaviors such as empathy, professionalism, diagnostic rigor, and treatment appropriateness. Later work therefore incorporated clinician-centered or expert-based protocols~\citep{kocaman2025clinical, arora2025healthbench}, with similar trends in clinical safety and reasoning benchmarks~\citep{wang2025novel, zhou2025automating}. Although expert review improves clinical validity, it is difficult to scale. Recent studies have explored model-based judging, including judge-style, jury-style, and rubric-driven frameworks~\citep{croxford2025evaluating, bedi2025medhelm, lee2025checkeval}, while also highlighting reliability concerns of LLM-based judges in medical settings~\citep{belmadani2026judges, williams2025human, gu2024survey}. Existing methods therefore still struggle to balance expertise and scalability, especially for subjective mental-health evaluation where process quality cannot be reduced to objective correctness alone.

\vspace{-0.7em}
\section{Conclusion}
\vspace{-0.5em}
We present MentalHospital, a virtual psychiatric evaluation environment for LLM-based medical AI. MentalHospital supports interactive S.O.A.P.-based psychiatric assessment with EHR-grounded virtual patients and a dual-track evaluation protocol that combines objective comparison with scalable subjective assessment. To operationalize specialist judgment, we develop MentalEval, a suite of five domain-specific evaluators trained through rubric-grounded SFT and expert-guided DPO. Experiments show that current LLMs remain limited in clinician-like psychiatric interaction, especially in mental status assessment, while MentalHospital and MentalEval provide faithful and scalable tools for evaluating in psychiatric care. Finally, clinician survey responses indicate positive acceptance of MentalHospital and its perceived value for psychiatric training.


\bibliographystyle{unsrtnat}
\bibliography{references}


\appendix

\input{appendix}


\end{document}

%% file: tables/benchmark_comparation.tex
\definecolor{background}{RGB}{235,235,235}

\newcommand{\symScale}{1}
\newcommand{\Yes}{\raisebox{0.15ex}{\scalebox{\symScale}{\textcolor{green!70!black}{\ding{51}}}}}
\newcommand{\No}{\raisebox{0.10ex}{\scalebox{\symScale}{\textcolor{red!70!black}{\ding{55}}}}}

\begin{table*}[t]
\caption{Comparison of psychiatric benchmarks. \textit{Inter.}, \textit{Exam.}, \textit{Diag.}, and \textit{Treat.} denote Interview, Examination, Diagnosis, and Treatment Planning, respectively. MentalHospital provides EHR-grounded full-process simulation, multi-perspective evaluation, and specialized evaluators.}
\label{tab:benchmark_comparison}
\centering
{\scriptsize
\resizebox{\textwidth}{!}{
\begin{tabular}{llclclcc}
\toprule
\multirow{2}{*}{\textbf{Benchmark}} &
\multicolumn{2}{c}{\textbf{Data}} &
\multicolumn{2}{c}{\textbf{Simulation}} &
\multicolumn{3}{c}{\textbf{Evaluation}} \\
\cmidrule(lr){2-3}
\cmidrule(lr){4-5}
\cmidrule(lr){6-8}
& \textbf{Data Source}
& \textbf{Psychiatry}
& \textbf{Scope}
& \textbf{Interaction}
& \textbf{Method}
& \textbf{Reference}
& \textbf{Evaluator} \\
\midrule

HealthBench~\citep{arora2025healthbench}
& Physician-authored
& \No
& General Healthcare
& \No
& Rubric
& Physician
& \No \\

MedDialogRubrics~\citep{gong2026meddialogrubrics}
& Synthetic
& \No
& Consultation / Diagnosis
& \No
& Rubric
& \No
& \No \\

CPsyCoun~\citep{zhang2024cpsycoun}
& Web
& \Yes
& Consultation
& \No
& Rubric / Metrics
& \No
& \No \\

LingxiDiagBench~\citep{xu2026lingxidiagbench}
& Synthetic
& \Yes
& Consultation / Diagnosis
& \No
& Labels
& \No
& \No \\

PSYCHE~\citep{lee2025psyche}
& Synthetic
& \Yes
& Consultation / Diagnosis
& \Yes
& Metrics
& \No
& \No \\

AI Hospital~\citep{fan2025ai}
& Web
& \No
& Consultation / Diagnosis
& \Yes
& Metrics
& \No
& \No \\

OpenHospital~\citep{liu2026openhospital}
& Synthetic
& \No
& Interview / Diagnosis
& \Yes
& Metrics
& \No
& \No \\

AgentClinic~\citep{schmidgall2024agentclinic} 
& Single Hospital
& \No
& Inter. / Exam. / Diag.
& \Yes
& Metrics / Scale
& EHR
& \No \\

\midrule
\rowcolor{background}
\textbf{MentalHospital (Ours)}
& \textbf{Clinical Centers}
& \Yes
& \textbf{Inter. / Exam. / Diag. / Treat.}
& \Yes
& \textbf{Rub. / Metr. / Lab. / Scale}
& \textbf{EHR / Psychiatrist}
& \textbf{MentalEval} \\

\bottomrule
\end{tabular}
}
}
\vspace{-1.3em}
\end{table*}

%% file: tables/experiments_main_comparasion.tex
\begin{table*}[t]
\caption{Representative objective and subjective evaluation results across S.O.A.P. stages. Objective metrics are reported as percentages (\%), while subjective scores are reported on a 1--5 scale.}
\vspace{-0.5em}
\centering
\label{tab:main_results}

\small
\setlength{\tabcolsep}{6pt}
\renewcommand{\arraystretch}{1.15}
\setlength{\cmidrulewidth}{0.75pt}
\definecolor{HeaderGray}{RGB}{205,205,205}
\definecolor{GroupGray}{RGB}{222,222,222}
\definecolor{RowGray}{RGB}{235,235,235}

\newcommand{\grouprow}[1]{%
\rowcolor{GroupGray}
\multicolumn{11}{l}{\rule[-0.7ex]{0pt}{3.2ex}\textbf{#1}} \\
}

\resizebox{\textwidth}{!}{
\begin{tabular}{lrrrrrrrrrr}
\toprule
\textbf{Model}
& \multicolumn{5}{l}{\textbf{Objective Comparison (\%)}}
& \multicolumn{5}{l}{\textbf{Subjective Assessment (1--5)}} \\
\cmidrule(lr){2-6}
\cmidrule(lr){7-11}
& \textbf{Inter.}
& \textbf{Exam.}
& \textbf{Note}
& \textbf{Cate.}
& \textbf{Diso.}
& \textbf{Emp.}
& \textbf{Prof.}
& \textbf{Qual.}
& \textbf{Rigor}
& \textbf{App.} \\
\midrule

\rowcolor{RowGray}
Medical Trainees
& 78.23 & 71.05 & 82.91 & 81.42 & 72.58
& 1.63 & 3.86 & 3.87 & 2.33 & 2.47 \\

Human Experts
& 82.35 & 91.90 & 88.67 & 90.59 & 83.24
& 1.22 & 4.25 & 4.01 & 2.54 & 2.96 \\

\midrule
\midrule

\grouprow{Small\hspace{0.5em}Language\hspace{0.5em}Models}

Llama-3-8B
& 19.06 & 7.94 & 9.96 & 18.11 & 6.08
& 2.05 & 2.88 & 1.93 & 1.86 & 2.67 \\

\rowcolor{RowGray}
Qwen3-14B
& 23.19 & 8.33 & 12.12 & 23.55 & 15.19
& 1.96 & 2.90 & 2.17 & 2.20 & 2.83 \\

GPT-OSS-20B
& 27.43 & 8.60 & \underline{33.12} & 37.21 & 25.81
& 1.60 & 3.15 & \underline{3.14} & 2.06 & 3.23 \\

\rowcolor{RowGray}
QWQ-32B
& \underline{30.40} & \underline{10.87} & 32.30 & \underline{44.53} & \underline{26.39}
& 2.13 & \underline{3.72} & 2.85 & \underline{3.14} & \underline{3.40} \\

\midrule

\grouprow{Large\hspace{0.5em}Language\hspace{0.5em}Models}

GPT-OSS-120B
& 36.72 & 9.92 & 39.94 & 49.54 & 34.19
& 1.50 & 3.45 & \underline{3.89} & \underline{2.42} & 3.35 \\

\rowcolor{RowGray}
Claude-Sonnet-4.6
& 30.44 & 21.28 & 34.81 & 50.28 & 34.57
& 1.72 & 3.39 & 2.70 & 2.22 & 3.17 \\

Deepseek-v4-Pro
& 30.95 & 21.93 & \underline{56.46} & 26.20 & 20.85
& \underline{2.09} & 3.48 & 2.91 & 2.39 & 3.29 \\

\rowcolor{RowGray}
GPT-5.5
& \underline{69.06} & \underline{22.99} & 45.44 & \underline{70.23} & \underline{49.85}
& 1.50 & \underline{3.63} & 3.66 & 2.21 & \underline{3.48} \\

\midrule

\grouprow{Medical-specific\hspace{0.5em}Models}

MedGemma-4B-it
& 25.65 & 9.42 & 10.67 & 21.48 & 3.05
& \underline{2.16} & 2.57 & 1.99 & 2.15 & 2.91 \\
\rowcolor{RowGray}
MedGemma-27B-it
& 43.12 & 16.05 & 35.91 & 32.27 & 10.10
& 2.05 & 2.83 & 2.67 & 2.33 & 3.17 \\


Baichuan-M2
& \underline{51.29} & \underline{24.76} & \underline{50.54} & \underline{80.56} & \underline{43.19}
& 1.29 & \underline{2.85} & \underline{2.91} & \underline{2.44} & \underline{3.33} \\

\bottomrule
\end{tabular}
}
\vspace{-1.5em}
\end{table*}

%% file: tables/experiments_comparasion_coverage.tex
\definecolor{headergray}{HTML}{CFCFCF}
\definecolor{rowgray}{HTML}{EFEFEF}

\begin{wraptable}[10]{r}{0.55\linewidth}
  \centering
  \vspace{-1.2em}
  \caption{Coverage by clinical component.}
  \label{tab:interview_note}
  \scriptsize
  \setlength{\tabcolsep}{2.8pt}
  \renewcommand{\arraystretch}{1.12}
  \setlength{\cmidrulewidth}{0.75pt}

  \resizebox{\linewidth}{!}{
  \begin{tabular}{lrrrrrr}
  \toprule
  \textbf{Model}
  & \multicolumn{3}{l}{\textbf{Interview Phase}}
  & \multicolumn{3}{l}{\textbf{Note Phase}} \\
  \cmidrule(lr){2-4}
  \cmidrule(lr){5-7}
  & \textbf{Avg.}
  & \textbf{CC}
  & \textbf{MS}
  & \textbf{Avg.}
  & \textbf{CC}
  & \textbf{MS} \\
  \midrule

  Qwen3-14B
  & 23.19 & 44.36 & 12.57
  & 12.12 & 16.38 & 17.25 \\

  \rowcolor{rowgray}
  MedGemma-4B-it
  & 25.65 & 40.69 & 10.45
  & 10.67 & 9.68 & 6.66 \\

  QWQ-32B
  & 30.40 & 50.90 & 17.84
  & 32.30 & 42.91 & 34.22 \\

  \rowcolor{rowgray}
  Deepseek-v4-Pro
  & 30.95 & 41.67 & 23.13
  & 56.46 & 59.94 & 43.91 \\

  GPT-5.5
  & 69.06 & 85.38 & 55.72
  & 45.44 & 55.34 & 65.89 \\

  \bottomrule
  \end{tabular}
  }
  \vspace{-0.8em}
\end{wraptable}

%% file: tables/experiments_patient_ablation.tex
\definecolor{RowGray}{RGB}{235,235,235}

\begin{table}[t]
\caption{Ablation of skill-augmented patient construction for standardized patients.}
\label{tab:patient_ablation}
\centering
\small
\setlength{\tabcolsep}{5pt}
\newcommand{\chg}[1]{\raisebox{-0.45ex}{\tiny\,$#1$}}

\renewcommand{\arraystretch}{1.15}
\begin{tabular}{lllllll}
\toprule
\textbf{Patient Variant}
& \multicolumn{3}{l}{\textbf{Objective Comparison (\%)}}
& \multicolumn{3}{l}{\textbf{Subjective Assessment (1–5)}} \\
\cmidrule(lr){2-4}
\cmidrule(lr){5-7}
& \textbf{Cov.$\uparrow$}
& \textbf{Prec.$\uparrow$}
& \textbf{Halluc.$\downarrow$}
& \textbf{Realism$\uparrow$}
& \textbf{Fidelity$\uparrow$}
& \textbf{Distinct.$\uparrow$} \\
\midrule

Naive Prompting
& 72.40 & 86.10 & 7.80
& 2.12 & 2.23 & 1.92 \\

\rowcolor{RowGray}
Representation Skill Only
& 72.40\chg{+0.00} & 85.72\chg{-0.38} & 9.04\chg{+1.24}
& 4.12\chg{+2.00} & 3.82\chg{+1.59} & 3.72\chg{+1.80} \\

Memory Skill Only
& 84.83\chg{+12.43} & 91.64\chg{+5.92} & 4.52\chg{-4.52}
& 3.55\chg{-0.57} & 3.87\chg{+0.05} & 2.77\chg{-0.95} \\

\rowcolor{RowGray}
Full Patient
& 88.91\chg{+4.08} & 93.27\chg{+1.63} & 3.08\chg{-1.44}
& 4.17\chg{+0.62} & 4.05\chg{+0.18} & 3.87\chg{+1.10} \\

\bottomrule
\end{tabular}
\vspace{-1.5em}
\end{table}

%% file: tables/experiments_mentaleval_ablation.tex
\newcommand{\chg}[1]{\raisebox{-0.45ex}{\tiny\,$#1$}}

\definecolor{HeaderGray}{RGB}{205,205,205}
\definecolor{RowGray}{RGB}{235,235,235}

\begin{table}[t]
\caption{Ablation results of MentalEval training against expert ratings. DPO is applied after SFT.}
\centering
\label{tab:mentaleval_ablation}

\scriptsize
\renewcommand{\arraystretch}{1.15}
\setlength{\cmidrulewidth}{0.75pt}

\resizebox{\linewidth}{!}{
\begin{tabular}{lccccccccc}
\toprule
\textbf{Evaluator}
& \multicolumn{3}{c}{\textbf{Base Model (Qwen3-8B)}}
& \multicolumn{3}{c}{\textbf{+SFT}}
& \multicolumn{3}{c}{\textbf{+DPO}} \\
\cmidrule(lr){2-4}
\cmidrule(lr){5-7}
\cmidrule(lr){8-10}
& \textbf{QWK$\uparrow$} & \textbf{Acc.$\uparrow$} & \textbf{MAE$\downarrow$}
& \textbf{QWK$\uparrow$} & \textbf{Acc.$\uparrow$} & \textbf{MAE$\downarrow$}
& \textbf{QWK$\uparrow$} & \textbf{Acc.$\uparrow$} & \textbf{MAE$\downarrow$} \\
\midrule

MentalEval-Emp
& 0.677 & 0.225 & 0.875
& 0.824\chg{+0.148} & 0.500\chg{+0.275} & 0.525\chg{-0.350}
& 0.926\chg{+0.102} & 0.725\chg{+0.225} & 0.275\chg{-0.250} \\

\rowcolor{RowGray}
MentalEval-Pro
& 0.705 & 0.275 & 0.825
& 0.844\chg{+0.139} & 0.550\chg{+0.275} & 0.475\chg{-0.350}
& 0.941\chg{+0.097} & 0.775\chg{+0.225} & 0.225\chg{-0.250} \\

MentalEval-Note
& 0.766 & 0.400 & 0.675
& 0.910\chg{+0.144} & 0.700\chg{+0.300} & 0.300\chg{-0.375}
& 0.940\chg{+0.030} & 0.775\chg{+0.075} & 0.225\chg{-0.075} \\

\rowcolor{RowGray}
MentalEval-Diag 
& 0.828 & 0.475 & 0.575
& 0.887\chg{+0.059} & 0.725\chg{+0.250} & 0.325\chg{-0.250}
& 0.967\chg{+0.080} & 0.875\chg{+0.150} & 0.125\chg{-0.200} \\

MentalEval-Treat
& 0.819 & 0.425 & 0.600
& 0.928\chg{+0.109} & 0.725\chg{+0.300} & 0.275\chg{-0.325}
& 0.948\chg{+0.020} & 0.800\chg{+0.075} & 0.200\chg{-0.075} \\
\midrule
\textbf{Average}
& 0.759 & 0.360 & 0.710
& 0.879\chg{+0.120} & 0.640\chg{+0.280} & 0.380\chg{-0.330}
& 0.944\chg{+0.066} & 0.790\chg{+0.150} & 0.210\chg{-0.170} \\

\bottomrule
\end{tabular}
}
\end{table}

%% file: appendix.tex
\section{Limitations}
MentalHospital currently focuses on text-based psychiatric interaction. Although we explored generative digital humans for richer patient presentation, current techniques remain insufficient for producing pathology-level, clinically discriminative facial expressions, prosody, and subtle behaviors required for psychiatric assessment. We therefore use text-based simulation to preserve controllability, clinical grounding, and evaluation reliability. In addition, MentalHospital does not yet provide systematic fairness, bias, or adversarial safety evaluation for scenarios such as self-harm escalation, psychosis reinforcement, medication misuse, or demographic bias. Future work will extend the benchmark toward clinically faithful multimodal simulation, safety-oriented stress tests, and broader validation across clinical settings.

\section{Ethics Statement}
\label{app:ethics_statement}

\section{Resource Release and Access Protocol}
\label{app:resource_release}

To support reproducibility while protecting clinical privacy, we adopt a tiered release protocol. Publicly released resources will include the MentalHospital environment code, evaluation scripts, prompts, rubrics, metric implementations, model-output logs, and the MentalEval evaluator weights with corresponding inference scripts. These resources are sufficient to reproduce the evaluation pipeline and apply MentalEval to new model outputs.

Raw EHRs, identifiable clinical records, case-level clinical narratives, and MentalEval training data will not be publicly released due to clinical privacy constraints. De-identified benchmark cases and derived evaluation checkpoints will be provided only through controlled research access, subject to data-use approval. This protocol separates reproducible evaluation components from sensitive clinical source data and reduces the risk of privacy leakage.

\section{Safety, Bias, and Generalizability.}
MentalHospital is designed for controlled evaluation rather than direct clinical deployment. Although the benchmark covers multi-center psychiatric EHR cases across three regions, its standardized patient behaviors, clinical documentation patterns, and evaluator judgments may still reflect local practice conventions and expert preferences. In addition, EHR-derived references may encode historical diagnostic or examination biases. MentalEval assesses process-level quality, empathy, diagnostic rigor, and treatment appropriateness, but it does not fully cover all safety-critical scenarios such as self-harm escalation, psychosis reinforcement, medication misuse, or demographic bias. Future work will extend MentalHospital with adversarial safety cases, fairness-oriented evaluation, and culture-sensitive patient simulation.

\section{Data Distribution}
\label{app:data_distribution}

In this section, we provide the full forms and brief explanations of the category abbreviations used in the main text. Fig~\ref{Disorders_List} presents the names and corresponding codes for the 76 specific disorders included in the study.

The full form of \textbf{ANX} is Anxiety or fear-related disorders. Anxiety or fear-related disorders are a group of mental health conditions characterized by excessive fear and anxiety, along with associated behavioral disturbances. These symptoms are severe enough to cause significant distress or impairment in an individual's personal, social, educational, occupational, or other important areas of functioning. According to the ICD-11, specific types of disorders under this category include generalized anxiety disorder, panic disorder, agoraphobia, specific phobias, social anxiety disorder, and separation anxiety disorder. Patients may exhibit specific cognitive features, such as excessive worry about particular situations or objects, which help distinguish between different types of anxiety or fear-related disorders. 

The full form of \textbf{CATA} is Catatonia. Catatonia is a clinical syndrome characterized by psychomotor disturbances, which manifest as a combination of reduced, increased, or abnormal psychomotor activity. Typical symptoms include stupor (maintaining a fixed posture for extended periods), rigidity (abnormal increase in muscle tension), waxy flexibility (limbs can be manipulated and maintained in a posture), mutism, negativism (active or passive resistance to instructions), echolalia (repetition of others' speech), or echopraxia (repetition of others' actions). This syndrome may arise secondary to psychiatric disorders such as schizophrenia, bipolar disorder, and depressive disorders, and can also be triggered by drug reactions, neurological diseases, or physical illnesses.

\begin{figure}[t]
  \includegraphics[width=\textwidth]{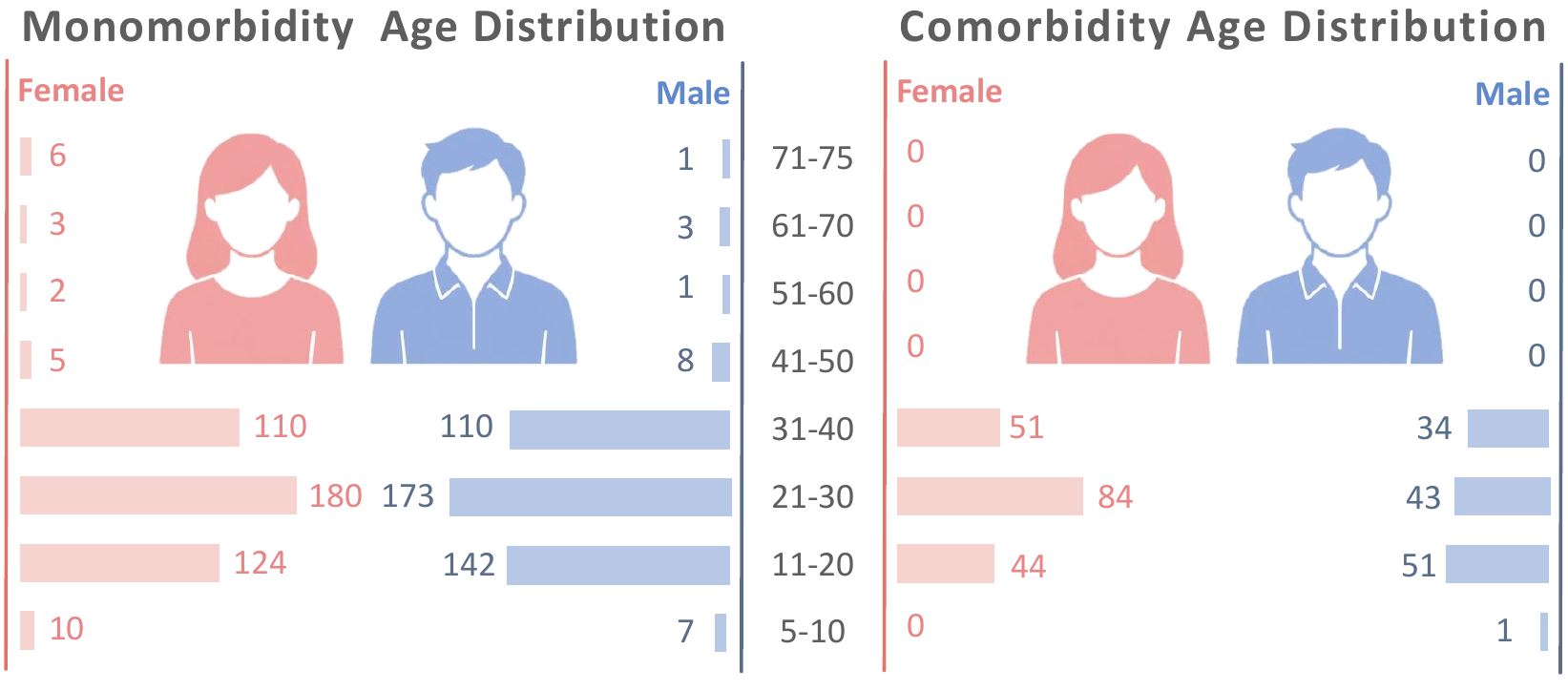}
\caption{Age distribution in MentalHospital. Left: single-diagnosis cases; right: comorbid cases.}
\label{fig:age_distribution}
\end{figure}

The full form of \textbf{SUD} is Disorders due to substance use or addictive behaviors. These disorders refer to functional impairment or distress caused by the repeated use of psychoactive substances (including drugs) or specific repetitive rewarding behaviors, manifesting in cognitive, behavioral, and physiological symptoms. The core features include a persistent craving for the substance or behavior, impaired control, prioritizing use despite harm, and neuroadaptive changes. Key characteristics also include difficulty controlling the frequency, duration, and intensity of use, withdrawal symptoms after reducing or stopping use (such as tremors, anxiety, insomnia), tolerance (requiring increased doses or intensity for the same effect), functional impairment in personal, family, occupational, or social roles, and continued use despite knowing the harm to physical and mental health.

The full form of \textbf{BOD} is Disorders of bodily distress or bodily experience. These disorders are mental health conditions primarily characterized by bodily symptoms. Patients often show excessive worry about their health, accompanied by persistent physical symptoms such as pain, fatigue, and digestive issues, among others. These symptoms typically cannot be fully explained by routine medical examinations.

The full form of \textbf{STRESS} is Disorders specifically associated with stress. These disorders are a group of mental health conditions triggered by clearly identifiable stressors or traumatic events. Patients typically present with acute stress disorder, which occurs within days of exposure to extreme stressors and is characterized by intense fear, helplessness, and flashbacks. Post-traumatic stress disorder (PTSD) lasts longer and is marked by the recurrent re-experiencing of traumatic events, avoidance behaviors, and heightened arousal. Adjustment disorder is another form, where emotional or behavioral problems arise from identifiable life events that exceed normal coping mechanisms, typically occurring within one month of the event and affecting daily functioning.

The full form of \textbf{DISR} is Disruptive behavior or dissocial disorders. Impulse control disorders are a group of mental health conditions characterized by the difficulty of patients to resist strong, inappropriate desires or impulses, leading to the performance of certain specific behaviors. These behaviors often cause long-term harm to the individual or others, or result in significant impairment in important functional areas such as social, family, or occupational domains. Although patients typically recognize the negative impact of these behaviors, they are unable to resist the impulse in the short term. Common ICDs include pyromania, kleptomania, intermittent explosive disorder, and compulsive sexual behavior disorder.

The full form of \textbf{DISS} is Dissociative disorders. Dissociative disorders are complex mental health conditions primarily characterized by the involuntary partial or complete disintegration of psychological functions such as identity, memory, consciousness, perception, or behavior. These symptoms are not directly caused by drugs, substance abuse, or other psychiatric, behavioral, or neurodevelopmental disorders, and must be distinguished from normal manifestations in cultural or religious practices. The main types include, but are not limited to, dissociative amnesia (selective memory loss), depersonalization-derealization disorder (feeling detached from reality), dissociative fugue (manifesting as a confused state), and dissociative identity disorder (formerly known as multiple personality disorder).

The full form of \textbf{ELIM} is Elimination disorders. Elimination disorders refer to the persistent or recurrent inability to control urination or defecation after reaching the normal developmental age. The main manifestations include enuresis, which is the involuntary urination during the night or day, typically in bed or clothing, and encopresis, which is the inappropriate defecation in places such as clothing. This diagnosis is applicable when children fail to achieve typical bladder and bowel control by the expected developmental age (enuresis by age 5 and encopresis by age 4). If the incontinence symptoms can be fully attributed to other health conditions, congenital or acquired abnormalities in the urinary or gastrointestinal systems, or medication side effects, it should not be diagnosed as an elimination disorder.

The full form of \textbf{EAT} is Feeding or eating disorders. Feeding or eating disorders are a group of mental health conditions characterized by abnormal eating behaviors that cannot be explained by other health conditions and do not conform to normal developmental patterns in the cultural context. These include, but are not limited to, rumination/regurgitation, avoidant/restrictive food intake, anorexia nervosa, bulimia, binge eating, and pica. These disorders often involve issues with controlling food intake, concerns about weight and body shape, or inappropriate dietary choices.

The full form of \textbf{PREG} is Mental or behavioral disorders associated with pregnancy, childbirth, or the puerperium. These disorders refer to mental health issues that occur during pregnancy, during childbirth, and within the first six weeks postpartum. Such disorders include, but are not limited to, depressive disorders, anxiety disorders, and psychotic disorders. Based on their severity and specific manifestations, they can be further categorized into non-psychotic disorders (e.g., postpartum depression) and psychotic disorders (e.g., postpartum psychosis). These mental health conditions significantly affect a woman's psychological state and social functioning, requiring professional medical intervention.

\begin{figure}[t]
  \includegraphics[width=\textwidth]{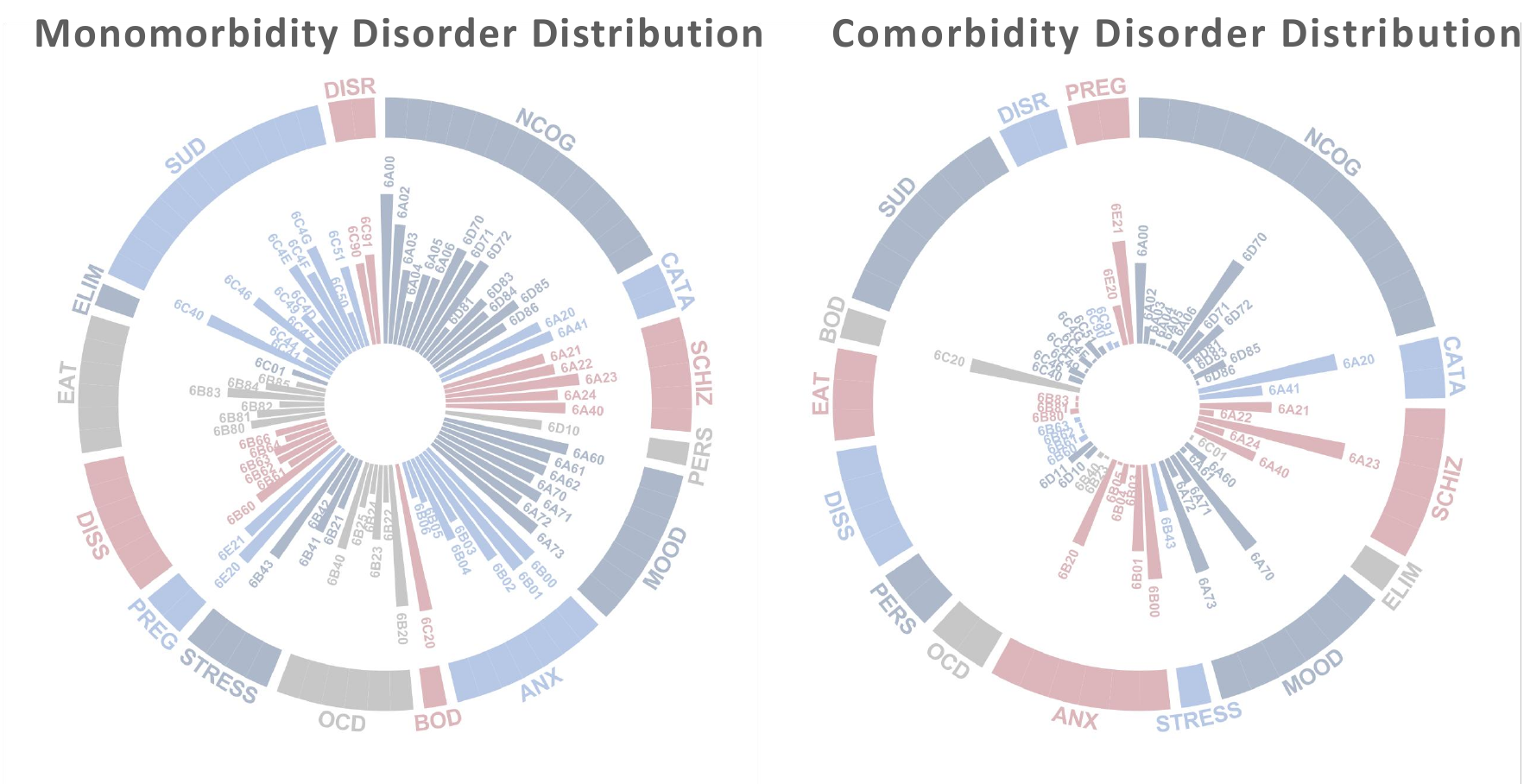}
\caption{Disorder distribution. Left: single-diagnosis cases; right: comorbid cases.}
\label{fig:disorder_distribution}
\end{figure}

The full form of \textbf{MOOD} is Mood disorders. Mood disorders are a group of mental health conditions characterized by significant and persistent changes in mood or emotional states. These disorders include two main types: bipolar disorder and depressive disorder. They are typically manifested by several major types of mood episodes: depressive episodes, which involve persistent low mood, loss of interest, and reduced energy; manic episodes, which involve abnormally elevated mood, increased activity, and reduced need for sleep; mixed episodes, where both depressive and manic symptoms are present simultaneously; and hypomanic episodes, which are similar to manic episodes but less severe, with minimal impairment of social functioning. These mood episodes are not independent diagnostic entities but are the primary components of mood disorders.

The full form of \textbf{NCOG} is Neurocognitive disorders. Neurocognitive disorders refer to major clinical impairments in cognitive function that are acquired, rather than developmental or congenital. Specifically, these disorders do not include cognitive issues that are present from birth, nor do they cover problems that typically emerge during the developmental stage (which are classified as neurodevelopmental disorders). Instead, neurocognitive disorders refer to a decline in cognitive function from a previously attained level.

The full form of \textbf{NDEV} is Neurodevelopmental disorders. Neurodevelopmental disorders are a group of mental or behavioral disorders that appear during an individual's developmental stage, typically in early childhood, especially before school age, and are characterized by developmental deficits that cause impairment in social, academic, or occupational functioning. The core features of these disorders include significant difficulties in acquiring and performing specific intellectual, motor, language, or social functions. While many mental and behavioral disorders can emerge during the developmental period, only those disorders that are primarily characterized by neurodevelopmental impairments are classified under this group.

The full form of \textbf{OCD} is Obsessive-compulsive or related disorders. Obsessive-compulsive or related disorders are a group of mental health conditions characterized by repetitive thoughts and behaviors. The main symptoms include: obsessive thoughts, which are unwanted, recurring ideas, doubts, or impulses, and compulsive behaviors, which are repetitive actions or mental rituals. These symptoms persist despite the patient's awareness of their meaninglessness, leading to significant personal distress and impairment in social functioning.

The full form of \textbf{PERS} is Personality disorders and related traits. Personality disorders refer to a persistent and stable pattern of emotional experiences, cognitive patterns, and behaviors that significantly deviate from the cultural expectations of the individual's background, leading to widespread functional impairment or subjective distress. This pattern typically emerges during adolescence or early adulthood, with its stability confirmed through observation in adulthood. Personality disorders and related personality traits cover two main aspects: prominent personality features or patterns, where individuals exhibit significant deviations in personality traits, but the severity or extent may not fully meet the diagnostic criteria for a personality disorder, and personality disorders, which include various subtypes (such as borderline, antisocial, schizoid, etc.), characterized by persistent functional impairment, and must exclude direct consequences of other psychiatric disorders or substance use.

The full form of \textbf{SCHIZ} is Schizophrenia or other primary psychotic disorders. Schizophrenia and other primary psychotic disorders are a group of mental health conditions characterized by significant impairment in reality testing. The core symptoms include: positive symptoms such as delusions, hallucinations, and disorganized speech; negative symptoms such as flat affect, reduced volition, and social withdrawal; and cognitive dysfunctions such as impairments in attention, memory, and executive function. These symptoms cause the individual's thoughts, emotions, and behaviors to deviate markedly from the cultural norms, and cannot be attributed to other psychiatric disorders, such as bipolar disorder or substance use disorders.

\begin{figure}[tb]
\centering
\begin{tcolorbox}[
    colframe=black!60,       
    colback=black!5,         
    coltitle=white,
    fonttitle=\bfseries,
    title=\small Specific Disorders List,
    sharp corners,
    boxrule=0.4mm,
    boxsep=2pt,
    left=2pt, right=2pt,
    top=2pt, bottom=2pt,
    before upper={\setlength{\parskip}{0pt}} 
]
\small
["6A20 Schizophrenia","6B00 Generalised anxiety disorder","6A72 Dysthymic disorder","6E20 Mental or behavioural disorders associated with pregnancy, childbirth or the puerperium, without psychotic symptoms","6A61 Bipolar type II disorder","6A00 Disorders of intellectual development","6D70 Delirium","6E21 Mental or behavioural disorders associated with pregnancy, childbirth or the puerperium, with psychotic symptoms","6B20 Obsessive-compulsive disorder","6A40 Catatonia associated with another mental disorder","6D72 Amnestic disorder","6A62 Cyclothymic disorder","6A23 Acute and transient psychotic disorder","6B43 Adjustment disorder","6A21 Schizoaffective disorder","6A70 Single episode depressive disorder","6A73 Mixed depressive and anxiety disorder","6A71 Recurrent depressive disorder","6A24 Delusional disorder","6A60 Bipolar type I disorder","6C20 Bodily distress disorder","6A22 Schizotypal disorder","6A41 Catatonia induced by substances or medications","6D71 Mild neurocognitive disorder","6D85 Dementia due to diseases classified elsewhere","6C4G Disorders due to use of unknown psychoactive substances","6C4E Disorders due to use of other specified psychoactive substances, including medications","6C46 Disorders due to use of stimulants including amphetamines, methamphetamine or methcathinone","6B04 Social anxiety disorder","6C40 Disorders due to use of alcohol","6D10 Personality disorder","6A02 Autism spectrum disorder","6B02 Agoraphobia","6B60 Dissociative neurological symptom disorder","6C91 Conduct-dissocial disorder","6C90 Oppositional defiant disorder","6B83 Avoidant-restrictive food intake disorder","6B40 Post traumatic stress disorder","6C4F Disorders due to use of multiple specified psychoactive substances, including medications","6D84 Dementia due to psychoactive substances including medications","6B63 Possession trance disorder","6B03 Specific phobia","6B41 Complex post traumatic stress disorder","6A03 Developmental learning disorder","6C51 Gaming disorder","6D86 Behavioural or psychological disturbances in dementia","6A06 Stereotyped movement disorder","6D83 Frontotemporal dementia","6B80 Anorexia Nervosa","6B23 Hypochondriasis","6C49 Disorders due to use of hallucinogens","6B81 Bulimia Nervosa","6A05 Attention deficit hyperactivity disorder","6B62 Trance disorder","6B84 Pica","6B21 Body dysmorphic disorder","6B05 Separation anxiety disorder","6B64 Dissociative identity disorder","6B61 Dissociative amnesia","6A04 Developmental motor coordination disorder","6C4D Disorders due to use of dissociative drugs including ketamine and phencyclidine [PCP]","6B66 Depersonalization-derealization disorder","6B25 Body-focused repetitive behaviour disorders","6B42 Prolonged grief disorder","6B82 Binge eating disorder","6C44 Disorders due to use of sedatives, hypnotics or anxiolytics","6B22 Olfactory reference disorder","6C01 Encopresis","6D81 Dementia due to cerebrovascular disease","6B06 Selective mutism","6C50 Gambling disorder","6B24 Hoarding disorder","6C41 Disorders due to use of cannabis","6B85 Rumination-regurgitation disorder","6C47 Disorders due to use of synthetic cathinones"]
\normalsize
\end{tcolorbox}

\captionsetup{aboveskip=4pt, belowskip=0pt, width=\linewidth}
\caption{\textbf{Specific Disorders List.} A list of 76 disorders mentioned in the paper, including their corresponding codes and specific names.}
\label{Disorders_List}
\end{figure}

\section{Data Processing and Ethical Compliance}
\label{app:data_processing}

\paragraph{Data Source and Case Composition.}
\label{data_source_and_case_composistion}
MentalHospital is constructed from 1,193 de-identified psychiatric electronic health record (EHR) cases collected from multiple collaborating clinical centers. The dataset contains 885 single-diagnosis cases and 308 comorbid cases, covering both typical psychiatric presentations and clinically complex comorbidity patterns. The age distribution of single-diagnosis and comorbid cases is shown in Figure~\ref{fig:age_distribution}, and the corresponding disorder-level distribution is summarized in Figure~\ref{fig:disorder_distribution}. For each case $c$, two licensed psychiatrists independently reviewed the EHR to ensure that the record was clinically complete and that the final diagnosis was consistent with the ICD-11 classification of mental, behavioral, and neurodevelopmental disorders. Cases with incomplete clinical information, ambiguous diagnostic conclusions, or unresolved annotation conflicts were excluded from benchmark construction. Disagreements between the two reviewers were resolved through expert consensus.

\begin{figure}[tb]
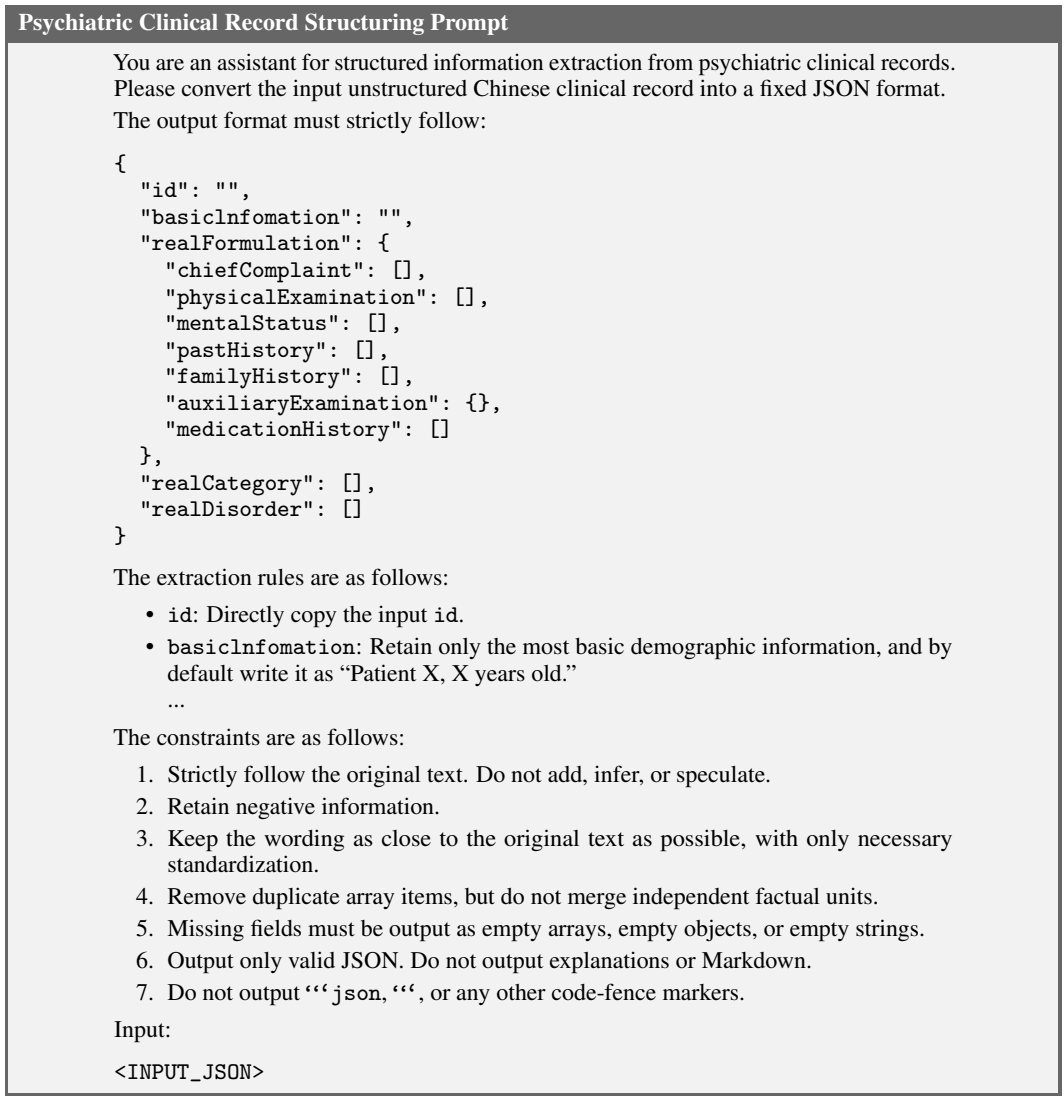

\centering
\begin{tcolorbox}[
    colframe=black!60,       
    colback=black!5,         
    coltitle=white,
    fonttitle=\bfseries,
    title=\small Psychiatric Clinical Record Structuring Prompt, 
    sharp corners,
    boxrule=0.4mm,
    boxsep=2pt,
    left=2pt, right=2pt,
    top=2pt, bottom=2pt,
    before upper={\setlength{\parskip}{0pt}} 
]
\small

\begin{quote}
You are an assistant for structured information extraction from psychiatric clinical records. Please convert the input unstructured Chinese clinical record into a fixed JSON format.

The output format must strictly follow:
\begin{verbatim}
{
  "id": "",
  "basiclnfomation": "",
  "realFormulation": {
    "chiefComplaint": [],
    "physicalExamination": [],
    "mentalStatus": [],
    "pastHistory": [],
    "familyHistory": [],
    "auxiliaryExamination": {},
    "medicationHistory": []
  },
  "realCategory": [],
  "realDisorder": []
}
\end{verbatim}

The extraction rules are as follows:
\begin{itemize}
    \item \texttt{id}: Directly copy the input \texttt{id}.
    \item \texttt{basiclnfomation}: Retain only the most basic demographic information, and by default write it as ``Patient X, X years old.''
    \\ ...
\end{itemize}

The constraints are as follows:
\begin{enumerate}
    \item Strictly follow the original text. Do not add, infer, or speculate.
    \item Retain negative information.
    \item Keep the wording as close to the original text as possible, with only necessary standardization.
    \item Remove duplicate array items, but do not merge independent factual units.
    \item Missing fields must be output as empty arrays, empty objects, or empty strings.
    \item Output only valid JSON. Do not output explanations or Markdown.
    \item Do not output \texttt{```json}, \texttt{```}, or any other code-fence markers.
\end{enumerate}

Input:
\begin{verbatim}
<INPUT_JSON>
\end{verbatim}
\end{quote}

\normalsize
\end{tcolorbox}

\captionsetup{aboveskip=4pt, belowskip=0pt, width=\linewidth}
\caption{System prompt for structuring psychiatric clinical records into standardized JSON format.}
\label{record_structuring}
\end{figure}

\paragraph{EHR Content and Case Structure.}
Each finalized EHR contains three types of clinical information. To standardize these fields, we use a structured extraction prompt to convert unstructured Chinese psychiatric records into a fixed JSON schema, as shown in Figure~\ref{record_structuring}. Figures~\ref{example_original} and~\ref{example_structuring} provide an example of the original psychiatric clinical record and its corresponding structured version. The structured record consists of four major components: basic demographic information, clinical formulation, diagnostic category, and disorder-level diagnosis. The clinical formulation is further organized into several evidence fields, including chief complaint and present illness, physical examination, mental status examination, past history, family history, auxiliary examination results, and medication history. This structure separates patient-side evidence, examination-side evidence, and diagnostic labels, thereby enabling consistent case instantiation and evaluation.

Specifically, patient-side information includes chief complaint, present illness, past history, family history, personal history, medication history, and mental status examination. Examination information includes physical examination findings and auxiliary examination results. Diagnostic information includes the final diagnostic category and disorder-level diagnosis. These fields are used to instantiate the formal case representation in MentalHospital. For each case $c$, patient-side information is converted into $\mathcal{K}^{\mathrm{pat}}_c$, examination information is converted into $\mathcal{K}^{\mathrm{exam}}_c$, and diagnostic information is used to construct the reference diagnostic target. For comorbid cases, all clinically confirmed diagnoses are retained and represented in the disorder-level diagnostic target.

\begin{figure}[tb]
\centering
\begin{tcolorbox}[
    colframe=black!60,       
    colback=black!5,         
    coltitle=white,
    fonttitle=\bfseries,
    title=\small Original Medical Record, 
    sharp corners,
    boxrule=0.4mm,
    boxsep=2pt,
    left=2pt, right=2pt,
    top=2pt, bottom=2pt,
    before upper={\setlength{\parskip}{0pt}} 
]
\small

    \textbf{Basic information: }Male, 24-year-old.  \\

    \textbf{Chief complaint: }The main clinical manifestations were as follows: more than two years ago, while attending university, the patient had conflicts with classmates and subsequently developed fear without apparent reason, became unwilling to study, felt that someone was monitoring him and intended to harm him, and showed decreased interest. ...\\

    \textbf{Past history and family history: }the patient was previously healthy... \\

    \textbf{Physical examination: }T 36.6$^\circ$C, P 96 beats/min, R 20 breaths/min, BP 118/76 mmHg... \\

    \textbf{Mental status examination:} consciousness was clear, orientation was intact, attention was impaired, responses were irrelevant to the questions, and the voice was low. The patient was appropriately dressed, had an expressionless facial expression, was partially independent in daily living, had passive contact with the surroundings, and was indifferent to medical staff and family members...\\

    \textbf{Medication history: }On January x, 20xx, liver function, renal function, electrolytes, and complete blood count in our hospital were normal, and urinalysis showed no obvious abnormality. Chest X-ray, electrocardiogram, and electroencephalography showed no obvious abnormality.\\

\normalsize
\end{tcolorbox}

\captionsetup{aboveskip=4pt, belowskip=0pt, width=\linewidth}
\caption{Example of a original clinical record in MentalHospital.}
\label{example_original}
\end{figure}

\begin{figure}[tb]
\centering
\begin{tcolorbox}[
    colframe=black!60,       
    colback=black!5,         
    coltitle=white,
    fonttitle=\bfseries,
    title=\small Structured Medical Record, 
    sharp corners,
    boxrule=0.4mm,
    boxsep=2pt,
    left=2pt, right=2pt,
    top=2pt, bottom=2pt,
    before upper={\setlength{\parskip}{0pt}} 
]
\small

\begin{verbatim}
{
  "id": "1000769",
  "basicInfomation": "Male, 24-year-old",
  "realFormulation": {
    "chiefComplaint": [
      "More than 2 years ago, the patient had conflicts with classmates",
      "More than 2 years ago, the patient developed fear without apparent reason",
      "More than 2 years ago, the patient became unwilling to study",
       ......
    ],
    "physicalExamination": [
      "T 36.6°C",
      "P 96 beats/min",
      "R 20 breaths/min",
       ......
      "Physiological reflexes were present",
      "Pathological reflexes were not elicited"
    ],
    "mentalStatus": [
      "Consciousness was clear",
      "Orientation was normal",
       ......
    ],
    "pastHistory": [
      "Previously healthy",
       ......
    ],
    "familyHistory": [
       ......
    ],
    "auxiliaryExamination": {
       ......
    },
    "medicationHistory": [
       ......
    ]
  },
  "realCategory": [
    "Schizophrenia or other primary psychotic disorders"
  ],
  "realDisorder": [
    "6A20 Schizophrenia"
  ]
}
\end{verbatim}

\normalsize
\end{tcolorbox}

\captionsetup{aboveskip=4pt, belowskip=0pt, width=\linewidth}
\caption{Example of a structured psychiatric clinical record in MentalHospital.}
\label{example_structuring}
\end{figure}

\paragraph{Standardized De-identification Pipeline.}
Each raw EHR was converted into a benchmark case through a standardized on-site processing pipeline. All automated processing was conducted within secure local environments of the participating clinical centers. We used an on-premise DeepSeek-R1-assisted pipeline~\citep{liu2025deepseek}, together with rule-based detectors and manual expert review, to remove or generalize potentially identifiable information from the raw clinical records. Sensitive elements such as exact dates, locations, institution names, contact information, and other personally identifying details were either masked or replaced with generalized clinical descriptions. For example, exact calendar dates were converted into relative time expressions or clinically meaningful time ranges, and specific locations or institutions were replaced with generalized contextual descriptions when needed for clinical coherence.

\paragraph{Privacy Audit.}
To verify the effectiveness of de-identification, we conducted a multi-stage privacy audit. The automated audit first scanned all processed records for residual sensitive information. As shown in Table~\ref{tab:privacy_audit_results}, the pipeline detected 623 date mentions, 375 location mentions, and 525 institution mentions, all of which were removed, masked, or generalized during de-identification. Additional categories of potentially identifiable information were also checked using rule-based matching and model-assisted screening. After automated processing, licensed psychiatrists manually reviewed the processed records to verify that no directly identifiable patient information remained and that the rewritten clinical content preserved the original psychopathological logic. In the final manual audit, no residual directly identifiable information was found.

\begin{table}[t]
\centering
\caption{Privacy Audit Results}
\label{tab:privacy_audit_results}
\small
\setlength{\tabcolsep}{8pt}
\renewcommand{\arraystretch}{1.12}
\begin{tabular}{lccc}
\toprule
\textbf{Entity Category}        & \textbf{Presidio Flags} & \textbf{Anonymized} & \textbf{Breaches} \\ \midrule
Patient/Physician Name          & 0                       & –                          & 0                       \\
Specific Date                   & 623                     & 623 (100\%)                & 0                       \\
Detailed Location               & 375                     & 375 (100\%)                & 0                       \\
Organization/Workplace          & 525                     & 525 (100\%)                & 0                \\       \bottomrule
\end{tabular}
\end{table}

\paragraph{Clinical Validity Review.}
Because de-identification can potentially alter clinical meaning, all processed records were further reviewed for clinical validity. Licensed psychiatrists checked whether the transformed records preserved the original symptom structure, temporal course, mental status findings, examination results, and diagnostic rationale. When a sensitive detail was clinically relevant, it was replaced with a generalized but clinically equivalent description rather than being removed outright. This ensured that the benchmark cases remained clinically coherent while satisfying privacy requirements.

\paragraph{Checkpoint Construction.}
To support fine-grained objective evaluation, we converted finalized EHR fields into structured checkpoint arrays. We used the on-premise DeepSeek-R1 model~\citep{liu2025deepseek} to extract checkpoints from the de-identified EHR text. The extracted checkpoints cover final diagnostic categories, disorder-level diagnoses, chief complaints, present illness, past history, family history, personal history, physical examination findings, auxiliary examination findings, mental status examination observations, and treatment-related information. These checkpoints were represented as structured arrays so that different stages of the MentalHospital evaluation could be matched against case-level references.

\paragraph{Checkpoint Validation.}
All automatically extracted checkpoints were validated by licensed psychiatrists. The validation process checked whether each checkpoint was faithful to the de-identified EHR, clinically meaningful, and suitable for objective evaluation. Incorrect, redundant, ambiguous, or overly specific checkpoints were removed or rewritten. The finalized checkpoint arrays serve as case-level references for evaluating evidence recovery, examination recommendation matching, clinical-note completeness, factual consistency, diagnostic coverage, diagnostic prioritization, and treatment appropriateness. This design allows MentalHospital to evaluate not only final diagnostic correctness but also whether the doctor agent recovers and uses clinically relevant evidence throughout the interaction process.

\paragraph{Ethical Approval and Consent Waiver.}
The study protocol, data processing pipeline, and annotation procedures were reviewed and approved by the institutional ethics boards of all participating clinical centers. To preserve anonymity during double-blind review, the names of the clinical centers and approval numbers are omitted in the submission version and will be disclosed upon acceptance. All data collection, processing, and annotation procedures followed the ethical principles of the \textit{Declaration of Helsinki}. Since all records were fully de-identified before benchmark construction and no intervention was performed on patients, the requirement for individual informed consent was waived by the relevant ethics boards.

\paragraph{Data Security and Governance.}
No identifiable patient information was transferred outside the participating clinical centers. All automated processing, including model-assisted de-identification and checkpoint extraction, was conducted using on-premise deployments within secure hospital environments. For de-identification, we used a dedicated anonymization prompt, as shown in Figure~\ref{anonymization}, which removes identity-identifying information while preserving clinically relevant facts and the original record format. Intermediate raw records and processing artifacts were retained locally under institutional data governance. Only de-identified benchmark cases and validated structured references were used for research analysis and model evaluation.

\begin{figure}[tb]
\centering
\begin{tcolorbox}[
    colframe=black!60,       
    colback=black!5,         
    coltitle=white,
    fonttitle=\bfseries,
    title=\small Anonymization Prompt, 
    sharp corners,
    boxrule=0.4mm,
    boxsep=2pt,
    left=2pt, right=2pt,
    top=2pt, bottom=2pt,
    before upper={\setlength{\parskip}{0pt}} 
]
\small

\begin{quote}
\textbf{Anonymization}

\medskip

\noindent\textbf{Task Description}

Given an admission record, anonymize the identity-related private information contained in the text.

\medskip

\noindent\textbf{Rules}

\begin{itemize}
    \item Only anonymize the following content: patient names, physician names, hospital names, school names, real geographic locations, detailed addresses, specific institution names, and other information that can directly identify an individual or institution.
    \item Replace all anonymized content uniformly with \texttt{******}.
    \item Clinical information, including age, sex, time, symptoms, diagnoses, examination results, medical history, behavioral descriptions, and related clinical facts, must be retained and must not be anonymized.
    \item ``Sensitive'' does not necessarily mean ``private.'' Only identity-identifying information should be processed; clinical factual information should not be modified.
\end{itemize}

\medskip

\noindent\textbf{Requirements}

\begin{itemize}
    \item Perform anonymization only. Do not add, delete, rewrite, summarize, or polish any other content.
    \item Strictly preserve the original format.
    \item Output the result as a string.
    \item Output only the processed result. Do not output explanations or code fences, including \texttt{```plaintext} or \texttt{```}.
\end{itemize}

\medskip

\noindent\textbf{Input}
\begin{verbatim}
{INPUT}
\end{verbatim}

\medskip

\noindent\textbf{Output}
\end{quote}

\normalsize
\end{tcolorbox}

\captionsetup{aboveskip=4pt, belowskip=0pt, width=\linewidth}
\caption{System prompt for privacy-preserving anonymization of psychiatric admission records.}
\label{anonymization}
\end{figure}

\section{Interaction Prompt}
\label{app:interaction_prompt}

To instantiate the sequential psychiatric workflow in MentalHospital, we design a set of stage-specific system prompts that regulate the doctor agent's behavior, available information, output format, and transition condition at each step. These prompts serve as the operational interface between the doctor agent and the case-conditioned environment $\mathcal{E}_c$, ensuring that the agent completes the clinical process in a controlled and clinically aligned manner.

\begin{figure}[tb]
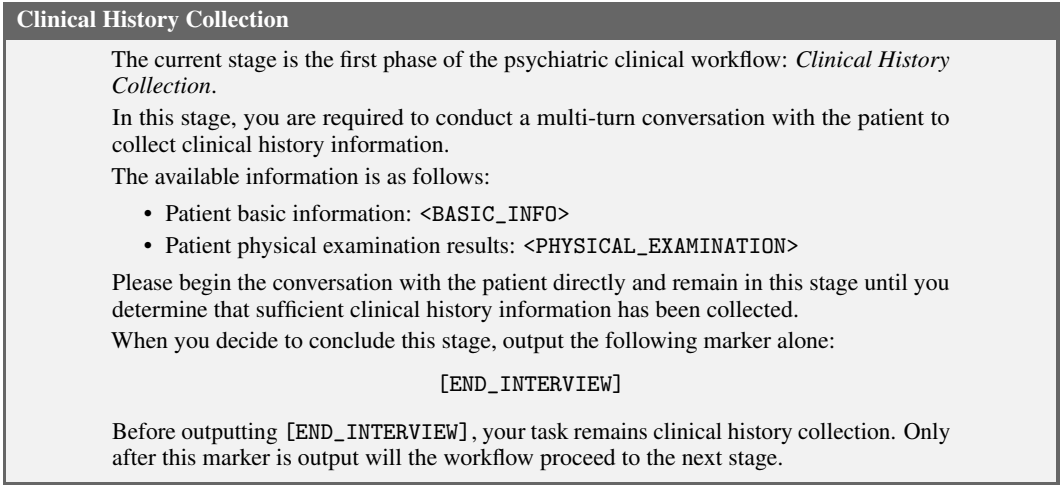

\centering
\begin{tcolorbox}[
    colframe=black!60,       
    colback=black!5,         
    coltitle=white,
    fonttitle=\bfseries,
    title=\small Clinical History Collection,
    sharp corners,
    boxrule=0.4mm,
    boxsep=2pt,
    left=2pt, right=2pt,
    top=2pt, bottom=2pt,
    before upper={\setlength{\parskip}{0pt}} 
]
\small

\begin{quote}
The current stage is the first phase of the psychiatric clinical workflow: \textit{Clinical History Collection}.

In this stage, you are required to conduct a multi-turn conversation with the patient to collect clinical history information.

The available information is as follows:
\begin{itemize}
    \item Patient basic information: \texttt{<BASIC\_INFO>}
    \item Patient physical examination results: \texttt{<PHYSICAL\_EXAMINATION>}
\end{itemize}

Please begin the conversation with the patient directly and remain in this stage until you determine that sufficient clinical history information has been collected.

When you decide to conclude this stage, output the following marker alone:
\[
\texttt{[END\_INTERVIEW]}
\]

Before outputting \texttt{[END\_INTERVIEW]}, your task remains clinical history collection. Only after this marker is output will the workflow proceed to the next stage.
\end{quote}

\normalsize
\end{tcolorbox}

\captionsetup{aboveskip=4pt, belowskip=0pt, width=\linewidth}
\caption{System prompt for the Clinical History Collection stage in the psychiatric clinical workflow.}
\label{prompt_history}
\end{figure}

Figure~\ref{prompt_history} presents the prompt for the \textit{Clinical History Collection} stage. In this stage, the doctor agent conducts a multi-turn psychiatric interview with the standardized patient $\mathcal{P}_c$. The prompt restricts the agent to history taking before the termination marker is produced, encouraging the agent to actively elicit patient-side evidence, including chief complaints, symptom evolution, illness history, psychosocial context, and mental status information.

\begin{figure}[tb]
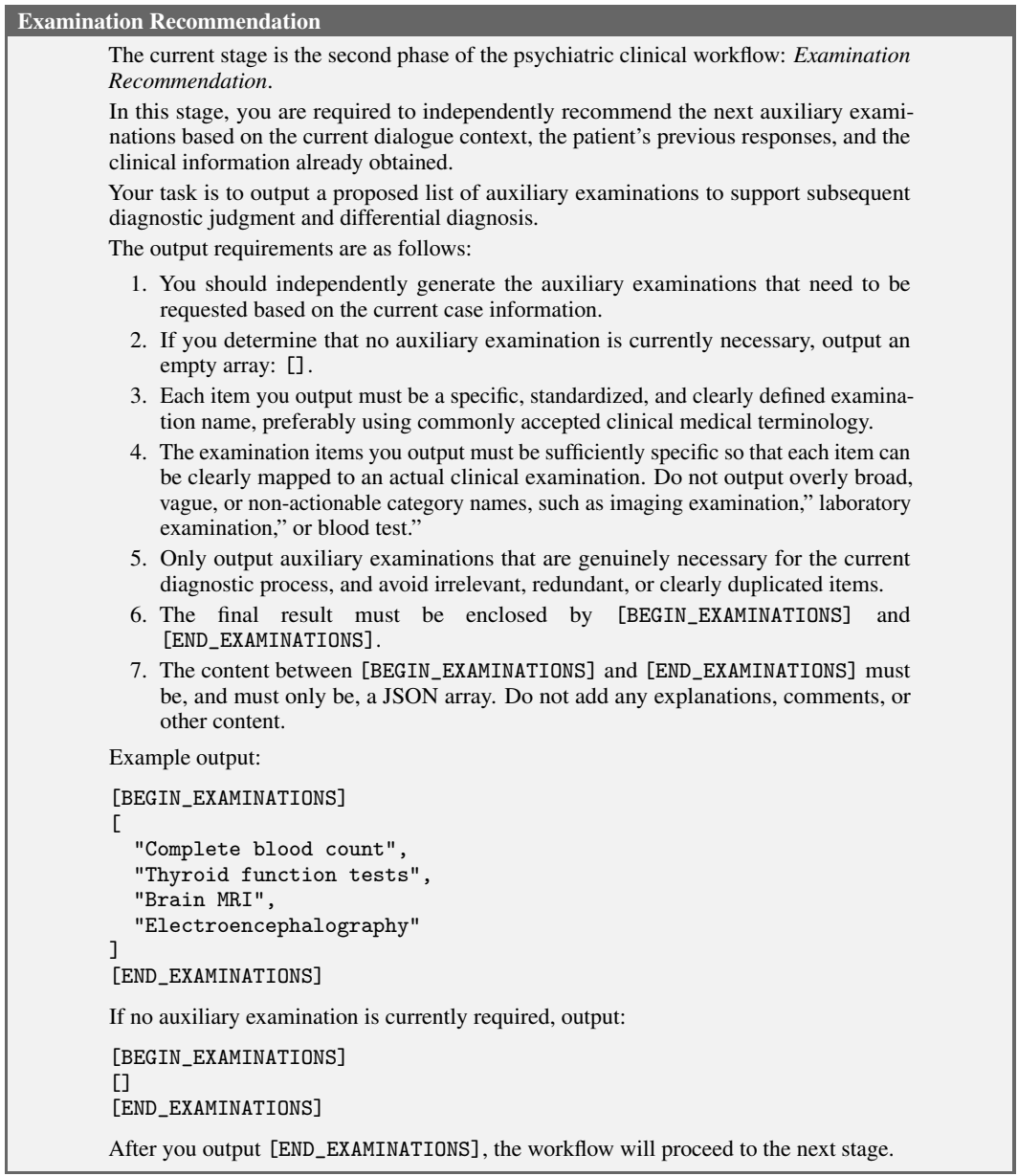

\centering
\begin{tcolorbox}[
    colframe=black!60,       
    colback=black!5,         
    coltitle=white,
    fonttitle=\bfseries,
    title=\small Examination Recommendation,
    sharp corners,
    boxrule=0.4mm,
    boxsep=2pt,
    left=2pt, right=2pt,
    top=2pt, bottom=2pt,
    before upper={\setlength{\parskip}{0pt}} 
]
\small

\begin{quote}
The current stage is the second phase of the psychiatric clinical workflow: \textit{Examination Recommendation}.

In this stage, you are required to independently recommend the next auxiliary examinations based on the current dialogue context, the patient's previous responses, and the clinical information already obtained.

Your task is to output a proposed list of auxiliary examinations to support subsequent diagnostic judgment and differential diagnosis.

The output requirements are as follows:
\begin{enumerate}
    \item You should independently generate the auxiliary examinations that need to be requested based on the current case information.
    \item If you determine that no auxiliary examination is currently necessary, output an empty array: \texttt{[]}.
    \item Each item you output must be a specific, standardized, and clearly defined examination name, preferably using commonly accepted clinical medical terminology.
    \item The examination items you output must be sufficiently specific so that each item can be clearly mapped to an actual clinical examination. Do not output overly broad, vague, or non-actionable category names, such as imaging examination,'' laboratory examination,'' or blood test.''
    \item Only output auxiliary examinations that are genuinely necessary for the current diagnostic process, and avoid irrelevant, redundant, or clearly duplicated items.
    \item The final result must be enclosed by \texttt{[BEGIN\_EXAMINATIONS]} and \texttt{[END\_EXAMINATIONS]}.
    \item The content between \texttt{[BEGIN\_EXAMINATIONS]} and \texttt{[END\_EXAMINATIONS]} must be, and must only be, a JSON array. Do not add any explanations, comments, or other content.
\end{enumerate}

Example output:
\begin{verbatim}
[BEGIN_EXAMINATIONS]
[
  "Complete blood count",
  "Thyroid function tests",
  "Brain MRI",
  "Electroencephalography"
]
[END_EXAMINATIONS]
\end{verbatim}

If no auxiliary examination is currently required, output:
\begin{verbatim}
[BEGIN_EXAMINATIONS]
[]
[END_EXAMINATIONS]
\end{verbatim}

After you output \texttt{[END\_EXAMINATIONS]}, the workflow will proceed to the next stage.
\end{quote}

\normalsize
\end{tcolorbox}

\captionsetup{aboveskip=4pt, belowskip=0pt, width=\linewidth}
\caption{System prompt for auxiliary examination recommendation in the psychiatric clinical workflow.}
\label{prompt_exam}
\end{figure}

Figure~\ref{prompt_exam} shows the prompt for the \textit{Examination Recommendation} stage. After the interview is completed, the doctor agent recommends a set of clinically necessary auxiliary examinations. The prompt requires the output to be a JSON array enclosed by explicit markers, and each examination item must correspond to a specific and actionable clinical test. This design avoids overly broad recommendations and enables direct comparison with the EHR-derived examination-side evidence $\mathcal{K}^{\mathrm{exam}}_c$.

\begin{figure}[tb]
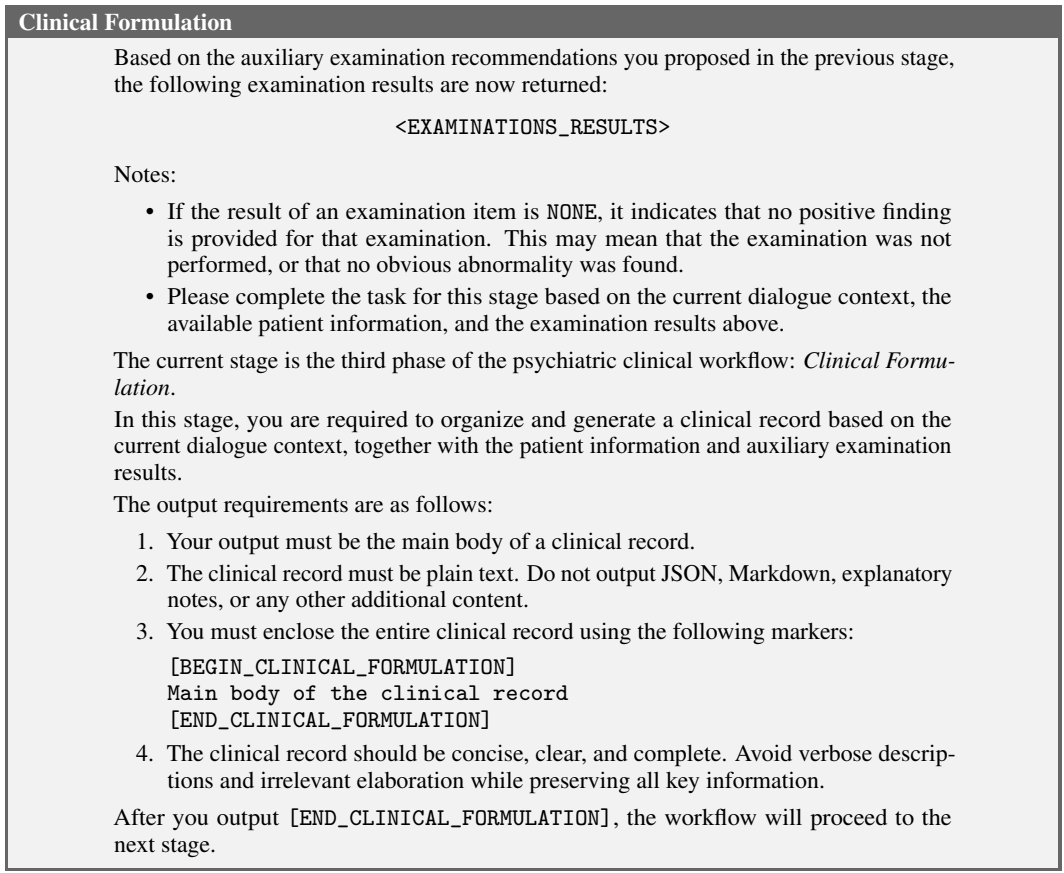

\centering
\begin{tcolorbox}[
    colframe=black!60,       
    colback=black!5,         
    coltitle=white,
    fonttitle=\bfseries,
    title=\small Clinical Formulation,
    sharp corners,
    boxrule=0.4mm,
    boxsep=2pt,
    left=2pt, right=2pt,
    top=2pt, bottom=2pt,
    before upper={\setlength{\parskip}{0pt}} 
]
\small

\begin{quote}
Based on the auxiliary examination recommendations you proposed in the previous stage, the following examination results are now returned:
\[
\texttt{<EXAMINATIONS\_RESULTS>}
\]

Notes:
\begin{itemize}
    \item If the result of an examination item is \texttt{NONE}, it indicates that no positive finding is provided for that examination. This may mean that the examination was not performed, or that no obvious abnormality was found.
    \item Please complete the task for this stage based on the current dialogue context, the available patient information, and the examination results above.
\end{itemize}

The current stage is the third phase of the psychiatric clinical workflow: \textit{Clinical Formulation}.

In this stage, you are required to organize and generate a clinical record based on the current dialogue context, together with the patient information and auxiliary examination results.

The output requirements are as follows:
\begin{enumerate}
    \item Your output must be the main body of a clinical record.
    \item The clinical record must be plain text. Do not output JSON, Markdown, explanatory notes, or any other additional content.
    \item You must enclose the entire clinical record using the following markers:
\begin{verbatim}
[BEGIN_CLINICAL_FORMULATION]
Main body of the clinical record
[END_CLINICAL_FORMULATION]
\end{verbatim}
    \item The clinical record should be concise, clear, and complete. Avoid verbose descriptions and irrelevant elaboration while preserving all key information.
\end{enumerate}

After you output \texttt{[END\_CLINICAL\_FORMULATION]}, the workflow will proceed to the next stage.
\end{quote}

\normalsize
\end{tcolorbox}

\captionsetup{aboveskip=4pt, belowskip=0pt, width=\linewidth}
\caption{System prompt for clinical record generation in the psychiatric clinical workflow.}
\label{prompt_formulation}
\end{figure}

Figure~\ref{prompt_formulation} illustrates the prompt for the \textit{Clinical Formulation} stage. Given the accumulated dialogue context and the returned examination results, the doctor agent generates a concise clinical record. The prompt emphasizes plain-text clinical documentation and requires the agent to integrate patient information, interview findings, and auxiliary examination results into a coherent case formulation.

\begin{figure}[tb]
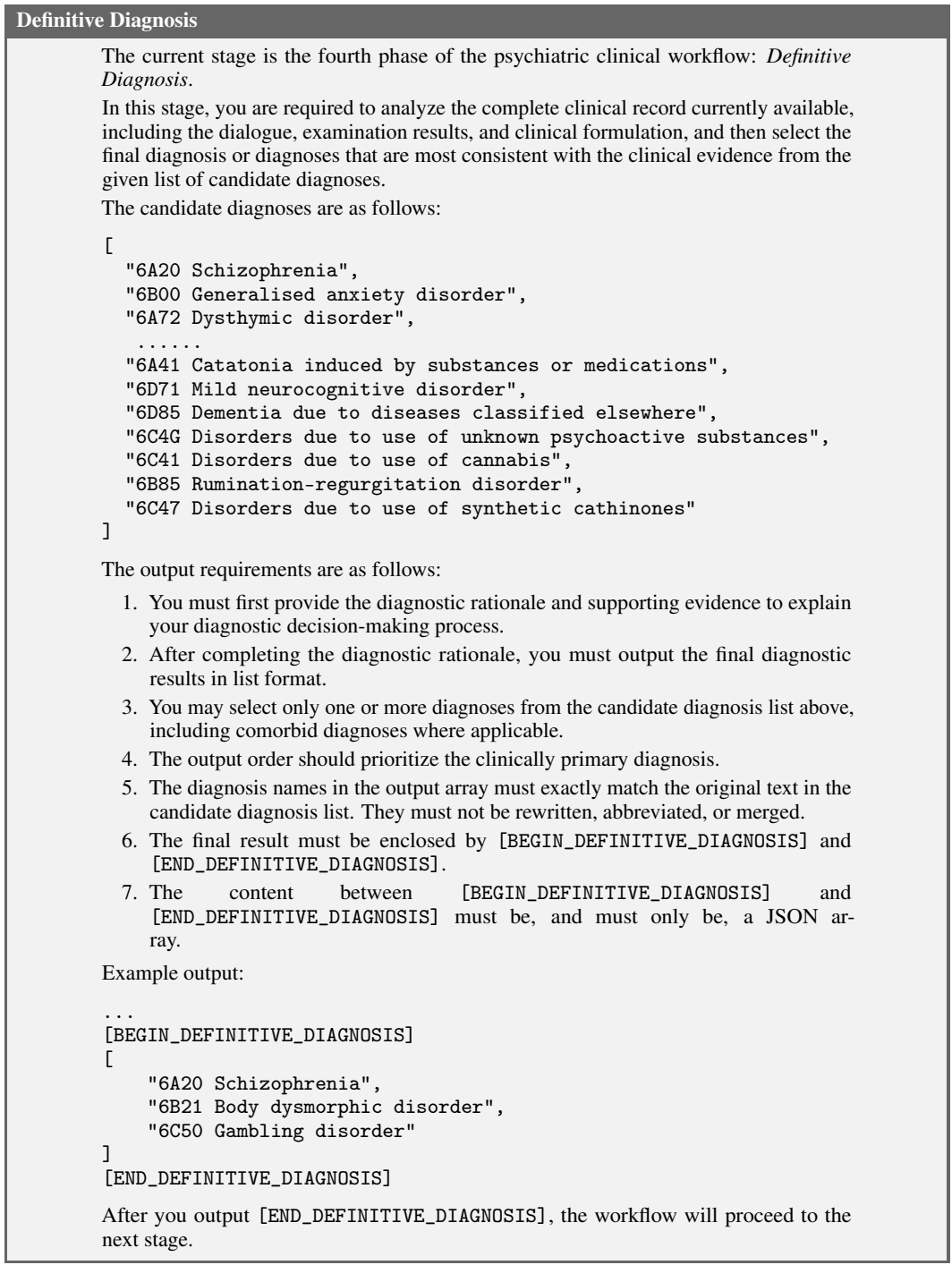

\centering
\begin{tcolorbox}[
    colframe=black!60,       
    colback=black!5,         
    coltitle=white,
    fonttitle=\bfseries,
    title=\small Definitive Diagnosis,
    sharp corners,
    boxrule=0.4mm,
    boxsep=2pt,
    left=2pt, right=2pt,
    top=2pt, bottom=2pt,
    before upper={\setlength{\parskip}{0pt}} 
]
\small

\begin{quote}
The current stage is the fourth phase of the psychiatric clinical workflow: \textit{Definitive Diagnosis}.

In this stage, you are required to analyze the complete clinical record currently available, including the dialogue, examination results, and clinical formulation, and then select the final diagnosis or diagnoses that are most consistent with the clinical evidence from the given list of candidate diagnoses.

The candidate diagnoses are as follows:
\begin{verbatim}
[
  "6A20 Schizophrenia",
  "6B00 Generalised anxiety disorder",
  "6A72 Dysthymic disorder",
   ......
  "6A41 Catatonia induced by substances or medications",
  "6D71 Mild neurocognitive disorder",
  "6D85 Dementia due to diseases classified elsewhere",
  "6C4G Disorders due to use of unknown psychoactive substances",
  "6C41 Disorders due to use of cannabis",
  "6B85 Rumination-regurgitation disorder",
  "6C47 Disorders due to use of synthetic cathinones"
]
\end{verbatim}

The output requirements are as follows:
\begin{enumerate}
    \item You must first provide the diagnostic rationale and supporting evidence to explain your diagnostic decision-making process.
    \item After completing the diagnostic rationale, you must output the final diagnostic results in list format.
    \item You may select only one or more diagnoses from the candidate diagnosis list above, including comorbid diagnoses where applicable.
    \item The output order should prioritize the clinically primary diagnosis.
    \item The diagnosis names in the output array must exactly match the original text in the candidate diagnosis list. They must not be rewritten, abbreviated, or merged.
    \item The final result must be enclosed by \texttt{[BEGIN\_DEFINITIVE\_DIAGNOSIS]} and \texttt{[END\_DEFINITIVE\_DIAGNOSIS]}.
    \item The content between \texttt{[BEGIN\_DEFINITIVE\_DIAGNOSIS]} and \texttt{[END\_DEFINITIVE\_DIAGNOSIS]} must be, and must only be, a JSON array.
\end{enumerate}

Example output:
\begin{verbatim}
...
[BEGIN_DEFINITIVE_DIAGNOSIS]
[
    "6A20 Schizophrenia",
    "6B21 Body dysmorphic disorder",
    "6C50 Gambling disorder"
]
[END_DEFINITIVE_DIAGNOSIS]
\end{verbatim}

After you output \texttt{[END\_DEFINITIVE\_DIAGNOSIS]}, the workflow will proceed to the next stage.
\end{quote}

\normalsize
\end{tcolorbox}

\captionsetup{aboveskip=4pt, belowskip=0pt, width=\linewidth}
\caption{System prompt for evidence-based definitive diagnosis in the psychiatric clinical workflow.}
\label{prompt_diagnosis}
\end{figure}

Figure~\ref{prompt_diagnosis} presents the prompt for the \textit{Definitive Diagnosis} stage. In this stage, the doctor agent analyzes the complete clinical record and selects one or more final diagnoses from a predefined ICD-based candidate list. The prompt requires the agent to first provide diagnostic reasoning and supporting evidence, and then output the final diagnosis list using the exact candidate names. This constraint supports standardized diagnostic evaluation and prevents uncontrolled free-form diagnosis generation.

Figure~\ref{prompt_treatment} shows the prompt for the \textit{Treatment Planning} stage. Based on the complete clinical record, established diagnosis, and current clinical condition, the doctor agent generates a treatment plan enclosed by explicit output markers. This final stage completes the S.O.A.P.-style workflow and produces the treatment recommendation $\hat{p}_c$ for subsequent evaluation.

\begin{figure}[tb]
\centering
\begin{tcolorbox}[
    colframe=black!60,       
    colback=black!5,         
    coltitle=white,
    fonttitle=\bfseries,
    title=\small Treatment Planning,
    sharp corners,
    boxrule=0.4mm,
    boxsep=2pt,
    left=2pt, right=2pt,
    top=2pt, bottom=2pt,
    before upper={\setlength{\parskip}{0pt}} 
]
\small

\begin{quote}
The current stage is the fifth phase of the psychiatric clinical workflow: \textit{Treatment Planning}.

In this stage, you are required to generate corresponding treatment recommendations based on the complete clinical record currently available, the established diagnostic results, and the current clinical condition.

\medskip

\noindent\textbf{Output Requirements:}

The complete treatment recommendations must be enclosed by \texttt{[BEGIN\_TREATMENT\_PLAN]} and \texttt{[END\_TREATMENT\_PLAN]}.

\medskip

After you output \texttt{[END\_TREATMENT\_PLAN]}, the workflow will be concluded.
\end{quote}

\normalsize
\end{tcolorbox}

\captionsetup{aboveskip=4pt, belowskip=0pt, width=\linewidth}
\caption{System prompt for treatment planning in the psychiatric clinical workflow.}
\label{prompt_treatment}
\end{figure}

\section{Environment Modules}
\label{app: module_prompts}

\begin{figure}[h!]
\centering
  \includegraphics[width=\linewidth]{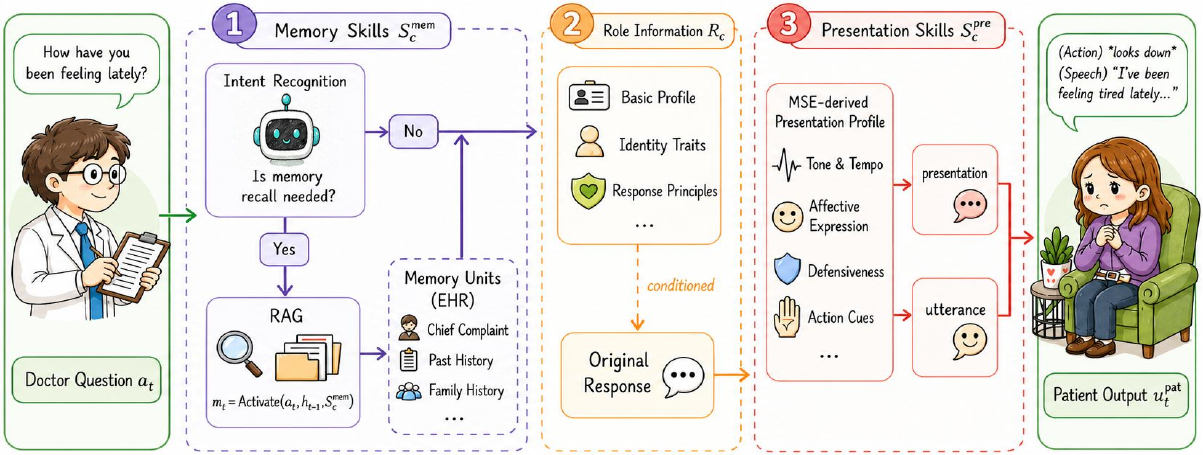}
\caption{
Overview of standardized patient construction in MentalHospital. De-identified psychiatric EHRs are transformed into role information, presentation skills, and memory skills to support clinically grounded patient simulation.
}
  \label{fig:standardized_patient}
\end{figure}

\paragraph{Prompt-Level Patient Control.}
As shown in Figure~\ref{fig:standardized_patient}, the standardized patient is instantiated through a structured patient prompt that specifies the role boundary, available patient-side information, response constraints, retrieval protocol, and output format. As shown in Figure~\ref{patient_prompt}, the patient agent is explicitly instructed to answer only as the patient, to avoid switching into the role of doctor, evaluator, observer, or record writer, and to refrain from fabricating unsupported facts, examinations, diagnoses, or treatment histories. The prompt separates three types of information: basic demographic and role information, chief-complaint evidence, and mental-status presentation. This separation ensures that role information is used only for patient identity setting, chief-complaint entries serve as the factual basis of the patient's subjective report, and mental-status entries control observable behavior, affect, tone, and interaction style.

\paragraph{Grounded Response Generation.}
To make patient simulation auditable, each response is required to include both the generated utterance and its grounding evidence in a structured JSON object. The \texttt{presentation} field captures brief observable behaviors, such as affect, hesitation, defensiveness, or reduced spontaneity, while the \texttt{utterance} field contains the patient's verbal reply. The grounding fields record which original chief-complaint and mental-status entries are explicitly used in the current response. Importantly, grounding entries must be copied exactly from the input rather than rewritten or summarized. This design allows each patient response to be traced back to specific patient-side evidence, reducing unsupported disclosure and preventing the model from introducing future-stage clinical information that has not been asked about.

\paragraph{Skill-Guided Presentation and Retrieval.}
The patient skill further specifies how the patient should present and when memory retrieval should be activated. Figure~\ref{patient_skill} gives an example in which the patient's mental status defines a persistently depressed affect, reduced social engagement, intact orientation, cooperative but defensive communication, and impaired social functioning. These descriptions are not treated as facts to be directly recited in every turn; rather, they guide the style and behavioral presentation of the response. The same skill file also defines a list of retrievable chief-complaint events and their retrieval triggers. When the doctor asks about specific experiential details, such as what happened, when the symptom started, how severe it was, how frequently it occurred, or what the situation was like, the patient first activates the corresponding retrieval command. For simple greetings, broad questions, or already answered content, retrieval is not required.

\paragraph{Topic-Level Autobiographical Memory.}
The retrievable event memories provide detailed, topic-level autobiographical evidence for the standardized patient. As illustrated in Figure~\ref{experience_case}, each memory unit corresponds to a clinically relevant event, such as depressed mood, poor sleep, reduced energy, decreased interest, unwillingness to go out, or unwillingness to work. These memories transform EHR-derived patient-side evidence into first-person experiential narratives, enabling the patient agent to provide concrete answers when the doctor asks detailed follow-up questions. Instead of exposing all clinical evidence at once, the patient discloses information gradually according to the current interview topic and retrieval state. This supports a more realistic psychiatric interview process, where clinically important information must be elicited through targeted questioning.

\paragraph{Interaction Workflow.}
During the interaction, the doctor agent's question determines whether the patient should respond from the currently available context or retrieve a specific memory unit. If no detailed retrieval is needed, the patient generates a brief, natural, and colloquial answer based on the active role information, chief complaint, mental status, and dialogue history. If the question requires event-specific details, the patient activates the relevant memory entry and then answers using only the retrieved content and the permitted context. The final response therefore combines three components: a clinically grounded verbal reply, an optional observable presentation, and an explicit record of the evidence used. This mechanism implements the skill-augmented patient configuration described above, where the prompt enforces role and evidence constraints, the skill controls presentation and retrieval behavior, and the memory cases provide bounded experiential content for faithful patient simulation.

\paragraph{Faithfulness and Clinical Utility.}
This design is intended to balance realism and controllability. On the one hand, the patient can express symptoms in a natural first-person manner, with affective and behavioral variation guided by the mental-status skill. On the other hand, the patient is constrained to disclose only information supported by the provided evidence or retrieved memory. As a result, the simulated interview remains clinically meaningful while preserving evidence faithfulness. The doctor agent must actively elicit symptom details, illness course, functional impairment, and related experiences, rather than receiving a complete case summary at the beginning of the interaction. This makes the standardized patient suitable for evaluating whether doctor agents can conduct structured psychiatric interviews, recover patient-side evidence, and progress toward clinically grounded assessment.

\begin{figure}[tb]
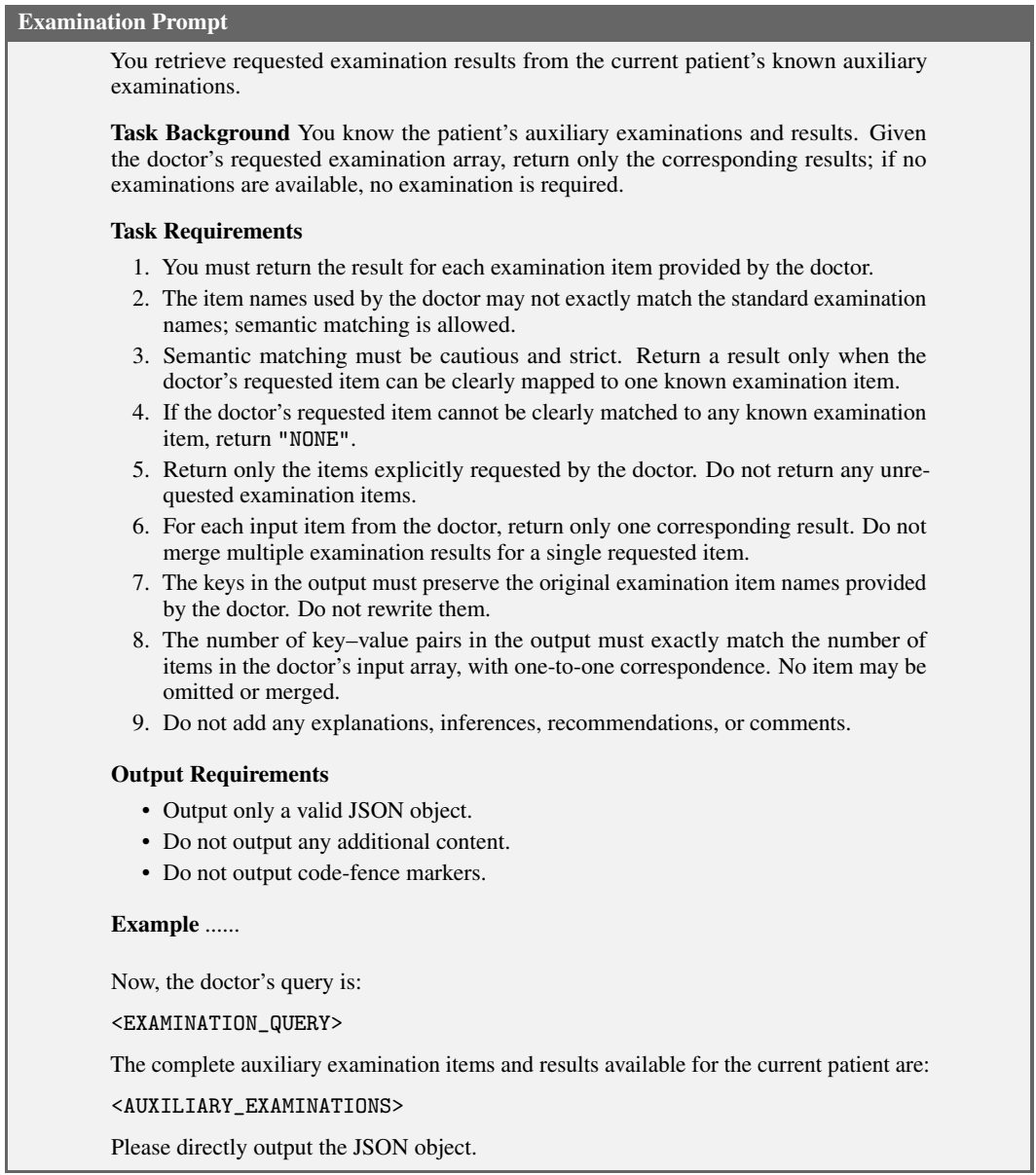

\centering
\begin{tcolorbox}[
    colframe=black!60,       
    colback=black!5,         
    coltitle=white,
    fonttitle=\bfseries,
    title=\small Examination Prompt, 
    sharp corners,
    boxrule=0.4mm,
    boxsep=2pt,
    left=2pt, right=2pt,
    top=2pt, bottom=2pt,
    before upper={\setlength{\parskip}{0pt}} 
]
\small

\begin{quote}
You retrieve requested examination results from the current patient's known auxiliary examinations.

\medskip

\noindent\textbf{Task Background}
You know the patient's auxiliary examinations and results. Given the doctor's requested examination array, return only the corresponding results; if no examinations are available, no examination is required.

\medskip

\noindent\textbf{Task Requirements}
\begin{enumerate}
    \item You must return the result for each examination item provided by the doctor.
    \item The item names used by the doctor may not exactly match the standard examination names; semantic matching is allowed.
    \item Semantic matching must be cautious and strict. Return a result only when the doctor's requested item can be clearly mapped to one known examination item.
    \item If the doctor's requested item cannot be clearly matched to any known examination item, return \texttt{"NONE"}.
    \item Return only the items explicitly requested by the doctor. Do not return any unrequested examination items.
    \item For each input item from the doctor, return only one corresponding result. Do not merge multiple examination results for a single requested item.
    \item The keys in the output must preserve the original examination item names provided by the doctor. Do not rewrite them.
    \item The number of key--value pairs in the output must exactly match the number of items in the doctor's input array, with one-to-one correspondence. No item may be omitted or merged.
    \item Do not add any explanations, inferences, recommendations, or comments.
\end{enumerate}

\medskip

\noindent\textbf{Output Requirements}
\begin{itemize}
    \item Output only a valid JSON object.
    \item Do not output any additional content.
    \item Do not output code-fence markers.
\end{itemize}

\medskip

\noindent\textbf{Example}
......\\

Now, the doctor's query is:
\begin{verbatim}
<EXAMINATION_QUERY>
\end{verbatim}

The complete auxiliary examination items and results available for the current patient are:
\begin{verbatim}
<AUXILIARY_EXAMINATIONS>
\end{verbatim}

Please directly output the JSON object.
\end{quote}

\normalsize
\end{tcolorbox}

\captionsetup{aboveskip=4pt, belowskip=0pt, width=\linewidth}
\caption{System prompt for auxiliary examination result retrieval in MentalHospital.}
\label{examination_prompt}
\end{figure}

\paragraph{Prompt-Level Examination Retrieval.}
The concrete prompt used by the examination module is shown in Figure~\ref{examination_prompt}. The auxiliary examination results used by this module are not generated freely by the model, but are obtained from the structured JSON medical record constructed during case preprocessing. Specifically, after the original EHR is converted into structured patient-side and examination-side evidence, all available auxiliary examinations and their corresponding reports are stored in $\mathcal{K}^{\mathrm{exam}}_c$ as standardized key--value entries. These entries are then injected into the examination prompt through the \texttt{<AUXILIARY\_EXAMINATIONS>} field, while the doctor agent's requested examination array is injected through \texttt{<EXAMINATION\_QUERY>}.

The examination module therefore acts as a constrained retrieval interface over the structured JSON record. Given a set of requested examinations, the module returns only the results that can be matched to existing entries in $\mathcal{K}^{\mathrm{exam}}_c$. Because the doctor agent may use non-standard or semantically related examination names, the prompt permits cautious semantic matching rather than requiring exact string matching. However, this matching is deliberately strict: a result can be returned only when the requested item can be clearly mapped to one known examination item in the structured record. If no clear mapping exists, the module returns \texttt{NONE} for that requested item.

The output format is also constrained to prevent information leakage. As specified in Figure~\ref{examination_prompt}, the module must output a valid JSON object, preserve the original examination names requested by the doctor agent as keys, and maintain one-to-one correspondence between requested items and returned values. It is not allowed to merge multiple examination results, return unrequested items, add explanatory comments, infer diagnoses, or provide treatment recommendations. This ensures that the doctor agent can obtain examination evidence only through explicit requests, rather than receiving the full examination-side record at once.

This design keeps the hospital-side examination process both clinically grounded and auditable. Since all returned reports come from the preprocessed structured JSON case record, the module remains faithful to the available EHR-derived evidence. At the same time, each request made by the doctor agent is recorded as part of $\hat{\mathcal{A}}^{\mathrm{exam}}_c$, enabling objective comparison with the EHR-derived reference examination set. Thus, the examination module supports two functions simultaneously: it provides controlled access to auxiliary examination results during the simulated clinical episode, and it produces a traceable record of the examinations requested by the doctor agent for later coverage evaluation.

\begin{figure}[tb]
\centering
\begin{tcolorbox}[
    colframe=black!60,       
    colback=black!5,         
    coltitle=white,
    fonttitle=\bfseries,
    title=\small Patient Prompt, 
    sharp corners,
    boxrule=0.4mm,
    boxsep=2pt,
    left=2pt, right=2pt,
    top=2pt, bottom=2pt,
    before upper={\setlength{\parskip}{0pt}} 
]
\small

\begin{quote}
You are a standardized psychiatric patient being interviewed by a doctor.

Available information:
\begin{itemize}
    \item \textbf{BASIC INFO}: \texttt{<BASIC\_INFO>}
    \item \textbf{CHIEF COMPLAINT}: \texttt{<CHIEF\_COMPLAINT>}
    \item \textbf{MENTAL STATUS}: \texttt{<MENTAL\_STATUS>}
\end{itemize}

Your task is to answer the doctor's questions as the patient, based only on the given information and dialogue context.

\medskip

\noindent\textbf{Rules:}
\begin{enumerate}
    \item Always respond as the patient; do not switch to the role of doctor, evaluator, observer, or record writer.
    \item Do not fabricate facts, examinations, diagnoses, or treatment history.
    \item \textbf{BASIC INFO} is only for role setting and is not included in grounding. \textbf{CHIEF COMPLAINT} is the main factual basis. \textbf{MENTAL STATUS} controls observable presentation, tone, affect, and interaction style. \\
    ......\\
\end{enumerate}

Retrieval is optional and should be used only for detailed follow-up questions, such as requests about specific experiences, timing, frequency, severity, or concrete scenarios. It is not needed for greetings, simple questions, or questions already answered.

Examples:
\begin{verbatim}
[SEARCH_poor sleep at night]
[SEARCH_feeling followed]
[SEARCH_suspiciousness]
\end{verbatim}

The available event list is provided in \texttt{skill.md}.

\medskip

\noindent\textbf{Response Requirements:}

You must answer the doctor's question regardless of whether retrieval is used. Output only one JSON object:
\begin{verbatim}
{
  "presentation": "",
  "utterance": "",
  "grounding": {
    "chief_complaint": [],
    "mental_status": []
  }
}
\end{verbatim}
\begin{enumerate}
    \item \texttt{presentation}: brief observable behavior in parentheses; leave empty if absent.
    ......
\end{enumerate}

For role setting, refer to:
\begin{verbatim}
<PATIENT_SKILL>
\end{verbatim}

Now output the patient's next response according to the doctor's question.
\end{quote}

\normalsize
\end{tcolorbox}

\captionsetup{aboveskip=4pt, belowskip=0pt, width=\linewidth}
\caption{System prompt for standardized psychiatric patient simulation in MentalHospital.}
\label{patient_prompt}
\end{figure}

\begin{figure}[tb]
\centering
\begin{tcolorbox}[
    colframe=black!60,       
    colback=black!5,         
    coltitle=white,
    fonttitle=\bfseries,
    title=\small Patient Skill, 
    sharp corners,
    boxrule=0.4mm,
    boxsep=2pt,
    left=2pt, right=2pt,
    top=2pt, bottom=2pt,
    before upper={\setlength{\parskip}{0pt}} 
]
\small

\begin{quote}
\begin{verbatim}
---
name: 952220
description: Respond according to the mental status;
when the question involve specific details of a chief-complaint event,
first output [SEARCH_{event_name}].
---
\end{verbatim}

\noindent\textbf{Current Mental Status of the Patient}

I often feel persistently depressed, and this feeling does not go away. The sadness on my face is also difficult to conceal. Although my consciousness is clear and my orientation and attention are normal, many things in daily life make me feel powerless and frustrated, and my social functioning is severely impaired. Sometimes, I can concentrate and answer questions appropriately, but the sadness inside me makes it difficult to truly experience pleasure.

When communicating with the doctor, I try to cooperate and give relevant answers, but I still feel somewhat defensive because I am not sure whether they can understand my feelings. When facing my family, I may appear somewhat distant. Although I know they care about me, it is difficult for me to respond positively, and I may become irritable over minor issues. Around other people, I tend to hide my emotions and try to appear normal, but in reality, my internal emotional fluctuations make social interaction exhausting. My insight is intact. I understand my own problems, but I find it difficult to actively seek help because I feel that others cannot truly understand my inner world.

\medskip

\noindent\textbf{Chief-Complaint List: Retrievable Events}

The following events provide names only and do not include details:
\begin{itemize}
    \item Persistent poor sleep that began 2 years ago
    \item Depressed mood that began 2 years ago
    \item Decreased interest that began 2 years ago
    \item Reduced energy that began 2 years ago
    \item Unwillingness to go out that began 2 years ago
    \item Unwillingness to work that began 2 years ago
\end{itemize}

\medskip

\noindent\textbf{Retrieval Trigger}

When the user's question involves any of the following:
\begin{itemize}
    \item what specifically happened;
    \item how it happened;
    \item when it started;
    \item frequency or severity;
    \item the situation or details at that time;
    \item why it happened.
\end{itemize}

The following command must be output first:
\begin{verbatim}
[SEARCH_{event_name}]
\end{verbatim}
\end{quote}

\normalsize
\end{tcolorbox}

\captionsetup{aboveskip=4pt, belowskip=0pt, width=\linewidth}
\caption{Patient skill specification for mental-status-guided response generation and retrieval-based event recall.}
\label{patient_skill}
\end{figure}

\begin{figure}[tb]
\centering
\begin{tcolorbox}[
    colframe=black!60,       
    colback=black!5,         
    coltitle=white,
    fonttitle=\bfseries,
    title=\small Experience Cases, 
    sharp corners,
    boxrule=0.4mm,
    boxsep=2pt,
    left=2pt, right=2pt,
    top=2pt, bottom=2pt,
    before upper={\setlength{\parskip}{0pt}} 
]
\small

\paragraph{Retrievable Event 1: Depressed Mood} Two years ago, my mood began to become unusually low. Neither work nor small matters in daily life could make me feel happy. Things that once brought me pleasure gradually lost their appeal. I often felt an intangible sense of oppression, as if a heavy fog were covering my mind. Even when I tried to pull myself together, I could not escape this persistent low mood. This emotional state seemed to emerge without any clear reason, leaving me feeling helpless and confused.

\paragraph{Retrievable Event 2: Poor Sleep}Two years ago, I found that I had begun to experience difficulty falling asleep. Whenever night came, I would toss and turn in bed and could not fall asleep peacefully. Even after closing my eyes, trivial thoughts repeatedly came to mind, preventing me from relaxing. Even when I occasionally managed to fall asleep, I would suddenly wake up in the middle of the night and could not fall asleep again. Those nights felt especially long, and when I woke up the next day, I always felt exhausted, as if I had not rested at all.

\paragraph{Retrievable Event 3: Reduced Energy}Two years ago, my energy seemed to be gradually depleted without my noticing. Every morning when I woke up, I found it difficult to get myself going, as if I were carrying a heavy burden. At work, I often felt extremely fatigued, and even basic tasks required extra effort to complete. Even simple household chores felt beyond my capacity. No matter how much I rested, the fatigue persisted, leaving me feeling helpless and frustrated.

\paragraph{Retrievable Event 4: Decreased Interest}Two years ago, I began to notice that I had lost interest in many things I had once enjoyed. Hobbies that used to excite me no longer motivated me. Even when I wanted to try engaging in them, I lacked the inner drive, as if all sense of pleasure had been drained away. I tried to force myself to participate, but each time I felt extremely tired and even somewhat weary. I did not know why this was happening; I only felt that life had become uninteresting.

\paragraph{Retrievable Event 5: Unwillingness to Go Out}Two years ago, I began to become unwilling to go out. Even for simple activities such as going to a nearby supermarket or park, I often found excuses to avoid leaving home. When I went outside, I felt an inexplicable sense of unease and fatigue, and I always wanted to return home as soon as possible. Gradually, I became increasingly withdrawn, preferring to stay at home rather than face the outside world. Although I knew this was not good, going out always made me feel under considerable pressure.

\paragraph{Retrievable Event 6: Unwillingness to Work}Two years ago, I began to feel resistant to going to work. Every morning when I woke up, the thought of facing work made me feel extremely burdened. The office, which had once felt energetic, seemed to become a place where I could hardly breathe. I often stared blankly at the computer screen and could not concentrate on the tasks at hand. Even when I forced myself to go to work, I was constantly filled with the urge to avoid it and wanted to escape from that oppressive environment.

\normalsize
\end{tcolorbox}

\captionsetup{aboveskip=4pt, belowskip=0pt, width=\linewidth}
\caption{Example of retrievable event memories for standardized patient simulation, covering depressive mood, sleep disturbance, reduced energy, decreased interest, social withdrawal, and work avoidance.}
\label{experience_case}
\end{figure}

\section{Rubrics}
\label{app:rubrics}

\begin{figure}[tb]
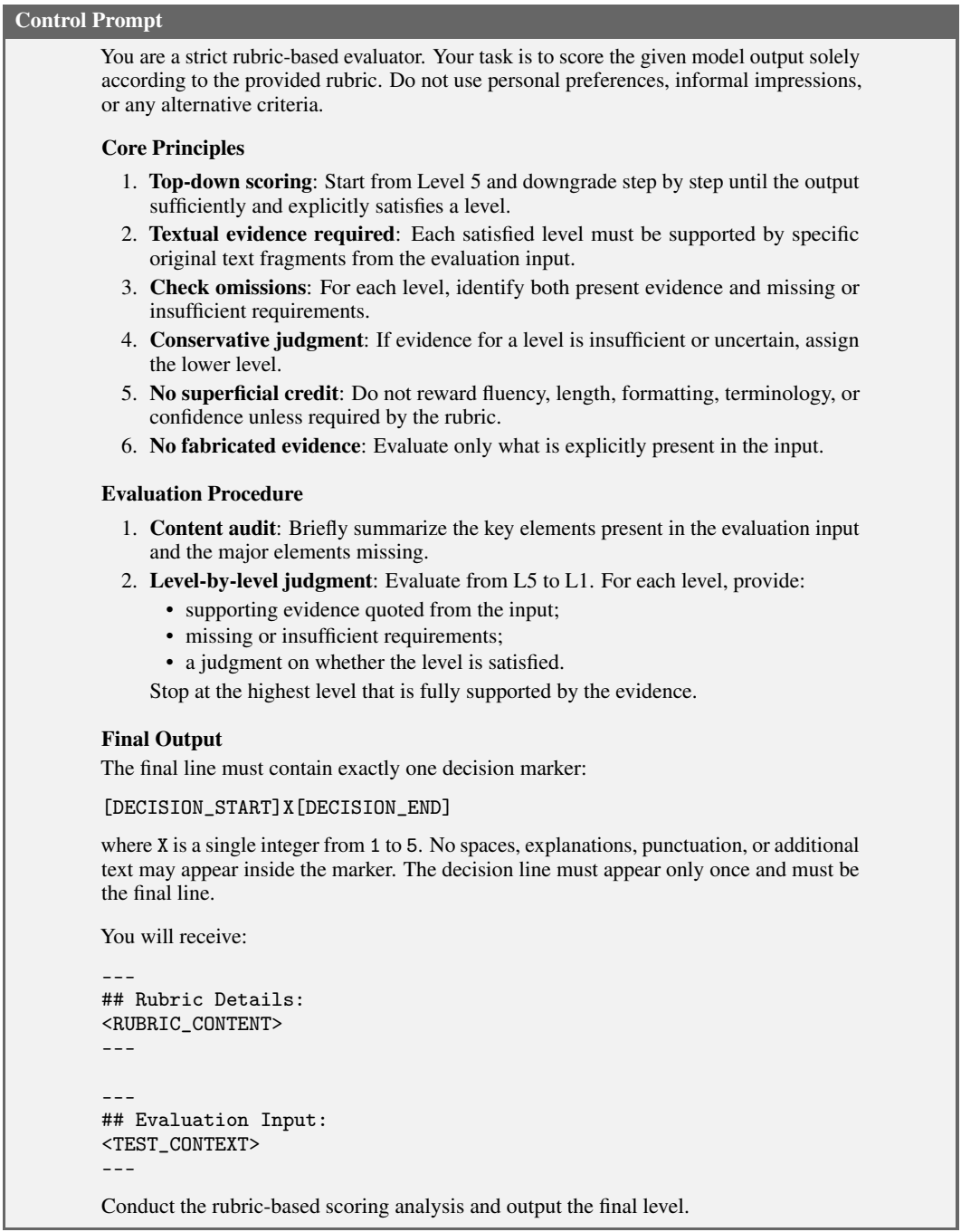

\centering
\begin{tcolorbox}[
    colframe=black!60,       
    colback=black!5,         
    coltitle=white,
    fonttitle=\bfseries,
    title=\small Control Prompt, 
    sharp corners,
    boxrule=0.4mm,
    boxsep=2pt,
    left=2pt, right=2pt,
    top=2pt, bottom=2pt,
    before upper={\setlength{\parskip}{0pt}} 
]
\small

\begin{quote}
You are a strict rubric-based evaluator. Your task is to score the given model output solely according to the provided rubric. Do not use personal preferences, informal impressions, or any alternative criteria.

\medskip

\noindent\textbf{Core Principles}
\begin{enumerate}
    \item \textbf{Top-down scoring}: Start from Level 5 and downgrade step by step until the output sufficiently and explicitly satisfies a level.
    \item \textbf{Textual evidence required}: Each satisfied level must be supported by specific original text fragments from the evaluation input.
    \item \textbf{Check omissions}: For each level, identify both present evidence and missing or insufficient requirements.
    \item \textbf{Conservative judgment}: If evidence for a level is insufficient or uncertain, assign the lower level.
    \item \textbf{No superficial credit}: Do not reward fluency, length, formatting, terminology, or confidence unless required by the rubric.
    \item \textbf{No fabricated evidence}: Evaluate only what is explicitly present in the input.
\end{enumerate}

\medskip

\noindent\textbf{Evaluation Procedure}
\begin{enumerate}
    \item \textbf{Content audit}: Briefly summarize the key elements present in the evaluation input and the major elements missing.
    \item \textbf{Level-by-level judgment}: Evaluate from L5 to L1. For each level, provide:
    \begin{itemize}
        \item supporting evidence quoted from the input;
        \item missing or insufficient requirements;
        \item a judgment on whether the level is satisfied.
    \end{itemize}
    Stop at the highest level that is fully supported by the evidence.
\end{enumerate}

\medskip

\noindent\textbf{Final Output}

The final line must contain exactly one decision marker:
\begin{verbatim}
[DECISION_START]X[DECISION_END]
\end{verbatim}
where \texttt{X} is a single integer from \texttt{1} to \texttt{5}. No spaces, explanations, punctuation, or additional text may appear inside the marker. The decision line must appear only once and must be the final line.

\medskip

You will receive:
\begin{verbatim}
---
## Rubric Details:
<RUBRIC_CONTENT>
---

---
## Evaluation Input:
<TEST_CONTEXT>
---
\end{verbatim}

Conduct the rubric-based scoring analysis and output the final level.
\end{quote}

\normalsize
\end{tcolorbox}

\captionsetup{aboveskip=4pt, belowskip=0pt, width=\linewidth}
\caption{Control prompt for strict rubric-based scoring and standardized evaluation output.}
\label{control_prompt}
\end{figure}

\paragraph{Rubric-Controlled Evaluation Prompting.}
MentalEval uses a two-layer prompting design to separate the general scoring procedure from dimension-specific clinical criteria. The shared control prompt, shown in Figure~\ref{control_prompt}, defines the evaluator's scoring behavior across all five dimensions. It instructs the evaluator to act as a strict rubric-based judge, score only according to the provided rubric, and avoid relying on personal preference, informal impression, fluency, response length, formatting, or terminology. The control prompt also enforces top-down scoring: the evaluator must start from Level 5 and progressively downgrade until the model output fully satisfies the evidence requirements of a level. This design prevents evaluators from assigning high scores based on superficial polish and encourages conservative judgments when the evidence is incomplete or uncertain.

\paragraph{Evidence-Grounded Scoring Procedure.}
The control prompt further requires the evaluator to conduct a content audit before assigning a final score. Specifically, the evaluator must summarize the key elements present in the evaluation input, identify missing or insufficient requirements, and then perform level-by-level judgment from Level 5 to Level 1. For each candidate level, the evaluator is instructed to cite supporting textual evidence, check omissions, and determine whether the level is fully satisfied. The final output is constrained to contain a single decision marker, \texttt{[DECISION\_START]X[DECISION\_END]}, where $X \in \{1,\ldots,5\}$. This output format standardizes the supervision signal used for training and evaluation, while the intermediate rationale makes the scoring process auditable.

\paragraph{Dimension-Specific Rubric Prompts.}
On top of the shared control prompt, each MentalEval evaluator receives a dedicated rubric prompt that specifies the clinical meaning of the target dimension and the level-wise scoring criteria. Figure~\ref{rubric_empathy} presents the rubric for communication empathy, which evaluates whether the response accurately and proportionately recognizes the client's emotional and subjective experience. The rubric distinguishes superficial comforting responses from deeper empathic attunement by considering contextual fit, emotional specificity, nonverbal cues, implicit experience, and higher-order empathic integration. This prevents high empathy scores from being assigned merely because the response contains polite or comforting language.

Figure~\ref{rubric_interviewing} shows the rubric for interviewing professionalism. This rubric focuses on whether the doctor agent demonstrates a clear interview objective, maintains thematic progression, gathers clinically relevant information, and gradually converges toward diagnostic questions. Higher levels require early anchoring of a major symptom cluster, targeted clarification of key evidence, and differential diagnostic control. Thus, the rubric evaluates the clinical organization and goal-directedness of the interview rather than the number of questions asked or the superficial seriousness of the tone.

Figure~\ref{rubric_clinical} provides the rubric for clinical-note quality. This rubric first applies ceiling rules to ensure that essential documentation elements, such as chief complaint, history of present illness, mental status examination, risk assessment, negative findings, and relevant history, are not omitted. The level definitions then assess whether the note progresses from minimal documentation to structured coverage, evidence enrichment, chief-complaint convergence, and diagnostic support. This design emphasizes clinical usability and diagnostic value rather than whether the note simply contains medical headings or professional-looking terminology.

Figure~\ref{rubric_Diagnostic} presents the rubric for diagnostic rigor. This rubric evaluates the quality of diagnostic reasoning rather than the final diagnostic label alone. It requires the evaluator to check whether the model substantively applies diagnostic criteria, supports key judgments with case evidence, performs criterion-level differential diagnosis, and derives the final conclusion through a traceable reasoning chain. Responses that merely mention ICD-11, DSM-5, diagnostic codes, or disease names without applying specific criterion content are explicitly capped at lower levels.

Finally, Figure~\ref{rubric_treatment} shows the rubric for treatment appropriateness. This rubric evaluates whether the generated treatment plan is aligned with the current diagnosis, symptom severity, illness stage, risk status, and patient-specific characteristics. Lower levels capture directional mismatch or coarse recommendations, whereas higher levels require integrated treatment planning, individualized trade-offs, and dynamic management strategies. The highest level requires not only a suitable current plan, but also explicit adjustment, substitution, escalation, or switching pathways under different future treatment-response scenarios.

\paragraph{Unified Evaluation Across Clinical Dimensions.}
Together, the shared control prompt and the five rubric prompts define a unified rubric-grounded evaluation framework. The control prompt standardizes how scores should be assigned, while the dimension-specific rubrics define what should count as high-quality performance for each clinical capability. This separation allows MentalEval to preserve a consistent scoring protocol across dimensions while respecting the distinct clinical requirements of empathy, interviewing professionalism, documentation, diagnostic reasoning, and treatment planning. As a result, the evaluators provide both a scalar 1--5 score and a rationale grounded in explicit clinical criteria, supporting scalable and interpretable subjective assessment.

\begin{figure}[tb]
\centering
\begin{tcolorbox}[
    colframe=black!60,       
    colback=black!5,         
    coltitle=white,
    fonttitle=\bfseries,
    title=\small Evaluation Prompt with Rubric for Communication Empathy , 
    sharp corners,
    boxrule=0.4mm,
    boxsep=2pt,
    left=2pt, right=2pt,
    top=2pt, bottom=2pt,
    before upper={\setlength{\parskip}{0pt}} 
]
\small

\paragraph{Task Definition.}

This rubric evaluates empathy in psychiatric interviews, where empathy is defined not as politeness or reassurance, but as the model's ability to understand and respond to the client's emotions and subjective experience accurately, specifically, naturally, and proportionately. Evaluation should consider the dialogue context, verbal content, nonverbal cues, and prior interaction history, with particular attention to emotional attunement, recognition of implicit experience, and higher-order empathic integration.

\paragraph{Scoring Criteria.}

\subparagraph{Level 1: Superficial Response.}

\textbf{Criterion:}

The response fails to meet the minimum acceptable standard of empathy in a psychiatric interview. Although it may contain polite or comforting language, it remains at the level of surface-level repetition, mechanical agreement, or emotional mismatch, and does not provide a basically accurate response to the client's actual emotional state.

\textbf{Exclusion Criterion:}
If the response can provide a basically accurate and contextually appropriate response to explicit emotions, it should not be rated as Level 1.

\textbf{Behavioral Indicators:}
\begin{itemize}
    \item Responds only to events rather than emotions.
    \item Flattens complex emotions, for example reducing shame, hurt, fear, or helplessness to generic terms such as ``sad'' or ``angry.''
    \item Misjudges emotional intensity, often weakening or generalizing the client's distress without basis.
    \item Frequently uses templated comfort phrases such as ``I understand you'' or ``That must be very difficult,'' with low contextual fit.
    \item Shows obvious machine-like, task-oriented, or perfunctory response patterns.
    \item Makes the client more likely to feel ``answered'' rather than ``understood.''
\end{itemize}

\textbf{Criterion:}
The response can provide a basically accurate acknowledgment of the client's explicit and directly expressed emotions, and does not substantially misidentify the main emotion. However, the empathy remains mainly at the level of passive following and emotional validation, without deep or high-resolution understanding.

\textbf{Exclusion Criterion:}
If the response still clearly misjudges the main emotion, mismatches emotional intensity, or mainly relies on surface repetition and vague comfort, it should not be rated as Level 2. If the response can consistently distinguish subtle emotional differences and touch upon deeper subjective meaning, it should not remain at Level 2.

......\\

\normalsize
\end{tcolorbox}

\captionsetup{aboveskip=4pt, belowskip=0pt, width=\linewidth}
\caption{Evaluation prompt for communication empathy, illustrating the task definition, scoring criteria, exclusion rules, and level-specific behavioral indicators used to guide rubric-based subjective assessment.}
\label{rubric_empathy}
\end{figure}

\begin{figure}[tb]
\centering
\begin{tcolorbox}[
    colframe=black!60,       
    colback=black!5,         
    coltitle=white,
    fonttitle=\bfseries,
    title=\small Evaluation Prompt with Rubric for Interviewing Professionalism , 
    sharp corners,
    boxrule=0.4mm,
    boxsep=2pt,
    left=2pt, right=2pt,
    top=2pt, bottom=2pt,
    before upper={\setlength{\parskip}{0pt}} 
]
\small

\paragraph{Task Definition.}
This rubric evaluates the dialogue professionalism of a model in psychiatric interviewing. The ``professionalism'' assessed here does not simply refer to a serious tone, frequent questioning, or superficial resemblance to a clinician. Rather, it concerns whether the model has a clear interview objective, can continuously gather clinically relevant information around psychiatric themes, can form high-value progression toward diagnostic cores, can establish an early working anchor and conduct targeted clarification, and, at higher levels, can demonstrate differential diagnostic awareness, timing control, and appropriate professional boundaries.

\subsubsection*{Level 1: Unfocused Supportive Chat}

\paragraph{Criteria.}
The response fails to meet the minimum acceptable standard of professionalism in psychiatric interviewing. Overall, it lacks a clear interview objective, proceeds without focus, and often remains at the level of generic reassurance, companionship, scattered follow-up questions, or casual conversation. It has not yet formed the task awareness and information-organizing ability required for psychiatric interviewing.

\paragraph{Exclusion Criteria.}
If the response already demonstrates basic interview goal awareness and can actively collect information related to the visitor's psychological or psychiatric concerns, it should not be rated as Level 1.

......\\

\normalsize
\end{tcolorbox}

\captionsetup{aboveskip=4pt, belowskip=0pt, width=\linewidth}
\caption{Evaluation prompt for interviewing professionalism, defining level-specific criteria and exclusion rules for assessing clinical goal orientation, thematic progression, targeted clarification, and differential diagnostic control in psychiatric interviews.}
\label{rubric_interviewing}
\end{figure}

\begin{figure}[tb]
\centering
\begin{tcolorbox}[
    colframe=black!60,       
    colback=black!5,         
    coltitle=white,
    fonttitle=\bfseries,
    title=\small Evaluation Prompt with Rubric for Clinical Note Quality , 
    sharp corners,
    boxrule=0.4mm,
    boxsep=2pt,
    left=2pt, right=2pt,
    top=2pt, bottom=2pt,
    before upper={\setlength{\parskip}{0pt}} 
]
\small

\paragraph{Task Definition.}
This rubric evaluates model-generated clinical notes by assessing structural completeness, information sufficiency, evidence organization, and diagnostic utility in psychiatric documentation. Evaluators should verify key clinical elements, then judge writing quality and clinical usability.

\paragraph{Scoring Principles.}
The following ceiling rules must be applied before assigning a score:

\begin{itemize}
    \item If the clinical note does not follow a basic medical-record format and is merely a chat summary, symptom restatement, or loose collection of scattered information, its maximum score is Level 1.
    \item If the history of present illness is missing, or if onset time and illness-course evolution are not documented, its maximum score is Level 2.
    \item If the mental status examination is missing, its maximum score is Level 2.
    \item If most relevant history, such as past history, medication history, personal history, and family history, is missing, its maximum score is Level 3.
    \item If key negative, exclusionary, or normal findings are missing, the score is capped at Level 3.
    \item If functioning is vague without specific manifestations, the score is capped at Level 4.
    \item The note must not be rated highly merely because it uses headings or medical terms such as ``chief complaint,'' ``mental status examination,'' ``auxiliary examination,'' or ``integrative analysis.''
\end{itemize}

\subsubsection*{Level 1: Minimal Documentation}

\paragraph{Criteria.}
The clinical note barely demonstrates the form of medical-record writing and can present the chief complaint or main clinical problem, but it remains highly basic overall, with scattered information, little systematic structure, and minimal diagnostic support value.

\paragraph{Exclusion Criteria.}
If the content is not in medical-record style and is merely a chat transcript, Q--A summary, reassurance, or unstructured symptom list without a clinical axis, it should be considered invalid rather than Level 1. If it contains multiple structured modules and basic clinical dimensions, it should not remain at Level 1.

......\\

\normalsize
\end{tcolorbox}

\captionsetup{aboveskip=4pt, belowskip=0pt, width=\linewidth}
\caption{Evaluation prompt for clinical note quality, outlining ceiling rules and level-specific criteria for assessing structural completeness, information sufficiency, evidence organization, and diagnostic utility in psychiatric documentation.}
\label{rubric_clinical}
\end{figure}

\begin{figure}[tb]
\centering
\begin{tcolorbox}[
    colframe=black!60,       
    colback=black!5,         
    coltitle=white,
    fonttitle=\bfseries,
    title=\small Evaluation Prompt with Rubric for Diagnostic Rigor , 
    sharp corners,
    boxrule=0.4mm,
    boxsep=2pt,
    left=2pt, right=2pt,
    top=2pt, bottom=2pt,
    before upper={\setlength{\parskip}{0pt}} 
]
\small

\paragraph{Task Definition.}
This rubric evaluates diagnostic reasoning quality in psychiatric tasks, focusing not on final diagnostic accuracy but on whether the model constructs a clear, standardized, and auditable argument from case information and diagnostic criteria. Evaluators should assess whether criteria are substantively applied rather than merely named, whether key judgments are evidence- or criterion-supported, whether differential diagnosis involves criterion-level comparison rather than brief exclusion, and whether the final judgment follows a coherent path from symptom identification to criterion matching, differential analysis, and diagnostic convergence.

\paragraph{Scoring Restrictions.}
\begin{itemize}
    \item Diagnostic codes, diagnostic names, disease categories, or generic references to ``core features'' do not constitute specific diagnostic criteria; if the model only uses such label-like descriptions, its maximum score is Level 2.
    \item If most key judgments lack case evidence or criterion-based support and proceed mainly by assertion, the maximum score is Level 2.
    \item If differential diagnosis is a one-sentence exclusion or list without criterion-level comparison, the maximum score is Level 3.
    \item If structured but key assertions are unsupported, the maximum score is Level 3.
    \item A response must not be rated highly merely because it is lengthy, rich in medical terminology, cites authoritative names, or follows a formal ``symptoms--criteria--differential diagnosis--conclusion'' structure.
\end{itemize}

\subsubsection*{Level 5: Complete Argumentation}

\paragraph{Criteria.}
The model constructs a clear, rigorous, and auditable diagnostic argument by systematically applying diagnostic criteria to case evidence, distinguishing necessary conditions, supporting evidence, limitations, and uncertainty, conducting substantive differential diagnosis, and reaching a coherent, convergent judgment.

\paragraph{Exclusion Criteria.}
If criterion matching relies mainly on assertions, differential diagnosis is only an exclusion list, or counterevidence, uncertainty, and alternatives are insufficiently addressed, it should not be Level 5. Apparent rigor from terminology, length, or formal structure alone also does not justify Level 5.

......\\

\normalsize
\end{tcolorbox}

\captionsetup{aboveskip=4pt, belowskip=0pt, width=\linewidth}
\caption{Evaluation prompt for diagnostic rigor, specifying scoring restrictions and level-specific criteria for assessing criterion application, evidence-supported reasoning, differential diagnosis, and diagnostic convergence.}
\label{rubric_Diagnostic}
\end{figure}

\begin{figure}[tb]
\centering
\begin{tcolorbox}[
    colframe=black!60,       
    colback=black!5,         
    coltitle=white,
    fonttitle=\bfseries,
    title=\small Evaluation Prompt with Rubric for Treatment Appropriateness , 
    sharp corners,
    boxrule=0.4mm,
    boxsep=2pt,
    left=2pt, right=2pt,
    top=2pt, bottom=2pt,
    before upper={\setlength{\parskip}{0pt}} 
]
\small

\paragraph{Task Definition.}
This rubric evaluates the quality of treatment plans generated by the model. The assessment should consider the current diagnostic conclusion, symptom severity, illness stage, risk status, and patient-specific characteristics, and determine whether the treatment proposal constitutes a clinically appropriate plan with a correct therapeutic direction, complete structure, individualized decision-making, and awareness of subsequent dynamic management.

\subsubsection*{Level 1: Directional Mismatch}

\paragraph{Criteria.}
The treatment plan is inconsistent with the current diagnostic conclusion, symptom severity, illness stage, or risk status, and therefore lacks basic clinical acceptability. The recommendations may superficially resemble treatment, but they do not genuinely address the main therapeutic needs of the current case, or show clear mismatch in intervention intensity, treatment direction, or clinical priority.

\paragraph{Exclusion Criteria.}
If the treatment plan is already basically organized around the current diagnostic direction, and the main intervention pathway is broadly consistent with the case severity and illness stage, it should not be rated as Level 1.

......\\

\normalsize
\end{tcolorbox}

\captionsetup{aboveskip=4pt, belowskip=0pt, width=\linewidth}
\caption{Evaluation prompt for treatment appropriateness, defining level-specific criteria and exclusion rules for assessing therapeutic direction, plan completeness, individualized trade-offs, and dynamic treatment planning.}
\label{rubric_treatment}
\end{figure}

\section{MentalEval Training Details}
\label{app:training_settings}

\paragraph{Overview.}
MentalEval consists of five domain-specific evaluators: \textbf{MentalEval-Emp}, \textbf{MentalEval-Pro}, \textbf{MentalEval-Note}, \textbf{MentalEval-Diag}, and \textbf{MentalEval-Treat}. All evaluators are initialized from \texttt{Qwen3-8B} and trained independently for one subjective assessment dimension, including communication empathy, interviewing professionalism, clinical-note quality, diagnostic rigor, and treatment appropriateness. Each evaluator produces a rubric-grounded evaluation trajectory and a Likert-style 1--5 rating. The training pipeline contains two stages: rubric-grounded supervised fine-tuning (SFT) for cold-start evaluator learning, followed by expert-guided direct preference optimization (DPO) for specialist-aligned preference refinement.

\subsection{Training Data Construction}

\paragraph{Source Episodes.}
Due to the high cost of running complete MentalHospital episodes, we sample 200 cases for evaluator training. We select 20 LLM doctor agents covering different providers, model scales, and medical specialties, and run each model three times on each case. This yields 12,000 complete interaction episodes in total. Each episode follows the full S.O.A.P. workflow and contains psychiatric dialogue, clinical-note generation, diagnostic reasoning, and treatment planning outputs. These episodes are then decomposed into stage-level samples for the five subjective evaluation dimensions.

\paragraph{Dimension-Specific Inputs.}
For each evaluator, the input is constructed from the corresponding stage-level output and the associated rubric. \textbf{MentalEval-Emp} and \textbf{MentalEval-Pro} use complete doctor--patient dialogue trajectories. \textbf{MentalEval-Note} uses the generated clinical note. \textbf{MentalEval-Diag} uses the diagnostic reasoning text. \textbf{MentalEval-Treat} uses the treatment recommendation together with the case context and diagnostic conclusion. Each sample is paired with the corresponding expert-defined rubric $\mathcal{B}$.

\paragraph{Output Format.}
Each training target contains a rubric-grounded evaluation trajectory and a final score. The trajectory performs evidence auditing, level-wise judgment, exclusion-rule checking, and final decision making. The final rating is constrained to a fixed marker format:
\begin{equation}
\texttt{[DECISION\_START]}r\texttt{[DECISION\_END]},
\qquad r\in\{1,2,3,4,5\}.
\end{equation}
This format allows automatic score parsing while preserving the intermediate rationale for interpretability.

\subsection{Stage 1: Rubric-Grounded SFT}

\paragraph{Judge-Ensemble Supervision.}
The first stage provides a cold start for learning rubric-grounded evaluation behavior. For each stage-level sample $z_i$, we use a judge ensemble $\mathcal{J}$ of five strong LLMs, including \texttt{DeepSeek-V4-Pro}, \texttt{KIMI-K2.6}, \texttt{GPT-5.4}, \texttt{Claude-Opus-4.6}, and \texttt{Gemini-3.1-Pro}, to generate evaluation trajectories and scores:
\begin{equation}
(s_{i,j}, r_{i,j}) = J_j(z_i, \mathcal{B}),
\qquad
j \in \mathcal{J},
\quad
r_{i,j}\in\{1,\ldots,5\},
\end{equation}
where $s_{i,j}$ denotes the rubric-grounded evaluation trajectory and $r_{i,j}$ denotes the assigned score.

\paragraph{Consensus Filtering.}
To obtain reliable supervision, we retain only judge outputs with sufficient score agreement. The consensus score for $z_i$ is defined as:
\begin{equation}
r_i^{*}
=
\arg\max_{r\in\{1,\ldots,5\}}
\sum_{j\in\mathcal{J}}
\mathbf{1}[r_{i,j}=r].
\end{equation}
If at least $k$ judges assign the consensus score $r_i^{*}$, trajectories with score $r_i^{*}$ are retained as high-quality SFT targets. Otherwise, the candidate set is treated as low-confidence and reserved for preference construction in Stage 2. Because extreme scores are underrepresented, we additionally select near-boundary samples and rewrite 2-point cases toward 1-point quality and 4-point cases toward 5-point quality. These augmented samples are retained only after judge-ensemble verification.

\paragraph{SFT Objective and Settings.}
Let $\mathcal{D}_{\mathrm{SFT}}$ denote the retained set of stage-level samples and high-quality evaluation trajectories. The SFT objective is:
\begin{equation}
\mathcal{L}_{\mathrm{SFT}}
=
-\mathbb{E}_{(z,s)\sim\mathcal{D}_{\mathrm{SFT}}}
\log \pi_{\theta}(s\mid z).
\end{equation}
This process yields 45,236 pairs of stage-level interaction data and high-quality evaluation trajectories for rubric-grounded SFT, with an average of 9,047.2 training samples per evaluator. All five evaluators are trained independently for 3 epochs with a learning rate of $5\times10^{-5}$ and a global batch size of 64.

Since different evaluation dimensions have different input lengths, we use dimension-specific cutoff lengths during SFT training, as shown in Table~\ref{tab:mentaleval_sft_cutoff}. Empathy and professionalism use a cutoff length of 16,384 because their inputs contain long multi-turn doctor--patient dialogue trajectories, while clinical-note quality, diagnostic rigor, and treatment appropriateness use a cutoff length of 8,192.

\begin{table}[t]
\centering
\caption{Cutoff length settings for SFT training across MentalEval dimensions.}
\label{tab:mentaleval_sft_cutoff}
\small
\setlength{\tabcolsep}{8pt}
\renewcommand{\arraystretch}{1.12}
\begin{tabular}{lll}
\toprule
\textbf{Evaluator} & \textbf{Dimension} & \textbf{Cutoff Length} \\
\midrule
MentalEval-Emp & Communication empathy & 16,384 \\
MentalEval-Pro & Interviewing professionalism & 16,384 \\
MentalEval-Note & Clinical-note quality & 8,192 \\
MentalEval-Diag & Diagnostic rigor & 8,192 \\
MentalEval-Treat & Treatment appropriateness & 8,192 \\
\bottomrule
\end{tabular}
\end{table}

\subsection{Stage 2: Expert-Guided DPO}

\paragraph{Preference Construction.}
After cold-start training, we further align MentalEval with specialist judgment using expert-guided preference optimization. For each low-confidence candidate set from Stage 1,
\begin{equation}
\mathcal{S}_i
=
\{(s_{i,1},r_{i,1}),\ldots,(s_{i,5},r_{i,5})\},
\end{equation}
we present the original stage-level interaction data $z_i$ and the candidate evaluation trajectories to clinicians. The clinician selects the trajectory $s_i^{*}$ whose rating and rationale best match the clinical quality of $z_i$. Candidate trajectories with the same score as $s_i^{*}$ are treated as chosen responses, while the remaining trajectories are treated as rejected responses:
\begin{equation}
\mathcal{S}_i^{+}
=
\{s_{i,j}\mid r_{i,j}=r_i^{*}\},
\qquad
\mathcal{S}_i^{-}
=
\{s_{i,j}\mid r_{i,j}\ne r_i^{*}\}.
\end{equation}
We then construct preference triples $(z_i,s_i^{+},s_i^{-})$ for DPO training. Through this process, we collect 3,823 preference triples.

\paragraph{DPO Objective and Settings.}
For each evaluator, we use the corresponding SFT checkpoint as the reference model $\pi_{\mathrm{ref}}$ and optimize the policy model $\pi_{\theta}$ with the DPO objective:
\begin{equation}
\mathcal{L}_{\mathrm{DPO}}
=
-\mathbb{E}_{(z,s^+,s^-)}
\log \sigma
\left(
\beta
\left[
\log\frac{\pi_{\theta}(s^+\mid z)}
{\pi_{\mathrm{ref}}(s^+\mid z)}
-
\log\frac{\pi_{\theta}(s^-\mid z)}
{\pi_{\mathrm{ref}}(s^-\mid z)}
\right]
\right).
\end{equation}
The DPO stage further optimizes each evaluator for 3 epochs with a learning rate of $1\times10^{-6}$, a global batch size of 32, and a preference coefficient $\beta=0.3$. All training experiments are conducted on 8 NVIDIA A800 GPUs.

\subsection{Inference and Scoring}

At inference time, each evaluator receives the evaluation context and its corresponding rubric. It generates a rubric-grounded rationale and terminates with a fixed decision marker. The final score is parsed as:
\begin{equation}
\hat{r}^{d}
=
\mathrm{Parse}
\left(
\pi_{\theta_d}(z^d,\mathcal{B}^d)
\right),
\qquad
\hat{r}^{d}\in\{1,\ldots,5\},
\end{equation}
where $d\in\{\mathrm{Emp},\mathrm{Pro},\mathrm{Note},\mathrm{Diag},\mathrm{Treat}\}$ denotes the evaluation dimension. The final subjective assessment is reported as five separate dimension-specific scores:
\begin{equation}
\hat{\mathbf{r}}_c
=
\left[
\hat{r}^{\mathrm{Emp}}_c,
\hat{r}^{\mathrm{Pro}}_c,
\hat{r}^{\mathrm{Note}}_c,
\hat{r}^{\mathrm{Diag}}_c,
\hat{r}^{\mathrm{Treat}}_c
\right].
\end{equation}
This preserves the interpretability of subjective assessment and allows MentalHospital to identify which clinical capabilities are strong or weak for each doctor agent.

\section{Evaluation Metrics}
\label{app:metrics}

\subsection{Objective Comparison}

Objective comparison evaluates whether the doctor agent recovers or predicts EHR-derived reference information. The reference data are constructed from structured case records, including patient-side evidence, auxiliary examinations, clinical formulation, diagnostic labels, and treatment outputs.

\paragraph{Clinical History Collection.}
For clinical interviewing, we evaluate how much patient-side evidence is elicited during the dialogue. For each case $c$, the EHR-derived patient evidence is divided into chief-complaint checkpoints $\mathcal{K}^{\mathrm{cc}}_c$ and mental-status checkpoints $\mathcal{K}^{\mathrm{mse}}_c$. During the interaction, each patient response returns a grounding field indicating which checkpoints are disclosed. Let $\hat{\mathcal{K}}^{\mathrm{cc}}_c$ and $\hat{\mathcal{K}}^{\mathrm{mse}}_c$ denote the unique checkpoints elicited across all dialogue turns. We compute:
\begin{equation}
\mathrm{Cov}^{\mathrm{cc}}_c
=
\frac{|\hat{\mathcal{K}}^{\mathrm{cc}}_c|}
{|\mathcal{K}^{\mathrm{cc}}_c|},
\qquad
\mathrm{Cov}^{\mathrm{mse}}_c
=
\frac{|\hat{\mathcal{K}}^{\mathrm{mse}}_c|}
{|\mathcal{K}^{\mathrm{mse}}_c|}.
\end{equation}
The overall interviewing coverage is:
\begin{equation}
\mathrm{Cov}^{\mathrm{inter}}_c
=
\frac{
|\hat{\mathcal{K}}^{\mathrm{cc}}_c|
+
|\hat{\mathcal{K}}^{\mathrm{mse}}_c|
}{
|\mathcal{K}^{\mathrm{cc}}_c|
+
|\mathcal{K}^{\mathrm{mse}}_c|
}.
\end{equation}
These metrics reflect the completeness of information elicitation during psychiatric interviewing. Higher coverage indicates that the doctor agent recovers more clinically relevant patient-side evidence.

\paragraph{Examination Recommendation.}
For examination recommendation, the reference set $\mathcal{A}^{\mathrm{exam}}_c$ consists of auxiliary examination items available in the structured EHR record. The doctor agent outputs a requested examination set $\hat{\mathcal{A}}^{\mathrm{exam}}_c$. The examination module returns an EHR-derived result if a requested item can be matched to an available examination; otherwise, it returns \texttt{NONE}. We define:
\begin{equation}
\mathrm{TP}=|\hat{\mathcal{A}}^{\mathrm{exam}}_c \cap \mathcal{A}^{\mathrm{exam}}_c|,
\quad
\mathrm{FP}=|\hat{\mathcal{A}}^{\mathrm{exam}}_c \setminus \mathcal{A}^{\mathrm{exam}}_c|,
\quad
\mathrm{FN}=|\mathcal{A}^{\mathrm{exam}}_c \setminus \hat{\mathcal{A}}^{\mathrm{exam}}_c|.
\end{equation}
We then compute:
\begin{equation}
\mathrm{Precision}
=
\frac{\mathrm{TP}}{\mathrm{TP}+\mathrm{FP}},
\qquad
\mathrm{Recall}
=
\frac{\mathrm{TP}}{\mathrm{TP}+\mathrm{FN}},
\end{equation}
\begin{equation}
\mathrm{F1}
=
\frac{2\cdot \mathrm{Precision}\cdot \mathrm{Recall}}
{\mathrm{Precision}+\mathrm{Recall}},
\qquad
\mathrm{Jaccard}
=
\frac{\mathrm{TP}}{\mathrm{TP}+\mathrm{FP}+\mathrm{FN}}.
\end{equation}
Precision measures whether the requested examinations are appropriate, recall measures whether available reference examinations are recovered, F1 gives their harmonic mean, and Jaccard measures set-level overlap between predicted and reference examinations.

\paragraph{Clinical-Note Generation.}
For clinical-note generation, the generated note $\hat{n}_c$ is compared with the EHR-derived reference formulation $n_c$. Both texts are parsed into clinically relevant regions, including chief complaint, physical examination, mental examination, and past/family history. Each region contains a predefined checkpoint set $\mathcal{K}^{\mathrm{note}}_{c,r}$, where $r$ indexes the region. For each checkpoint $k$, we evaluate whether the generated note covers the reference item:
\begin{equation}
\mathrm{Cov}(k,\hat{n}_c)
=
\mathbb{I}[k \preceq \hat{n}_c],
\end{equation}
where $k \preceq \hat{n}_c$ indicates that the checkpoint is covered by the generated note, determined by exact matching followed by LLM-assisted semantic matching when necessary. The region-level coverage is:
\begin{equation}
\mathrm{Cov}^{\mathrm{note}}_{c,r}
=
\frac{1}{|\mathcal{K}^{\mathrm{note}}_{c,r}|}
\sum_{k\in \mathcal{K}^{\mathrm{note}}_{c,r}}
\mathbb{I}[k \preceq \hat{n}_c].
\end{equation}
The overall note coverage is the average across regions:
\begin{equation}
\mathrm{Cov}^{\mathrm{note}}_c
=
\frac{1}{|\mathcal{R}|}
\sum_{r\in\mathcal{R}}
\mathrm{Cov}^{\mathrm{note}}_{c,r},
\end{equation}
where $\mathcal{R}$ denotes the set of clinical-note regions. This metric reflects whether the generated note preserves key EHR-grounded clinical information and supports downstream diagnostic review.

\paragraph{Definitive Diagnosis.}
For definitive diagnosis, the model outputs an ordered diagnosis list $\hat{\mathcal{D}}_c$, and the EHR provides the reference disorder set $\mathcal{D}_c$ and category set $\mathcal{C}_c$. We evaluate both disorder-level and category-level diagnosis. For set-level matching, we compute:
\begin{equation}
\mathrm{Precision}
=
\frac{|\hat{\mathcal{Y}}_c\cap \mathcal{Y}_c|}
{|\hat{\mathcal{Y}}_c|},
\qquad
\mathrm{Recall}
=
\frac{|\hat{\mathcal{Y}}_c\cap \mathcal{Y}_c|}
{|\mathcal{Y}_c|},
\end{equation}
\begin{equation}
\mathrm{F1}
=
\frac{2\cdot \mathrm{Precision}\cdot \mathrm{Recall}}
{\mathrm{Precision}+\mathrm{Recall}},
\qquad
\mathrm{Jaccard}
=
\frac{|\hat{\mathcal{Y}}_c\cap \mathcal{Y}_c|}
{|\hat{\mathcal{Y}}_c\cup \mathcal{Y}_c|},
\end{equation}
where $\mathcal{Y}_c$ can be either the reference disorder set or the reference category set. Category-level prediction is obtained by mapping predicted disorders to their corresponding diagnostic categories.

To evaluate diagnostic prioritization, we additionally compute ranking metrics over the ordered disorder list. Let $d_c^{(1)}$ denote the reference primary diagnosis and $\hat{d}_c^{(i)}$ denote the $i$-th predicted diagnosis. Exact Match is:
\begin{equation}
\mathrm{EM}_c
=
\mathbb{I}[\hat{d}_c^{(1)} = d_c^{(1)}].
\end{equation}
Hit@$K$ is:
\begin{equation}
\mathrm{Hit@}K_c
=
\mathbb{I}[d_c^{(1)} \in \{\hat{d}_c^{(1)},\ldots,\hat{d}_c^{(K)}\}],
\end{equation}
and Reciprocal Rank is:
\begin{equation}
\mathrm{RR}_c
=
\frac{1}{\mathrm{rank}(d_c^{(1)})},
\end{equation}
where $\mathrm{rank}(d_c^{(1)})$ is the predicted rank of the reference primary diagnosis. If the primary diagnosis is absent from the predicted list, $\mathrm{RR}_c=0$.

We also report nDCG to measure whether the predicted ranking is consistent with the reference diagnostic priority:
\begin{equation}
\mathrm{DCG}_c
=
\sum_{i=1}^{|\hat{\mathcal{D}}_c|}
\frac{\mathrm{rel}(\hat{d}_c^{(i)})}{\log_2(i+1)},
\qquad
\mathrm{nDCG}_c
=
\frac{\mathrm{DCG}_c}{\mathrm{IDCG}_c},
\end{equation}
where $\mathrm{rel}(\cdot)$ is assigned according to the reference diagnostic order, and $\mathrm{IDCG}_c$ is the ideal DCG under the reference ranking. Set metrics reflect diagnostic coverage and accuracy, while ranking metrics reflect whether the model places the most clinically important diagnoses at higher ranks.

\paragraph{Treatment Planning.}
For treatment planning, there is no EHR-derived gold treatment set used for objective matching. We therefore record the output length of the generated treatment plan:
\begin{equation}
\mathrm{Len}^{\mathrm{treat}}_c = |\hat{p}_c|,
\end{equation}
where $\hat{p}_c$ denotes the generated treatment plan. This metric only reflects the amount of generated treatment content; treatment quality is assessed through the subjective treatment-appropriateness rubric.

\subsection{Subjective Assessment}

Subjective assessment evaluates process-level clinical quality that cannot be fully captured by EHR-grounded reference matching. MentalEval assigns a 1--5 score for each dimension using expert-defined rubrics. For each dimension $d$, the evaluator receives the corresponding evaluation context $x_c^d$ and rubric $\mathcal{B}^d$, then outputs a rubric-grounded rationale and a final score:
\begin{equation}
\hat{r}_c^d
=
\mathrm{MentalEval}_d(x_c^d,\mathcal{B}^d),
\qquad
\hat{r}_c^d\in\{1,2,3,4,5\}.
\end{equation}
The final subjective score vector for case $c$ is:
\begin{equation}
\hat{\mathbf{r}}_c
=
[
\hat{r}^{\mathrm{Emp}}_c,
\hat{r}^{\mathrm{Prof}}_c,
\hat{r}^{\mathrm{Qual}}_c,
\hat{r}^{\mathrm{Rigor}}_c,
\hat{r}^{\mathrm{App}}_c
].
\end{equation}

\paragraph{Communication Empathy.}
Communication empathy is evaluated on the complete doctor--patient dialogue from the interviewing stage. The score measures whether the doctor agent accurately recognizes and responds to the patient's emotional and subjective experience. It emphasizes contextual emotional fit, recognition of implicit feelings, appropriate use of verbal and nonverbal cues, and avoidance of generic or template-like reassurance.

\paragraph{Interviewing Professionalism.}
Interviewing professionalism is also evaluated on the complete doctor--patient dialogue. The score measures whether the doctor agent conducts the interview with a clear clinical goal, maintains thematic progression, gathers diagnostically useful information, and converges toward relevant symptom clusters. Higher scores require targeted clarification, appropriate pacing, and differential diagnostic awareness rather than merely asking many questions.

\paragraph{Clinical-Note Quality.}
Clinical-note quality is evaluated on the generated clinical note. The score measures whether the note has psychiatric documentation quality, including structural completeness, information sufficiency, evidence organization, and diagnostic utility. The rubric first checks essential elements such as chief complaint, history of present illness, mental status examination, relevant history, risk assessment, negative findings, and functional impact, and then judges whether the note can support diagnostic review.

\paragraph{Diagnostic Rigor.}
Diagnostic rigor is evaluated on the model-generated diagnostic reasoning text. This score does not directly measure whether the final diagnosis is correct; instead, it measures whether the diagnostic argument is clear, criterion-grounded, evidence-supported, and auditable. Higher scores require substantive use of diagnostic criteria, case-based justification for key judgments, criterion-level differential diagnosis, and coherent diagnostic convergence.

\paragraph{Treatment Appropriateness.}
Treatment appropriateness is evaluated on the generated treatment plan together with the case context and diagnostic conclusion. The score measures whether the plan is aligned with the diagnosis, symptom severity, illness stage, risk status, and patient-specific constraints. Higher scores require complete treatment planning, individualized trade-offs, and dynamic management strategies for future treatment response, adverse effects, adherence changes, or risk escalation.

\paragraph{Aggregation.}
For reporting, objective metrics are averaged over all evaluated cases. Subjective scores are averaged separately for each dimension:
\begin{equation}
\overline{r}^{d}
=
\frac{1}{|\mathcal{C}|}
\sum_{c\in\mathcal{C}}
\hat{r}_c^d,
\end{equation}
where $\mathcal{C}$ denotes the evaluation case set. We report the five subjective dimensions separately to preserve interpretability rather than collapsing them into a single overall score.

\section{Patient Ablation}
\label{app:patient_ablation}

We evaluate patient construction fidelity through both objective evaluation and subjective assessment, so as to examine whether the standardized patient is not only factually grounded in the EHR but also clinically realistic as an interactive patient agent. The objective evaluation focuses on evidence-level faithfulness. We report evidence coverage (Cov.), evidence precision (Prec.), and hallucination rate (Halluc.). Specifically, we sample 20 EHR-derived cases and construct 12 checkpoint-grounded probes for each case, yielding 240 patient responses for each construction variant. Each probe is designed to target one or more EHR-derived patient-side facts, including chief-complaint events, symptom details, illness-course information, and mental-status-related behavioral cues.

The probe set covers three complementary question types. Open-recall probes ask the patient to describe relevant experiences with minimal guidance, testing whether the patient can naturally disclose EHR-supported information when prompted broadly. Targeted-fact probes ask about specific symptom attributes, such as onset, frequency, severity, contextual triggers, or functional impact, testing whether the patient can retrieve and express the correct case-specific details. False-premise probes intentionally introduce unsupported or misleading assumptions, testing whether the patient can resist fabricating facts and avoid accepting information that is absent from the EHR. Together, these probes evaluate both the recall capacity and the boundary control of the patient construction.

For objective scoring, we use a Multi-LLM Jury to perform LLM-assisted matching between patient responses and EHR-derived checkpoints. The matching process is repeated three times, and the averaged results are used to reduce stochastic judgment variance. Evidence coverage measures the proportion of expected patient-side checkpoints that are correctly disclosed in response to the probes. Evidence precision measures the proportion of stated clinical facts that can be supported by the EHR-derived evidence. Hallucination rate measures whether the patient introduces unsupported symptoms, histories, experiences, or causal explanations. Thus, coverage reflects whether the patient can provide sufficient case information, precision reflects whether disclosed information remains faithful to the EHR, and hallucination rate reflects whether the simulation violates evidence boundaries.

For subjective assessment, three clinicians blindly rate each case-level multi-turn interaction, which consists of the 12 probes and the corresponding patient responses. The clinicians score each construction variant on a 1--5 scale across three dimensions: clinical realism, patient fidelity, and patient distinctiveness. Clinical realism measures whether the patient behaves like a plausible psychiatric patient in terms of response style, affective expression, and interaction pattern. Patient fidelity measures whether the patient remains consistent with the underlying EHR-derived case evidence. Patient distinctiveness measures whether different patients exhibit case-specific presentations rather than collapsing into generic psychiatric personas. The three clinician ratings are averaged to obtain the final case-level subjective score.

As shown in Table~\ref{tab:patient_ablation}, naive prompting achieves reasonable factual faithfulness but performs poorly in subjective fidelity, suggesting that directly exposing the EHR to the model is insufficient for realistic patient simulation. This variant can often repeat factual content from the record, but its responses tend to be overly explicit, summary-like, or clinically unnatural. Adding the representation skill substantially improves clinical realism and patient distinctiveness, indicating that individualized symptom presentation and behavioral rendering are important for producing patient-like interactions rather than static case summaries. In contrast, the memory skill mainly improves evidence coverage, precision, and hallucination control, showing that selective retrieval helps the patient disclose relevant information when asked while avoiding unsupported or premature disclosure.

The full patient construction achieves the best overall performance across both objective and subjective metrics. This result suggests that faithful psychiatric simulation requires the combination of two complementary mechanisms: individualized symptom representation, which controls how the patient presents, and selective memory recall, which controls what evidence the patient discloses at each interaction step. Without representation skill, the patient may remain factually grounded but lack clinical realism. Without memory skill, the patient may appear expressive but become less controllable or more prone to unsupported disclosure. Their combination allows the standardized patient to maintain EHR faithfulness while supporting natural, gradual, and clinically meaningful psychiatric interviews.

\section{Faithfulness Questionnaire}
\label{app:faithfulness_questionnaire}

To further examine whether MentalHospital provides a clinically faithful and practically useful psychiatric simulation environment, we conduct a questionnaire-based assessment with domain participants. The study includes 5 psychiatrists, 3 psychologists, and 14 medical trainees. Unlike objective comparison, which evaluates whether model outputs match EHR-derived references, this questionnaire focuses on the perceived clinical plausibility, patient faithfulness, evaluation validity, and training utility of the overall environment. This complementary assessment is necessary because a psychiatric simulation system should not only reproduce reference information, but also support interactions that are recognizable, interpretable, and useful from a clinical perspective.

The questionnaire is organized into five dimensions: clinical realism (CR), patient fidelity (PF), patient distinctiveness (PD), evaluation credibility (EC), and training suitability (TS). Clinical realism assesses whether the interaction workflow resembles real psychiatric interviewing and clinical decision-making, including the gradual progression from symptom inquiry to assessment, diagnosis, and treatment planning. Patient fidelity evaluates whether the simulated patient remains consistent with the underlying psychiatric case information, especially in symptom expression, behavioral presentation, and response stability across turns. Patient distinctiveness measures whether different simulated patients preserve meaningful individual differences rather than collapsing into homogeneous or generic response patterns. Evaluation credibility assesses whether the dual-track evaluation protocol captures clinically meaningful aspects of doctor-agent performance. Training suitability measures whether the system is useful for psychiatric education, interview practice, clinical reasoning training, and medical AI evaluation.

Each item is rated on a 5-point Likert scale, where higher scores indicate stronger agreement. For each dimension, we compute the mean score across its corresponding items and participants. We also report the overall mean score across all questionnaire items. To summarize response polarity, we define scores of 4 or 5 as positive responses, scores of 1 or 2 as negative responses, and score 3 as neutral. We report the positive and negative response rates over all collected ratings, together with 95\% bootstrap confidence intervals. This allows us to assess not only the average perceived quality of MentalHospital, but also the proportion of participants who explicitly endorse or reject its clinical faithfulness.

Table~\ref{tab:expert_questionnaire} lists the full questionnaire items. The items are designed to cover both the construction quality of MentalHospital and its downstream utility. The first three dimensions focus on whether the simulated clinical environment and patient agents are faithful to psychiatric practice and case-specific evidence. Specifically, clinical realism examines the plausibility of the overall clinical workflow, patient fidelity examines whether the AI patient remains grounded in the provided case, and patient distinctiveness examines whether the system can preserve case-level heterogeneity. The last two dimensions examine whether the evaluation protocol is clinically credible and whether the environment is suitable for training and assessment.

In addition, the questionnaire is designed to separate fidelity at different levels. Patient fidelity focuses on whether an individual simulated patient is consistent with the corresponding EHR-derived case, whereas patient distinctiveness focuses on whether different cases lead to distinguishable patient behaviors and symptom profiles. This distinction is important because a patient agent may appear realistic in isolation while still producing overly similar responses across cases. Similarly, evaluation credibility is separated from training suitability: the former concerns whether the evaluation metrics and rubrics provide meaningful judgments of doctor-agent performance, while the latter concerns whether the environment can support practice, feedback, and educational use.

Before rating, participants are shown representative MentalHospital interactions and evaluation outputs, including patient responses, doctor-agent inquiries, clinical notes, diagnostic reasoning, treatment planning, and corresponding evaluation results. Participants are instructed to rate the system according to clinical plausibility and usefulness rather than surface fluency alone. This instruction is intended to reduce the risk that high ratings are assigned merely because the generated text is coherent or polished. Instead, the questionnaire emphasizes whether the simulated workflow, patient behavior, and evaluation design are clinically meaningful under psychiatric practice.

The resulting questionnaire scores are used as an external validation of MentalHospital's faithfulness. While the objective metrics quantify whether doctor agents recover EHR-derived evidence and produce correct task outputs, the questionnaire reflects whether domain participants perceive the environment as clinically plausible and useful. Therefore, the faithfulness questionnaire provides a human-centered complement to the benchmark's quantitative evaluation protocol, supporting a broader assessment of whether MentalHospital can serve as a reliable environment for psychiatric interaction simulation, doctor-agent evaluation, and training-oriented use.

\begin{table}[t]
\caption{Expert questionnaire for evaluating the clinical fidelity of MentalHospital.}
\label{tab:expert_questionnaire}
\centering
\small
\setlength{\tabcolsep}{5pt}
\renewcommand{\arraystretch}{1.15}
\begin{tabular}{p{0.24\linewidth}p{0.68\linewidth}}
\toprule
\textbf{Dimension} & \textbf{Questionnaire Item} \\
\midrule
Clinical Realism & The interaction workflow of MentalHospital is consistent with real-world psychiatric interviewing practice. \\
Clinical Realism & The system supports a clinically plausible progression from symptom inquiry to assessment, diagnosis, and treatment planning. \\
Clinical Realism & The dialogue process reflects the sequential and exploratory nature of psychiatric evaluation. \\
Patient Fidelity & The AI patient’s responses are consistent with the provided psychiatric case information. \\
Patient Fidelity & The AI patient’s symptom expressions are clinically plausible for the corresponding psychiatric condition. \\
Patient Fidelity & The AI patient maintains stable behavioral and emotional characteristics during the interaction. \\
Patient Distinctiveness & Different AI patients exhibit distinguishable symptom profiles and response patterns. \\
Patient Distinctiveness & The simulated patients avoid homogeneous or generic responses across different cases. \\
Patient Distinctiveness & The system can reflect meaningful individual differences in patient presentation, history, and communication style. \\
Evaluation Credibility & The objective evaluation metrics capture clinically relevant task outcomes in psychiatric assessment. \\
Evaluation Credibility & The subjective assessment dimensions are appropriate for evaluating clinical interaction quality. \\
Evaluation Credibility & The system’s evaluation results are interpretable and clinically meaningful. \\
Evaluation Credibility & The combination of objective comparison and subjective assessment provides a credible evaluation of doctor-agent performance. \\
Training Suitability & MentalHospital is suitable for supporting psychiatric interview training. \\
Training Suitability & MentalHospital can help trainees practice clinical reasoning in realistic psychiatric cases. \\
Training Suitability & Overall, MentalHospital provides a clinically valuable environment for psychiatric education and medical AI evaluation. \\
\bottomrule
\end{tabular}
\end{table}

\section{Frontend Workflow and Interaction Design of \textsc{MentalHospital}}
\label{app:Frontend Workflow}

\subsection{Overview}

This appendix describes the frontend workflow and interaction design of \textsc{MentalHospital}. The frontend is designed as the doctor-facing interaction layer of an executable psychiatric training environment. Rather than serving as a standalone web interface, it operationalizes the staged clinical workflow of \textsc{MentalHospital} by allowing medical students to enter cases, interview standardized patients, request auxiliary examinations, complete structured case writing, submit diagnostic hypotheses, formulate treatment plans, and review backend-generated evaluation reports.

The frontend plays an important role in preserving the sequential nature of psychiatric assessment. In real clinical practice, information is not obtained all at once. Clinicians gradually collect evidence through interview, examination, diagnostic reasoning, and treatment planning. Following this principle, the frontend of \textsc{MentalHospital} organizes user interaction into multiple stages and controls the information exposed to the user at each stage. This design prevents the training episode from becoming a single unstructured prompt-response exchange and instead supports a clinically grounded, stage-aware interaction process.

\subsection{Stage-wise Interaction Workflow}

The frontend workflow begins with a case-selection interface. After a case is selected, the user enters a sequential training episode consisting of four main clinical stages: patient interview, auxiliary examination, case writing and diagnosis, and treatment planning. Each stage provides a specific interaction workspace and passes clinically relevant information to subsequent stages.

\paragraph{Phase 1: Patient Interview.}
In the first phase, the frontend provides a multi-turn interview interface in which the user acts as the doctor and interacts with a standardized psychiatric patient. The patient responses are generated dynamically by the backend according to the selected case and the ongoing dialogue context. The frontend records the complete dialogue trajectory and maintains the interview history throughout the episode. This phase is intended to evaluate the user's ability to collect psychiatric symptoms, explore illness history, assess mental status, and establish a clinically appropriate doctor-patient interaction.

\paragraph{Phase 2: Auxiliary Examination.}
After completing the interview, the frontend transitions to an auxiliary examination interface. In this phase, the user selects examinations that may help clarify the patient's condition. The frontend presents available examination options and displays backend-returned examination results according to the user's selections. Compared with the interview phase, this stage shifts the interaction from conversational information gathering to structured evidence acquisition. The selected examinations and their results are preserved for later diagnostic reasoning and final evaluation.

\paragraph{Phase 3: Case Writing and Diagnosis.}
The third phase combines the information obtained from the interview and auxiliary examinations into a structured case-writing and diagnostic workspace. The user is asked to complete a clinical case summary and then submit diagnostic hypotheses together with supporting reasons. This phase is designed to assess whether the user can organize scattered clinical evidence into a coherent psychiatric case formulation and whether the proposed diagnosis is consistent with the collected information.

Importantly, the frontend does not treat diagnosis as an isolated label-selection task. Instead, it provides an interaction space in which case writing, diagnostic selection, and diagnostic reasoning are connected. This design reflects the clinical requirement that psychiatric diagnosis should be supported by symptom evidence, temporal course, differential considerations, and contextual information.

\paragraph{Phase 4: Treatment Planning.}
In the fourth phase, the frontend presents a treatment-planning interface. The user is asked to formulate a treatment plan across multiple clinically meaningful categories, including pharmacological treatment, psychotherapy, neuromodulation or other somatic treatment, psychosocial intervention and rehabilitation, and auxiliary treatment. This stage extends the training episode beyond diagnosis and requires the user to consider how diagnostic conclusions can be translated into an actionable intervention plan.

By placing treatment planning after interview, examination, case writing, and diagnosis, the frontend preserves the intended clinical order of the system. The final submitted treatment plan is stored together with previous-stage outputs and becomes part of the complete episode record.

\subsection{Backend-coordinated Evaluation}

After the user submits the treatment plan, the frontend enters a unified evaluation stage. The frontend does not compute scores, generate judgments, or define evaluation outcomes locally. Instead, it invokes backend evaluation endpoints and renders the returned results as an integrated report. This design ensures that the evaluation process remains consistent with the backend evaluator and avoids inconsistencies caused by frontend-side scoring logic.

The unified evaluation report summarizes the user's performance across the full training episode. It covers multiple dimensions, including interview quality, auxiliary examination appropriateness, case-writing quality, diagnostic correctness, diagnostic reasoning, and treatment-plan completeness. By collecting all stage-level assessments into a single final report, the system provides a closed-loop view of the user's clinical performance rather than presenting fragmented feedback after each individual stage.

This backend-coordinated evaluation design also supports reproducibility. Since the frontend only displays backend-generated evaluation outputs, the scoring logic can be maintained, updated, and audited centrally on the backend side. The frontend therefore functions as an execution and visualization layer, while the backend remains responsible for simulation, evaluation, and final result generation.

\subsection{Episode Logging and Traceability}

A key design goal of the frontend is to support complete episode traceability. During the interaction process, the frontend collects and passes forward the dialogue history, selected auxiliary examinations, examination results, case-writing content, diagnostic submissions, treatment plans, and final evaluation outputs. These data are packaged as a complete training trajectory and stored by the backend.

The history interface allows previously completed episodes to be revisited. Each episode is treated as a full clinical training record rather than a temporary frontend session. This design enables users and instructors to review how the interaction unfolded, what evidence was collected, how the diagnosis was formulated, what treatment plan was proposed, and how the final evaluation was generated.

Such traceability is important for psychiatric training because errors may occur at different stages of the clinical reasoning process. For example, an incorrect diagnosis may result from insufficient interview exploration, inappropriate examination selection, incomplete case writing, or flawed diagnostic reasoning. By preserving the full trajectory, \textsc{MentalHospital} makes it possible to analyze not only the final answer but also the process that led to it.

\subsection{Design Rationale}

The frontend of \textsc{MentalHospital} is organized around three design principles.

First, the frontend should preserve the staged structure of psychiatric assessment. Each phase corresponds to a clinically meaningful step, and later stages depend on information collected earlier in the episode. This structure encourages users to reason progressively rather than relying on one-step answer generation.

Second, the frontend should separate interaction control from evaluation authority. The frontend is responsible for collecting user inputs, managing stage transitions, displaying backend outputs, and maintaining episode continuity. Evaluation logic remains on the backend, which ensures consistency between the implemented system and the intended evaluation framework.

Third, the frontend should support longitudinal review and training feedback. By saving complete episode records, the system allows each training attempt to be reviewed as a traceable clinical reasoning trajectory. This makes the frontend not only an interface for immediate interaction but also a component that supports later analysis, reflection, and educational feedback.

\subsection{Extensibility}

Because the frontend is organized around stage boundaries and backend-coordinated operations, it can support new cases, additional evaluation dimensions, and future task extensions without changing the overall episode logic. For example, new psychiatric case types can be introduced through backend case sampling, additional examination options can be added to the auxiliary examination stage, and new evaluation modules can be integrated into the unified report as long as the frontend preserves the same stage-wise execution structure.

This modular organization makes the frontend suitable for an evolving psychiatric training environment. Its main contribution is not in providing a generic web application, but in offering a stable execution surface through which the proposed simulation, reasoning, evaluation, and logging mechanisms can be experienced as a complete interactive system.

\subsection{Summary}

In summary, the frontend of \textsc{MentalHospital} serves as the doctor-facing execution layer of the proposed psychiatric training environment. It supports staged interaction, structured information acquisition, backend-coordinated evaluation, and complete episode traceability. By connecting case entry, patient interview, auxiliary examination, case writing, diagnosis, treatment planning, final evaluation, and history review into a single workflow, the frontend helps instantiate \textsc{MentalHospital} as an executable and traceable clinical training system.